\pdfoutput=1

\PassOptionsToPackage{table,dvipsnames}{xcolor}

\documentclass[11pt]{article}

\usepackage{enumitem}
\usepackage{geometry}
\usepackage{xcolor}

\usepackage[most]{tcolorbox}
\usepackage[final]{acl}

\usepackage[T1]{fontenc}
\usepackage[utf8]{inputenc}
\usepackage{times}
\usepackage{latexsym}
\usepackage{microtype}
\usepackage{inconsolata}

\usepackage{amsmath,amssymb}

\usepackage{graphicx}
\usepackage{subcaption}
\usepackage{caption}
\usepackage{booktabs}
\usepackage{multirow}
\usepackage{array}
\usepackage{colortbl}    

\usepackage{natbib}

\usepackage{algorithm}
\usepackage{algpseudocode}
\usepackage{stfloats}
\usepackage{rotating}

\usepackage{xspace}
\usepackage{siunitx}
\usepackage{pifont}
\usepackage{titletoc}

\definecolor{light_blue}{RGB}{153,187,255}
\definecolor{cellred}{RGB}{255,71,76} 
\usepackage{svg}
\usepackage{twemojis}
\usepackage{multicol}

\definecolor{avgcolor}{RGB}{220,230,241}
\definecolor{avgcol}{RGB}{255,249,196}
\definecolor{maxcolor}{RGB}{34,139,34}
\definecolor{mincolor}{RGB}{178,34,34}
\definecolor{softline}{gray}{0.7}
\title{\textit{Spot the BlindSpots}: Systematic Identification and Quantification of Fine-Grained LLM Biases in Contact Center Summaries}

\author{Kawin Mayilvaghanan, Siddhant Gupta\thanks{\hspace{1pt} Work done during internship at Observe.AI}, \and Ayush Kumar \\
        \texttt{\{kawin.m,  siddhant.gupta, ayush\}@observe.ai}\\
        Observe.AI \\ Bangalore, India}

\begin{document}
\maketitle

\begin{abstract}
Abstractive summarization is a core application in contact centers, where Large Language Models (LLMs) generate millions of summaries of call transcripts daily. Despite their apparent quality, it remains unclear whether LLMs systematically under- or over-attend to specific aspects of the transcript, potentially introducing biases in the generated summary. While prior work has examined social and positional biases, the specific forms of bias pertinent to contact center operations—which we term `Operational Bias'—have remained unexplored. To address this gap, we introduce \textit{\textbf{BlindSpot}}, a framework built upon a taxonomy of 15 operational bias dimensions (e.g., disfluency, speaker, topic)  for the identification and quantification of these biases. \textit{\textbf{BlindSpot}} leverages an LLM as a zero-shot classifier to derive categorical distributions for each bias dimension in a pair of transcript and its summary. The bias is then quantified using two metrics: \textit{Fidelity Gap} (the JS Divergence between distributions) and \textit{Coverage} (the percentage of source labels omitted). Using \textit{\textbf{BlindSpot}}, we conducted an empirical study with 2500 real call transcripts and their summaries generated by 20 LLMs of varying scales and families (e.g., GPT, Llama, Claude). Our analysis reveals that biases are systemic and present across all evaluated models, regardless of size or family.

\end{abstract}

\section{Introduction and Related Works}

Contact centers are central to business operations, serving as the primary interface for customer support. Their capacity to deliver superior customer service is crucial for maintaining satisfaction, cultivating loyalty, and ultimately ensuring business success across various industries.  Within this context, abstractive call summarization~\citep{yuan2019abstractivedialogsummarizationsemantic} is a critical task that enables contact center agents to effectively document interactions for regulatory compliance, contextual handoffs to other agents, and future reference. These summaries also underpin crucial downstream processes such as agent performance evaluation, business intelligence, insights discovery, and regulatory audits. For instance, supervisors use them to assess protocol adherence, while aggregated data highlights issues to inform strategic decisions.

\begin{figure}[t]
 \centering
  \includegraphics[width=1\columnwidth]{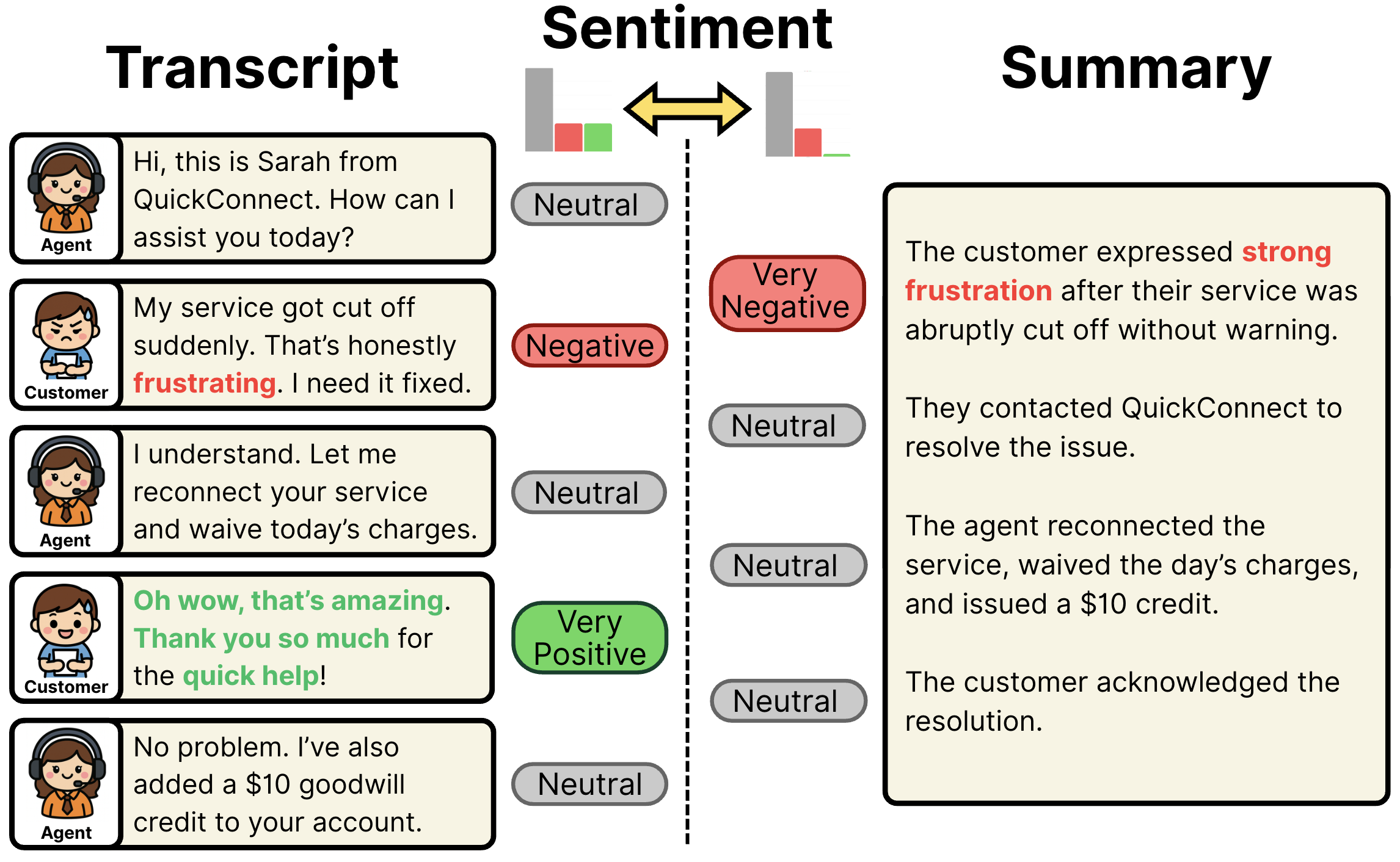}
  \caption{A call transcript (left) with mixed sentiment is contrasted with its summary (right). Although the summary is factually correct and complete, it \textbf{amplifies} the customer’s negative sentiment and \textbf{neutralizes} their positive sentiment towards resolution. This sentiment bias, invisible through contemporary summary evaluation metrics, underscores the importance of bias evaluation.}
  \label{fig:example}
\end{figure}

The emergence of Large Language Models (LLMs) has facilitated the automated generation of call summaries, producing fluent and coherent summaries at scale~\citep{sachdeva23_interspeech, thulke2024promptingfinetuningsmallllms}. Evaluating the quality of LLM-generated summaries presents a multifaceted challenge. Existing metrics~\citep{fabbri2021summeval, gao-wan-2022-dialsummeval} effectively capture general qualities like factual correctness, relevance, and coherence, often relying on human judgments via Likert-scale annotations. Complementing these assessments are automated reference-based metrics like BLEU~\citep{papineni-etal-2002-bleu}, ROUGE~\citep{lin-2004-rouge}, and BERTScore~\citep{zhang2019bertscore}, which provide quantitative measures of text similarity and overlap. More recently, "LLM-as-a-Judge"~\citep{liu2023gevalnlgevaluationusing, kim2024prometheus2opensource} has emerged, where another LLM is utilized to evaluate the quality of a summary, offering a potentially scalable and efficient alternative to human annotation.

However, these established quality metrics overlook a crucial aspect of fidelity: a summary can be factually correct and fluent, yet still be biased in how it represents the original interaction. While the field has extensively studied bias, their work has predominantly focused on two categories. First, \textbf{social and demographic biases}, focusing on attributes such as gender, race, and nationality \citep{nadeem-etal-2021-stereoset, li2020leveraginggraph, rudinger2018genderbias, zhu2024nationality}. Numerous methods have been proposed for detecting and mitigating these biases, including fairness-focused QA assessments \citep{wang2023causalentity}, similarity based \citep{zhou2023entitypolitical} and metrics like Information Density Metric (IDM) \citep{wang2024usersumbench}, Total Variation Distance (TVD) \citep{steen-markert-2024-bias}, and Fairness Gap \citep{olabisi2024positionbias}. Second, \textbf{structural biases}, notably position bias, the tendency to favor information based on its location in the text, have been documented \citep{wan2024positionalbiasfaithfulnesslongform, olabisi2024positionbias}.

Despite the complexity of these metrics, a critical gap remains: they fail to address a category of distortions that, while not necessarily factual errors, can severely undermine a summary's utility in a business context. This raises crucial questions of fidelity: do summaries accurately preserve customer sentiment? Do they equitably represent all parts of the conversation, or do they overstate the efficacy of an agent's proposed solution? We term these systematic deviations as operational biases: distortions in a summary that misrepresent the context of the original interaction. Such biases carry significant downstream consequences for agent evaluation, business intelligence, and customer satisfaction. To systematically identify and quantify these biases, our work makes the following contributions:
\begin{enumerate}
\item \textbf{Taxonomy of Operational Bias:} We define a taxonomy of 15 bias dimensions specific to the operational requirements of contact center summarization, grouped into five classes.
\item \textbf{The \textit{BlindSpot} Framework:} We introduce a fully-automated framework that quantifies bias by comparing the distributional properties of source transcripts and their summaries. 
\item \textbf{An Empirical Audit:} We conduct the first comprehensive benchmark of operational bias, evaluating 20  LLMs on a corpus of 2500 contact center transcripts.
\end{enumerate}

Our analysis extends beyond aggregate bias scores, using the \textit{BlindSpot} framework to provide a fine-grained view of representation. This allows us to identify specific labels that are systematically over- or under-represented by each model and reveal common failure modes. Crucially, this analysis is actionable: a targeted system prompt engineered from our findings reduced bias across nine different models, increasing average \textit{Coverage} by up to +4.87\% and measurably reducing the \textit{Fidelity Gap}.

Ultimately, this work provides a crucial toolset for moving beyond quality metrics toward a rigorous evaluation of summary biases. By systematically identifying and quantifying these biases, we lay the groundwork for building more accountable, reliable summarization systems for practical environments.

\section{Methodology}
In this section, we detail our methodology for identifying and quantifying biases in summaries. 

\subsection{Taxonomy of Operational Bias}
\label{subsec:bias_taxonomy}

\begin{table}[!hbt]
\centering
\scriptsize
\setlength{\tabcolsep}{3pt}
\begin{tabular}{p{0.28\linewidth} p{0.65\linewidth}}
\toprule
\textbf{Bias Dimension} & \textbf{Description} \\
\midrule
\multicolumn{2}{l}{\textit{1. Content \& Information Fidelity Dimensions}} \\
Entity Type & A bias here would reflect over- or under-representation of certain named entity type. \\
Topic & A bias in this dimension would indicate selective focus on certain topics while omitting others. \\
Solution & A bias would occur if certain solution types are consistently highlighted or downplayed. \\
Information Repetition & A bias in representing repeated information. \\
\midrule
\multicolumn{2}{l}{\textit{2. Conversational Structure \& Flow Dimensions}} \\
Position & A bias here would suggest overemphasis or neglect of particular stages of the interaction. \\
Turn Length & Bias in attending to short, medium, or long turns.  \\
Temporal Sequence & A bias would indicate reshuffling of the original order. \\
\midrule
\multicolumn{2}{l}{\textit{3. Speaker \& Role Representation Dimensions}} \\
Speaker & A bias here would suggest preferential inclusion of one speaker's perspective over the other. \\
Agent Action & A bias here would reflect selective inclusion or omission of particular agent behaviors. \\
\midrule
\multicolumn{2}{l}{\textit{4. Linguistic \& Stylistic Dimensions}} \\
Language Complexity & A bias here might result in oversimplification or unwarranted complexity. \\
Disfluency & A bias here would be reflected in either over-sanitizing or preserving disfluencies inconsistently. \\
Politeness & A bias here would occur if the summary changes the tone to be more neutral, impolite, or overly formal. \\
\midrule
\multicolumn{2}{l}{\textit{5. Affective \& Pragmatic Interpretation Dimensions}} \\
Sentiment & A bias would misrepresent the speaker's sentiment. \\
Emotion Shift &  A bias in emotion shift as amplified, attenuated, or neutralized relative to the original sentiment. \\
Urgency & A bias here would reflect over- or under-emphasizing the immediacy of concerns or requests. \\
\bottomrule
\end{tabular}
\caption{Taxonomy of 15 bias dimensions, defined across five classes for contact center summarization. See Appendix~\ref{appendix:taxonomy} and Table \ref{tab:bias_dimensions_full} for detailed definitions, and labels. }
\label{tab:bias_taxonomy}
\end{table}

To evaluate operational bias, we propose a taxonomy of 15 dimensions (Table \ref{tab:bias_taxonomy}). The framework moves beyond simple bias identification to link specific bias dimension to tangible operational outcomes, grouping dimensions into five classes based on core functional requirements of a contact center summary.

The first three classes address the foundational integrity of the summary: its narrative structure, and participant representation. \textit{\textbf{Content \& Information Fidelity}} ensures the summary is a reliable and actionable record; for instance, \textit{Entity Type} bias can render a summary useless by omitting key identifiers, while \textit{Solution Bias} corrupts business metrics like First Call Resolution. \textit{\textbf{Conversational Structure \& Flow}} maintains narrative integrity, as \textit{Temporal Sequence} bias can alter cause-and-effect interpretations, and \textit{Position} Bias can omit crucial mid-conversation resolution steps. Finally, \textit{\textbf{Speaker \& Role Representation}} ensures fair attribution, with \textit{Speaker} Bias being critical for balanced performance evaluations.

The remaining two classes evaluate more nuanced aspects of the interaction that are vital for risk management and quality assurance. The \textit{\textbf{Linguistic \& Stylistic}} class addresses distortions in conversational tone; \textit{Politeness} Bias, for example, can conceal agent behavior vital for performance evaluation, while \textit{Disfluency} Bias can mask customer confusion. Similarly, \textit{\textbf{Affective \& Pragmatic Interpretation}} focuses on subtext and intent. \textit{Sentiment} and \textit{Emotion Shift} Bias can obscure significant customer dissatisfaction and churn risks, while \textit{Urgency} Bias addresses the failure to capture time-sensitive requests. 

The proposed taxonomy therefore provides a structured framework that connects summarization bias to specific operational requirements. Although not exhaustive, this approach offers a crucial tool for holistically assessing a summary's true operational value and guiding its improvement, moving beyond generic metrics. A detailed description of each dimension is provided in Appendix~\ref{appendix:taxonomy}.

\subsection{Problem Formulation}

\begin{figure*}[t]
    \centering
    \includegraphics[width=0.98\textwidth]{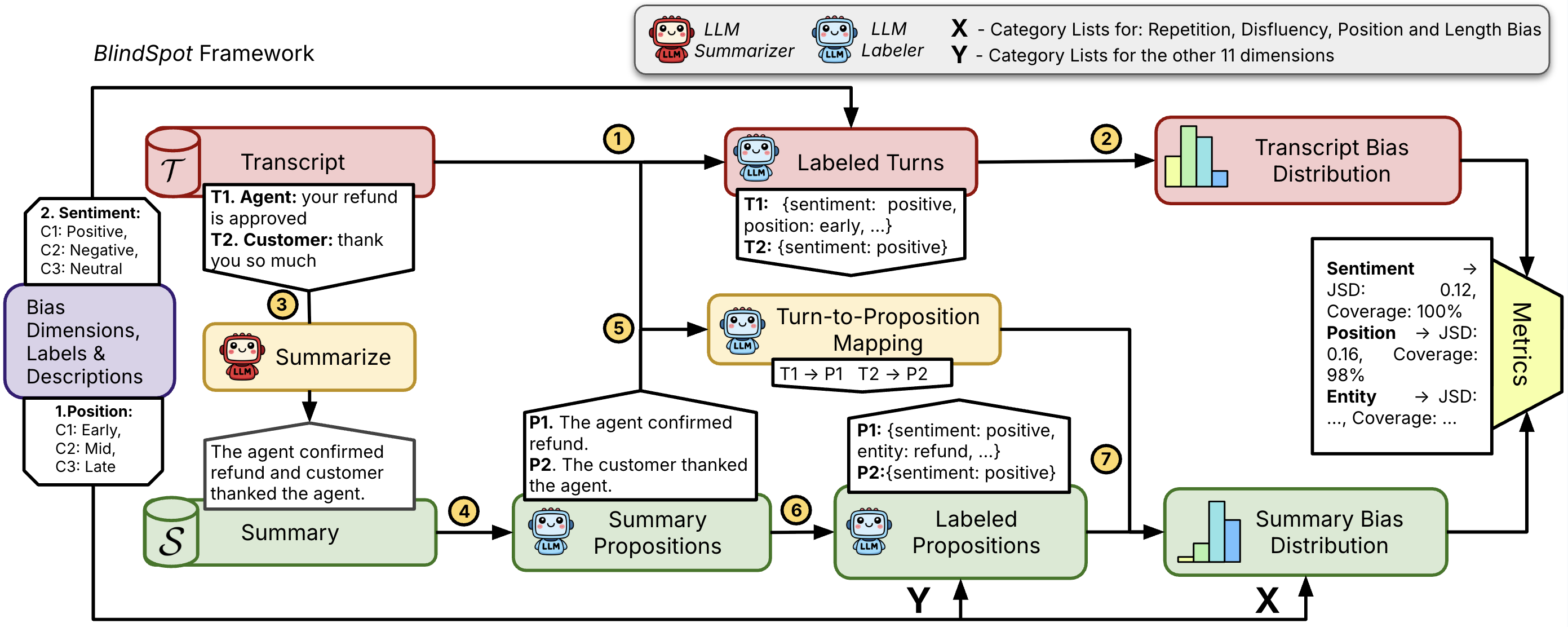}
    \caption{The \textit{BlindSpot} framework evaluates bias in call summaries. The transcript pipeline (red) creates a reference bias distribution by labeling turns for each dimension with an \textit{LLM Labeler}. The summary pipeline (green) creates a summary distribution by labeling propositions. Bias is quantified by comparing these distributions using \textit{Fidelity Gap} (JSD) and \textit{Coverage} \%. White boxes provide examples. (Best viewed in color.)}
    \label{fig:experiments}
\end{figure*}
Let \(\mathcal{T}=\{T_1,\dots,T_N\}\) be a corpus of \(N\) contact‐center transcripts.  Each transcript \(T_i\) consists of \(n_i\) turns, where a turn is a continuous utterance from a single speaker. An LLM summarizer \(\mathcal{M}\) produces a summary \(S_i\) composed of \(m_i\) \emph{propositions}—atomic units of information, typically a single claim or clause:
\(
S_i = \mathcal{M}(T_i)=(s_{i,1},\,s_{i,2},\,\dots,\,s_{i,m_i})\,.
\)

We define 15 bias dimensions \(d\), each associated with a discrete set of labels
\(\mathcal{C}_d=\{c_{d,1},\dots,c_{d,k}\}\).   For any unit $u$, a turn or proposition, a multi-label classifier LLM $\mathcal{L}_d$ assigns a subset of these labels:
\[
\mathcal{L}_d(u)\;\subseteq\;\mathcal{C}_d
\quad
\forall\,u\in T_i\cup S_i\,.
\]
For each transcript \(T_i\) and dimension \(d\), we compute the label distribution
\[
P_{i,d}(c)
=\frac{1}{n_i}\Bigl|\{\,t_{i,j}\in T_i: c\in \mathcal{L}_d(t_{i,j})\}\Bigr|,
\quad
c\in\mathcal{C}_d.
\]
Likewise for the summary \( S_i\):
\[
Q_{i,d}(c)
=\frac{1}{m_i}\Bigl|\{\,s_{i,j}\in S_i: c\in \mathcal{L}_d(s_{i,j})\}\Bigr|.
\]
We measure \emph{fidelity gap} in distributions in dimension \(d\) for pair \((T_i, S_i)\) via Jensen–Shannon divergence (JSD)~\citep{MENENDEZ1997307}:
\(
\mathrm{FidelityGap}_{i,d}
=
D_{\mathrm{JS}}(P_{i,d}\,\|\,Q_{i,d}).
\)
 The overall \textit{fidelity gap} in \(d\) is
\[
\mathrm{FidelityGap}_d
=\frac{1}{N}\sum_{i=1}^N \mathrm{FidelityGap}_{i,d}.
\]
To detect outright omissions, we also define \emph{coverage} for dimension \(d\):

\begin{subequations}
\begin{align*}
\mathrm{Coverage}_{i,d}
&= \frac{\#\{\,c: P_{i,d}(c)>0,\;Q_{i,d}(c)>0\}}
        {\#\{\,c: P_{i,d}(c)>0\}}\,,\\
\mathrm{Coverage}_{d} \%
&= \frac{1}{N}\sum_{i=1}^{N}\mathrm{Coverage}_{i,d}\, \times 100.
\end{align*}
\end{subequations}

Thus, for each bias dimension \(d\), two complementary metrics—\(\mathrm{FidelityGap}_d\) and \(\mathrm{Coverage}_d\) \% —jointly quantify how summaries distort or omit labels relative to the original transcripts.

\subsection{Framework Design and Workflow}
The \textit{BlindSpot} framework quantifies operational bias in three stages: generating the reference distribution from transcript, deriving the summary distribution, and computing bias scores from their comparison.
\paragraph{Transcript Pipeline:}
To establish a ground-truth representation, we first generate a categorical distribution $P_d$ for each bias dimension from the source transcript. Turn-level labels are produced using a hybrid approach. For dimensions requiring semantic interpretation (e.g., \textit{Sentiment, Topic, Politeness, Entity Type}), we leverage an LLM Labeler $\mathcal{L}$ to identify labels. For structural dimensions, we use direct computation: \textit{Speaker} is extracted from metadata, while \textit{Turn Length} and \textit{Position} are calculated from turn and its index. Finally, derived dimensions like \textit{Emotion Shift} and \textit{Temporal Sequence} are inferred from the labels of \textit{Sentiment} and \textit{Position}. The LLM Labeler, GPT-4o, was validated against a human-annotated set, achieving 93.7\% accuracy (see Appendix~\ref{subsec:llm_validation} for validation details).
\paragraph{Summary Pipeline:}
Next, we generate a distribution $Q_d$ from the summary produced by the LLM under evaluation. First, the model generates a complete summary from the full transcript, in single forward pass, mirroring real-world application. To enable fine-grained analysis, this summary is then decomposed into minimal semantic units, or "propositions," using an LLM. This step ensures a uniform and granular basis for labeling. Each proposition is then annotated using the same hybrid methodology as the transcript turns. To handle turn-dependent dimensions (e.g., \textit{Position}, \textit{Disfluency}), we perform a mapping step, linking each proposition back to the one or more source turns it summarizes.
\paragraph{Bias Quantification.}
Finally, we quantify bias for each dimension by calculating the \textit{Fidelity Gap} and \textit{Coverage} between the transcript distribution $P_d$ and the summary distribution $Q_d$. For derived dimensions like \textit{Temporal Sequence}, the reference distribution  is defined as a one-hot vector representing the ideal label. Consequently, only \textit{Fidelity Gap} is computed, as \textit{Coverage} is not applicable.

Full workflow and implementation details are in Appendix~\ref{appendix:methodology}, and labeler prompts are in Appendix~\ref{appendix:prompts}.

\section{Experimental Setup}

\begin{table*}[!hbt]
  \centering
  \small
  \renewcommand{\arraystretch}{1.2}
  \resizebox{\textwidth}{!}{%
    \begin{tabular}{@{}l
      !{\color{softline}\vrule width .5pt}           
    *{20}{>{\centering\arraybackslash}p{1.2cm}}
    !{\color{softline}\vrule width .5pt}           
    >{\centering\arraybackslash}p{1.2cm}%
  @{}}

      \textbf{Metric / Bias} 
        & \rotatebox{90}{\texttt{llama-3.2-1b}} 
        & \rotatebox{90}{\texttt{llama-3.2-3b}} 
        & \rotatebox{90}{\texttt{llama-3.3-70b}} 
        & \rotatebox{90}{\texttt{llama-4-maverick}} 
        & \rotatebox{90}{\texttt{nova-micro}} 
        & \rotatebox{90}{\texttt{nova-lite}} 
        & \rotatebox{90}{\texttt{nova-pro}} 
        & \rotatebox{90}{\texttt{claude-3.5-haiku}} 
        & \rotatebox{90}{\texttt{claude-3.7-sonnet}} 
        & \rotatebox{90}{\texttt{claude-4-sonnet}} 
        & \rotatebox{90}{\texttt{deepseek-r1}} 
        & \rotatebox{90}{\texttt{gemini-2.0-flash}} 
        & \rotatebox{90}{\texttt{gemini-2.0-flash-lite}} 
        & \rotatebox{90}{\texttt{gpt-4o-mini}} 
        & \rotatebox{90}{\texttt{gpt-4o}} 
        & \rotatebox{90}{\texttt{gpt-4.1-nano}} 
        & \rotatebox{90}{\texttt{gpt-4.1-mini}} 
        & \rotatebox{90}{\texttt{gpt-4.1}} 
        & \rotatebox{90}{\texttt{o3-mini}} 
        & \rotatebox{90}{\texttt{o4-mini}}
        & \rotatebox{90}
        {\textbf{Average}}\\
      \midrule
      \multicolumn{21}{@{}l}{\textit{\textbf{Fidelity Gap (JSD)}} ($\downarrow$ better)} \\
      Turn Length         & 0.016 & 0.014 & \textcolor{maxcolor}{\textbf{0.013}} & 0.015 & 0.015 & 0.014 & 0.015 & 0.014 & 0.015 & 0.014 & 0.015 & \textcolor{mincolor}{\textbf{0.048}} & \textcolor{mincolor}{\textbf{0.048}} & \textcolor{maxcolor}{\textbf{0.013}} & 0.014 & \textcolor{maxcolor}{\textbf{0.013}} & \textcolor{maxcolor}{\textbf{0.013}} & \textcolor{maxcolor}{\textbf{0.013}} & \textcolor{maxcolor}{\textbf{0.013}} & 0.015 & 0.017 \\
      Speaker        & 0.016 & 0.016 & 0.014 & 0.014 & 0.018 & 0.016 & 0.016 & 0.013 & 0.012 & \textcolor{maxcolor}{\textbf{0.011}} & 0.014 & \textcolor{mincolor}{\textbf{0.048}} & \textcolor{mincolor}{\textbf{0.048}} & 0.012 & 0.014 & 0.015 & 0.013 & 0.013 & 0.015 & 0.014 & 0.018 \\
      Position       & 0.026 & 0.019 & 0.016 & 0.017 & 0.017 & 0.016 & 0.019 & 0.016 & 0.017 & 0.017 & 0.017 & \textcolor{mincolor}{\textbf{0.077}} & 0.076 & 0.017 & 0.017 & \textcolor{maxcolor}{\textbf{0.014}} & 0.015 & 0.016 & 0.015 & 0.017 & 0.023 \\
      Urgency        & 0.025 & 0.023 & 0.023 & 0.023 & 0.024 & 0.023 & 0.024 & 0.025 & 0.025 & 0.027 & 0.026 & \textcolor{mincolor}{\textbf{0.049}} & 0.045 & \textcolor{maxcolor}{\textbf{0.022}}  & 0.024 & \textcolor{maxcolor}{\textbf{0.022}}  & 0.023 & 0.024 & \textcolor{maxcolor}{\textbf{0.022}} & 0.024 & 0.026 \\
      Solution       & 0.046 & 0.030 & 0.029 & 0.029 & 0.028 & 0.027 & 0.032 & 0.031 & 0.035 & 0.035 & 0.032 & \textcolor{mincolor}{\textbf{0.073}} & 0.068 & 0.028 & 0.027 & \textcolor{maxcolor}{\textbf{0.023}} & 0.027 & 0.025 & 0.024 & 0.027 & 0.034 \\
      Politeness     & 0.036 & 0.038 & 0.035 & 0.035 & 0.038 & 0.037 & 0.037 & 0.033 & 0.032 & 0.031 & 0.035 & \textcolor{mincolor}{\textbf{0.066}} & 0.063 & 0.034 & 0.035 & 0.031 & 0.033 & 0.033 & \textcolor{maxcolor}{\textbf{0.031}} & 0.035 & 0.037 \\
      Language Complexity      & 0.041 & 0.038 & 0.035 & 0.037 & 0.038 & 0.036 & 0.039 & 0.036 & 0.036 & 0.036 & 0.039 & 0.081 & \textcolor{mincolor}{\textbf{0.083}} & 0.034 & 0.035 & 0.034 & 0.035 & 0.035 & \textcolor{maxcolor}{\textbf{0.033}} & 0.038 & 0.041 \\
      Sentiment      & 0.041 & 0.041 & 0.038 & 0.040 & 0.039 & 0.040 & 0.040 & 0.043 & 0.046 & 0.048 & 0.046 & \textcolor{mincolor}{\textbf{0.069}} & 0.068 & 0.038 & 0.040 & \textcolor{maxcolor}{\textbf{0.036}} & 0.040 & 0.039 & \textcolor{maxcolor}{\textbf{0.036}} & 0.043 & 0.044 \\
      Disfluency     & 0.055 & 0.052 & 0.050 & 0.051 & 0.050 & 0.052 & 0.054 & 0.051 & 0.052 & 0.053 & 0.053 & \textcolor{mincolor}{\textbf{0.076}} & 0.075 & 0.049 & 0.051 & \textcolor{maxcolor}{\textbf{0.048}} & 0.051 & 0.051 & 0.049 & 0.054 & 0.054 \\
      Topic          & 0.058 & 0.050 & 0.047 & 0.048 & 0.052 & 0.050 & 0.054 & 0.054 & 0.057 & 0.058 & 0.057 & \textcolor{mincolor}{\textbf{0.128}} & 0.121 & 0.045 & 0.050 & 0.047 & 0.048 & 0.047 & \textcolor{maxcolor}{\textbf{0.046}} & 0.053 & 0.060 \\
      Information Repetition     & 0.091 & 0.093 & 0.084 & 0.080 & 0.086 & 0.090 & 0.087 & 0.084 & 0.089 & 0.086 & 0.087 & \textcolor{mincolor}{\textbf{0.100}} & \textcolor{mincolor}{\textbf{0.100}} & 0.078 & 0.089 & 0.078 & 0.079 & 0.082 & \textcolor{maxcolor}{\textbf{0.075}} & 0.085 & 0.087 \\
      Emotion Shift   & 0.116 & 0.144 & 0.138 & 0.129 & 0.140 & 0.137 & 0.132 & 0.131 & 0.116 & 0.119 & 0.128 & 0.119 & 0.112 & \textcolor{mincolor}{\textbf{0.149}} & 0.137 & 0.137 & 0.129 & 0.125 & 0.122 & \textcolor{maxcolor}{\textbf{0.107}} & 0.128 \\
      Entity Type        & 0.170 & 0.158 & 0.147 & 0.136 & 0.180 & 0.173 & 0.176 & 0.116 & 0.096 & \textcolor{maxcolor}{\textbf{0.086}} & 0.120 & 0.169 & \textcolor{mincolor}{\textbf{0.190}} & 0.181 & 0.169 & \textcolor{mincolor}{\textbf{0.190}} & 0.146 & 0.149 & 0.173 & 0.111 & 0.152 \\
      Agent Action   & 0.180 & 0.178 & \textcolor{maxcolor}{\textbf{0.174}} & 0.178 & 0.182 & 0.182 & 0.184 & 0.182 & 0.188 & 0.188 & 0.189 & \textcolor{mincolor}{\textbf{0.215}} & 0.213 & 0.175 & 0.178 & 0.176 & 0.180 & 0.178 & 0.177 & 0.185 & 0.184 \\
      Temporal Sequence          & 0.394 & 0.358 & 0.337 & 0.356 & 0.382 & 0.370 & 0.387 & 0.362 & 0.358 & 0.348 & 0.347 & \textcolor{mincolor}{\textbf{0.467}} & \textcolor{mincolor}{\textbf{0.467}} & 0.380 & 0.385 & 0.351 & \textcolor{maxcolor}{\textbf{0.326}} & 0.333 & 0.353 & 0.349 & 0.370 \\
      \midrule
    \rowcolor{avgcol} \textbf{Average} & 0.087 & 0.084 & 0.079 & 0.079 & 0.086 & 0.084 & 0.086 & 0.079 & 0.078 & \textcolor{maxcolor}{\textbf{0.077}} & 0.080 & \textcolor{mincolor}{\textbf{0.119}} & \textcolor{mincolor}{\textbf{0.119}} & 0.084 & 0.084 & 0.081 & \textcolor{maxcolor}{\textbf{0.077}} & 0.078 & 0.079 & \textcolor{maxcolor}{\textbf{0.077}} & \textbf{0.081} \\
    
      \midrule
      \multicolumn{21}{@{}l}{\textit{\textbf{Coverage}} (↑ better)} \\
      Turn Length         & 87.00 & 86.77 & 87.82 & 86.12 & 86.60 & \textcolor{maxcolor}{\textbf{87.83}} & 85.63 & 85.77 & 85.94 & 85.32 & 85.48 & \textcolor{mincolor}{\textbf{69.32}} & 71.16 & 87.65 & 87.00 & 87.67 & 87.16 & 87.38 & 87.81 & 85.72 & 85.01 \\
      Speaker        & \textcolor{maxcolor}{\textbf{99.16}} & 97.83 & 98.17 & 97.67 & 98.17 & 98.00 & 98.17 & 98.00 & 98.17 & 97.83 & 97.83 & 84.50 & \textcolor{mincolor}{\textbf{86.08}} & 98.33 & 98.50 & 97.83 & 98.00 & 97.83 & 98.00 & 98.17 & 96.81 \\
      Position       & \textcolor{maxcolor}{\textbf{98.79}} & 97.77 & 98.17 & 97.50 & 98.07 & 97.93 & 98.03 & 97.93 & 98.03 & 97.67 & 97.66 & \textcolor{mincolor}{\textbf{79.39}} & 80.57 & 98.23 & 98.40 & 97.80 & 98.00 & 97.77 & 97.97 & 98.03 & 96.18\\
      Urgency        & 92.09 & 91.93 & \textcolor{maxcolor}{\textbf{93.73}} & 91.86 & 92.57 & 92.17 & 92.21 & 92.26 & 92.61 & 91.21 & 91.60 & 74.53 & \textcolor{mincolor}{\textbf{73.78}} & 93.02 & 92.96 & 92.82 & 93.16 & 92.41 & 93.63 & 92.18 & 90.64 \\
      Solution       & 80.32 & 85.02 & 86.44 & 84.87 & 85.36 & 86.54 & 84.45 & 83.74 & 82.61 & 82.91 & 84.00 & \textcolor{mincolor}{\textbf{63.07}} & 65.50 & 85.42 & 86.11 & \textcolor{maxcolor}{\textbf{87.33}} & 86.42 & 85.99 & 86.96 & 85.28 & 82.92 \\
      Politeness     & 95.15 & 95.42 & 95.82 & 94.90 & 94.69 & 94.76 & 94.96 & 93.68 & 93.22 & 92.90 & 94.00 & \textcolor{mincolor}{\textbf{78.53}} & 79.88 & \textcolor{maxcolor}{\textbf{96.01}} & 95.46 & 95.11 & 95.13 & 95.12 & 94.78 & 93.61 & 93.16 \\
      Language Complexity      & 82.51 & 83.30 & 84.56 & 83.27 & 83.37 & 82.93 & 82.10 & 82.81 & 82.91 & 82.75 & 82.96 & \textcolor{mincolor}{\textbf{63.13}} & 65.01 & \textcolor{maxcolor}{\textbf{84.60}} & 83.92 & 84.46 & 83.10 & 83.72 & 84.40 & 83.19 & 81.45 \\  
      Sentiment      & 89.00 & 90.13 & 91.52 & 90.15 & 89.54 & 90.74 & 89.44 & 89.31 & 88.93 & 88.23 & 88.99 & \textcolor{mincolor}{\textbf{71.42}} & 72.89 & \textcolor{maxcolor}{\textbf{92.05}} & 90.72 & 91.13 & 91.17 & 90.25 & 90.16 & 88.70 & 88.22 \\
      Disfluency     & 67.91 & 68.20 & 70.23 & 68.16 & 69.37 & 68.64 & 67.40 & 69.17 & 68.35 & 67.96 & 67.96 & 51.44 & \textcolor{mincolor}{\textbf{52.93}} & 69.99 & 69.42 & 70.48 & 69.66 & 69.43 & \textcolor{maxcolor}{\textbf{70.65}} & 67.52 & 67.24 \\
      Topic          & 75.54 & 79.11 & 81.03 & 79.44 & 78.12 & 78.83 & 76.58 & 76.04 & 74.55 & 72.85 & 75.72 & \textcolor{mincolor}{\textbf{54.42}} & 56.96 & \textcolor{maxcolor}{\textbf{81.53}} & 79.59 & 80.37 & 79.12 & 79.62 & 78.96 & 75.20 & 75.68 \\
      Information Repetition     & 60.83 & 61.83 & 61.52 & 63.04 & 61.57 & 61.84 & 60.43 & 59.83 & 60.34 & 61.64 & 60.47 & 42.91 & \textcolor{mincolor}{\textbf{42.15}} & 63.61 & 61.23 & 65.60 & \textcolor{maxcolor}{\textbf{63.84}} & 61.84 & 65.85 & 62.49 & 60.14 \\
      Entity Type        & 50.66 & 52.03 & 54.04 & 56.52 & 47.02 & 48.82 & 48.73 & 60.34 & 67.39 & \textcolor{maxcolor}{\textbf{70.96}} & 60.00 & 34.32 & \textcolor{mincolor}{\textbf{32.37}} & 46.29 & 49.57 & 44.64 & 54.48 & 53.62 & 48.91 & 63.07 & 52.19 \\
      Agent Action   & 67.74 & 68.19 & \textcolor{maxcolor}{\textbf{70.62}} & 68.80 & 67.00 & 68.22 & 65.96 & 66.90 & 64.71 & 64.71 & 64.40 & \textcolor{mincolor}{\textbf{51.29}} & 53.41 & 70.12 & 68.69 & 70.59 & 68.81 & 68.77 & 69.66 & 65.92 & 66.23 \\
      \midrule
    \rowcolor{avgcol} \textbf{Average} & 80.52 & 81.35 & \textcolor{maxcolor}{\textbf{82.59}} & 81.72 & 80.88 & 81.25 & 80.31 & 81.21 & 81.37 & 81.30 & 80.85 & \textcolor{mincolor}{\textbf{62.94}} & 64.05 & 82.07 & 81.66 & 81.99 & 82.16 & 81.83 & 82.13 & 81.47 & \textbf{79.68} \\
    
      \midrule
    \textbf{LLM Judge Score} & \textcolor{mincolor}{\textbf{2.07}} & 4.04 & 4.79 & \textcolor{maxcolor}{\textbf{4.87}} & 4.68 & 4.61 & 4.85 & 4.83 & 4.72 & 4.81 & 4.71 & 3.87 & 3.96 & 4.71 & 4.85 & 4.72 & 4.78 & 4.78 & 4.74 & 4.79 & \textbf{4.64}\\
    
    \textbf{Compression Factor} & 
    \textbf{10.98} & 18.83 & 17.23 & 20.75 & 27.44 & 25.29 & 31.2 & 22.86 & 19.05 & 17.29 & 21.87 & \textbf{62.01} & 60.78 & 26.37 & 27.73 & 29.19 & 20.84 & 17.68 & 20.13 & 21.25 & \textbf{25.94} \\

      \bottomrule

    \end{tabular}%
      }
  \caption{Main evaluation results for $20$ LLMs on $15$ bias dimensions in call summarization. Reported metrics include: \textit{Fidelity Gap} (JSD) ($0$–$1$, $\downarrow$ better), \textit{Coverage} $\%$ ($0$–$100$, $\uparrow$ better), LLM Judge Score ($1$–$5$, $\uparrow$ better), and Compression Factor. We highlight the best scores in green and worst scores in red for each row.}
  \label{table:main_results}
\end{table*}
\paragraph{Dataset and Models}
We evaluate on 2500 real contact-center transcripts\footnote{The dataset cannot be released due to its proprietary nature.} from 12 domains (e.g., FinTech, Healthcare), summarized by 20 LLMs under uniform prompting (details in Appendix~\ref{appendix:experimental_setup}).

\paragraph{Evaluation Metrics}
Our evaluation pairs two metrics to quantify bias for each dimension. We use \textbf{Jensen-Shannon Divergence (JSD)} to measure the distributional shift, which serves as a robust and symmetric measure of the \textit{fidelity gap}. We also compute \textbf{Coverage} \%: the percentage of source labels that appear in the summary. To contextualize these findings, we also report \textbf{LLM-Judge score} (1–5 scale; see Appendix~\ref{subsec:llm_judge} for details) for overall summary quality and \textbf{Compression Factor} (transcript/summary tokens) to measure the degree of abstraction. Additional divergence metrics are in Appendix~\ref{subsec:alternate_metrics}.

\section{Results}
We evaluated 20 LLMs across 15 bias dimensions (Table \ref{table:main_results}) and highlight the key findings below.
\paragraph{Overall Model Performance}
The majority of evaluated models demonstrate similar performance, occupying a narrow range for both average JSD (0.077–0.087) and Coverage (80.31–82.59\%). However, our analysis reveals three key observations. First, model performance is not solely determined by scale; top performers include both large models like \texttt{claude-4-sonnet} and \texttt{llama-3.3-70b} and smaller ones like \texttt{gpt-4.1-mini}. Second, \texttt{gemini-2.0-flash} and \texttt{gemini-2.0-flash-lite} are notable outliers, exhibiting significantly higher average JSD (0.119). Finally, we observe modest but consistent improvements from intra-family scaling. In the Llama series, for example, JSD drops from 0.087 (1B) to 0.079 (70B) as Coverage increases by 2\%. This pattern holds for other model families.
\paragraph{Analysis by Bias Dimension}
The results reveal two clear groups of bias dimensions: \textbf{Most Challenging Dimensions:} The preservation of \textit{Temporal Sequence} presents the most significant challenge, with the highest average JSD (0.370) by a large margin. This indicates models frequently alter event chronology, obscuring cause-and-effect. Furthermore, dimensions requiring granular detail show the lowest information retention. \textit{Entity type} coverage is the lowest on average at 52.19\%, meaning nearly half of all named entities are typically omitted. Models also struggle with \textit{Repetition} (60.14\% coverage) and \textit{Agent Actions} (66.23\% coverage), suggesting a difficulty in capturing the significance of repeated points and agent activities. \textbf{Most Robust Dimensions:} In contrast, models are highly effective at preserving high-level structural information. The \textit{Speaker} and \textit{Position} dimensions show minimal bias, with very low average JSD (0.018 and 0.023) and high coverage (96.81\% and 96.18\%, respectively). This suggests that while models can reliably attribute statements and identify general location in conversation, they fail to preserve fine-grained details within those structural boundaries.
\paragraph{Influence of Compression on Bias}
Bias increases with compression: Pearson correlation shows that JSD increases ($r=0.76$) and coverage drops ($r=-0.88$) as compression increases. An exception is \texttt{llama-3.2-1b}, which has the lowest compression ($10.98$) but still a high bias.
\paragraph{Insufficiency of Quality Metrics}
Holistic metrics like LLM-Judge score weakly correlate with bias: Pearson coefficients show modest improvements in JSD ($r = -0.34$) and coverage ($r = 0.33$) as scores increase. However, high-scoring models like \texttt{nova-pro} (score = 4.85) can still exhibit severe \textit{Temporal Sequence} bias (JSD = 0.387), revealing that such metrics overlook structural fidelity.

\begin{figure}[H]
    \centering
    \includegraphics[width=0.81\linewidth]{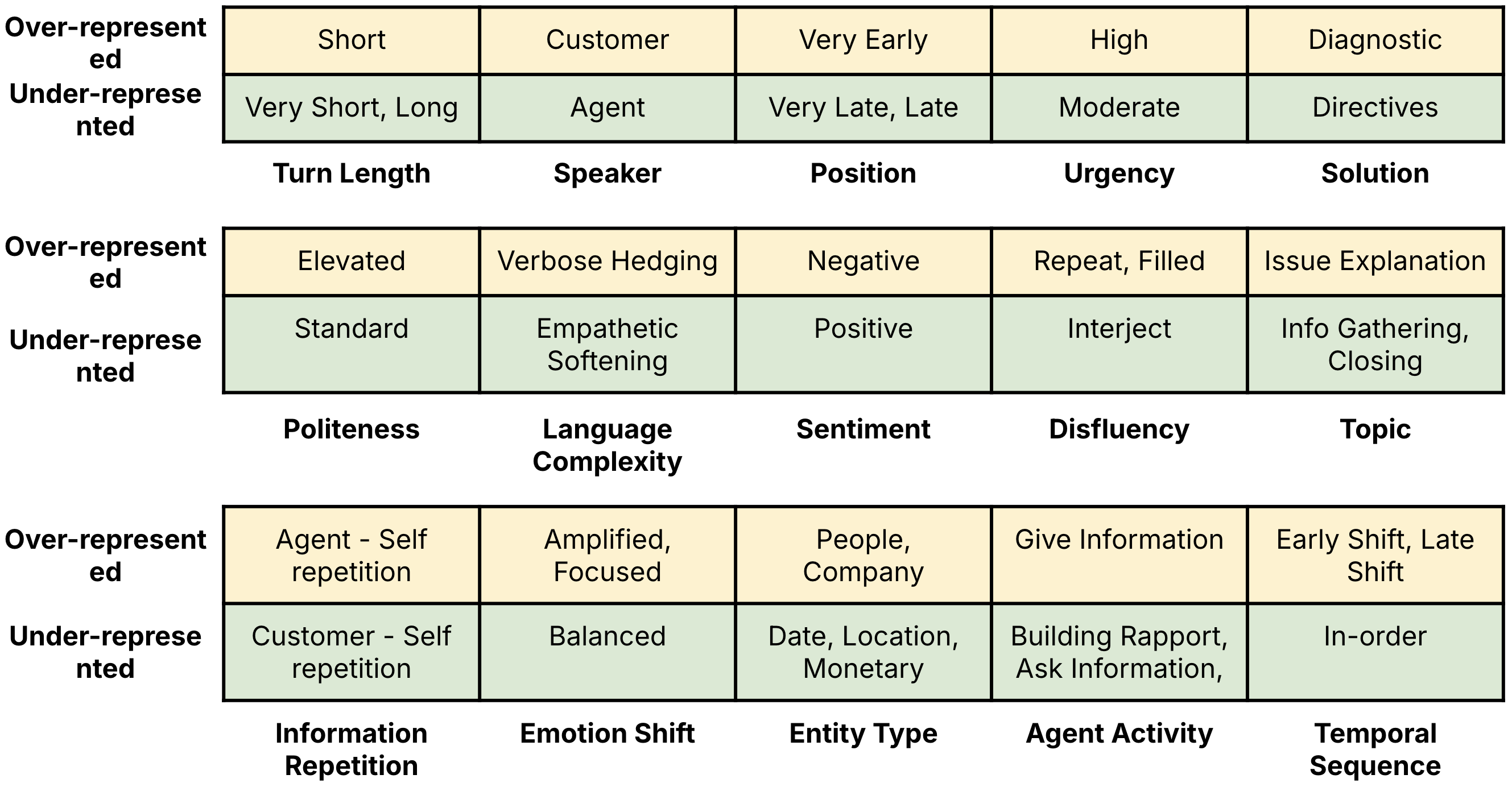}
    \caption{Specific labels that are over- or under-represented consistently across all models.}
    \label{figure:over-under}
\end{figure}

\paragraph{Analysis of Representation Patterns}
Our fine-grained analysis reveals systematic biases (Figure \ref{figure:over-under}). Models consistently over-represent labels like Negative sentiment and Early segments, while under-representing labels like Building-Rapport and Directives. This indicates a model tendency to construct simplified, problem-focused narratives, sacrificing crucial interactional context.

\begin{figure}
    \centering
    \includegraphics[width=0.78\linewidth]{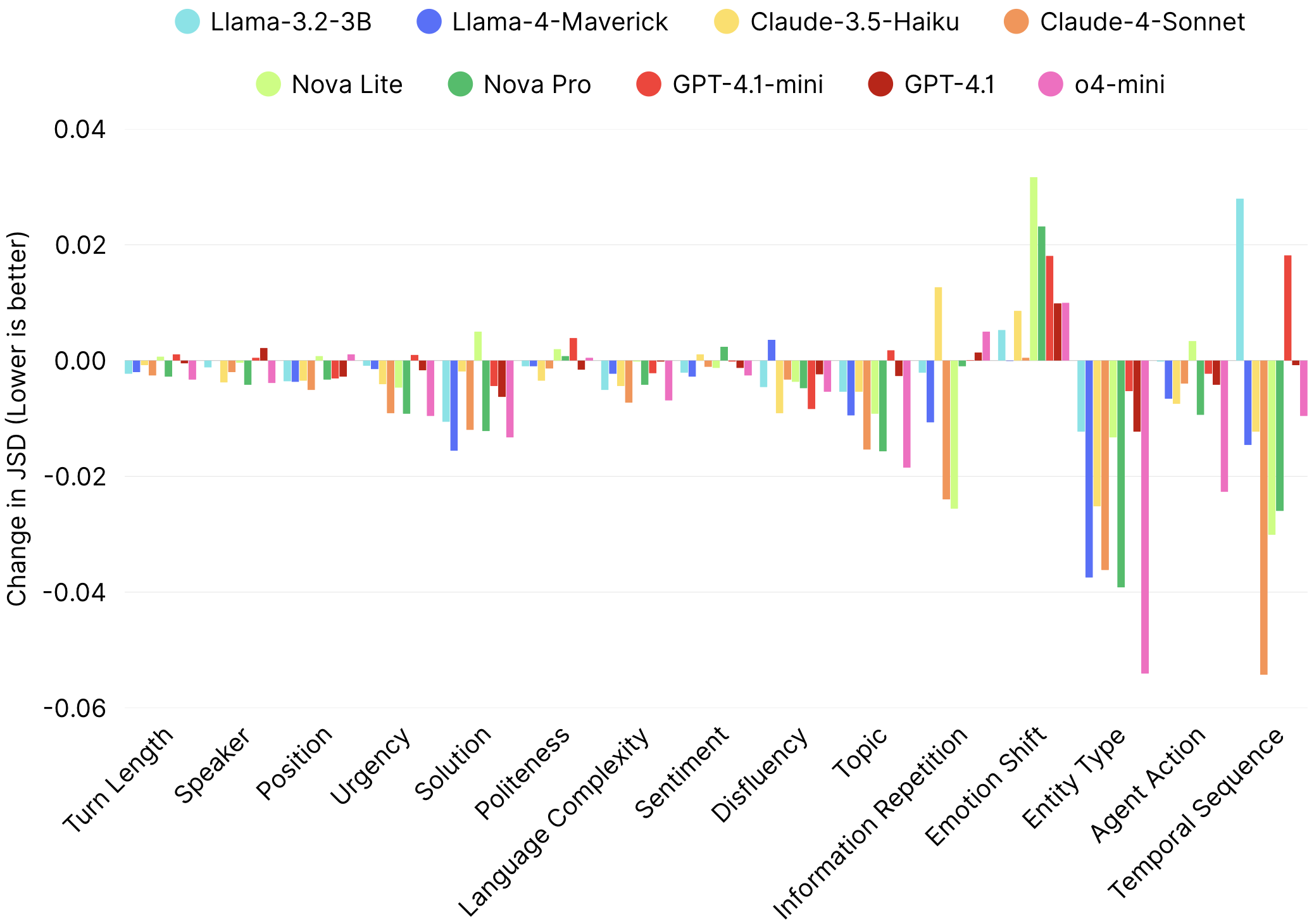}
    \includegraphics[width=0.78\linewidth]{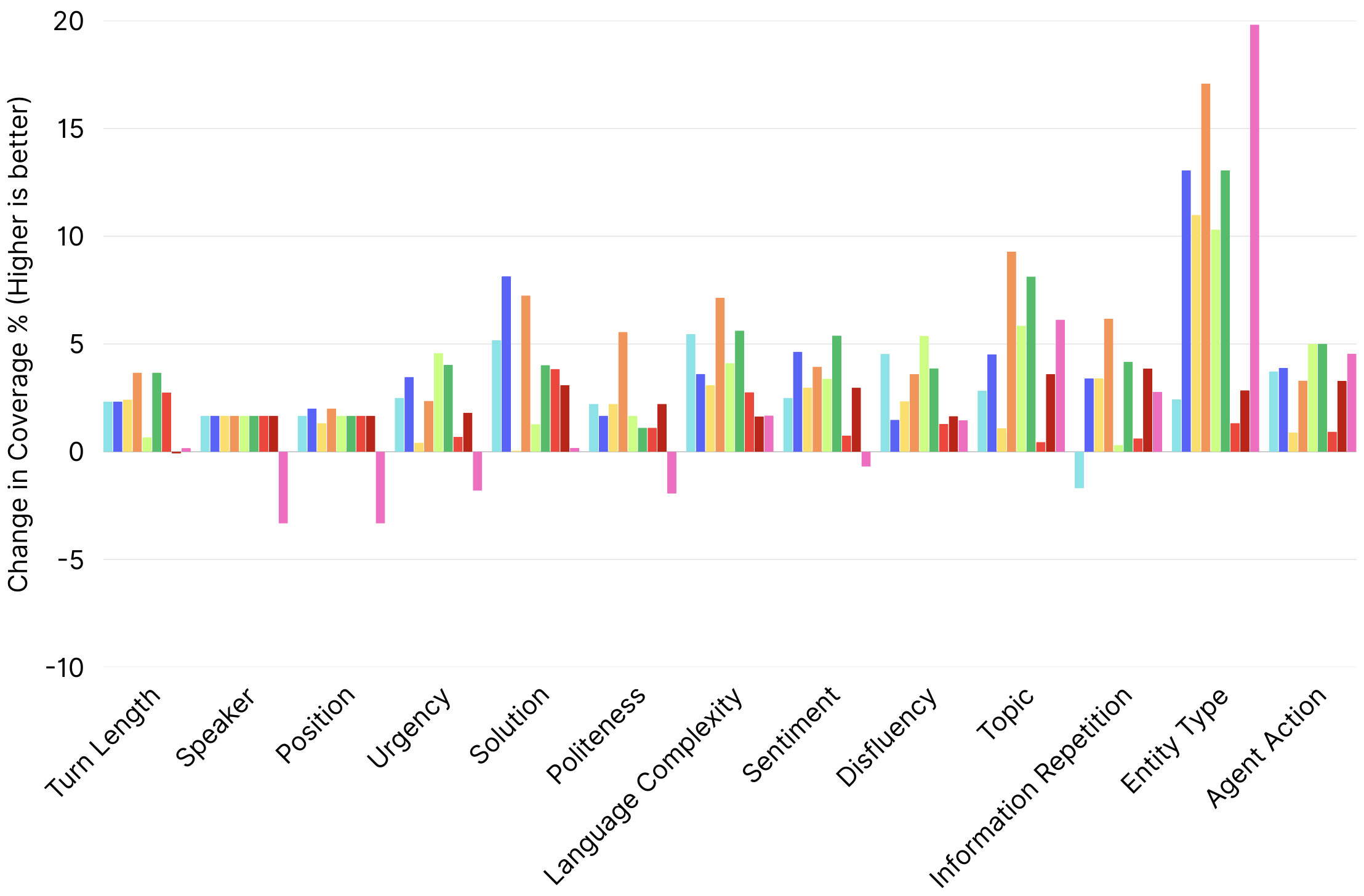}
    \caption{Change in JSD (top) and Coverage \% (bottom) after bias mitigation.}
    \label{fig:mitigation_results}
\end{figure}

\paragraph{Bias Mitigation via Targeted Prompting}
To demonstrate our framework’s utility, we investigated bias mitigation by constructing a system prompt (see ~\ref{box:bias-mitigation-prompt}) based on our analysis. We evaluated this prompt on nine models: a small and large variant from four families, plus a reasoning model. As shown in Figure~\ref{fig:mitigation_results}, all models reduced bias, with lower JSD (except \textit{Emotion Shift}) and higher Coverage. This led to substantial gains for \texttt{sonnet}, with a +4.87\% Coverage increase and 0.012 JSD reduction. \texttt{llama-4} and \texttt{nova-pro} also improved Coverage by 3.59\% and 4.09\%. {\textbf{Notably, we observe a scaling effect: larger models consistently showed greater bias reduction than smaller ones in the same family.} For instance, \texttt{sonnet} showed more JSD reduction (-0.012 vs. -0.004 for \texttt{haiku}), while \texttt{llama-4} achieved a higher Coverage gain (+3.59\% vs. +2.36\% for \texttt{llama-3b}). While full mitigation is beyond this paper’s scope, this experiment shows that \textit{BlindSpot} provides actionable feedback to improve model behavior.

\section{Conclusion}
\label{sec:conclusion}
This work demonstrates that while LLMs produce fluent summaries of contact center conversations, they contain systematic operational biases. To address this, we introduce BlindSpot, a framework that quantifies these distortions across 15 contact center specific dimensions using divergence and coverage metrics. We show that the detailed analysis from BlindSpot is actionable; its findings enabled us to construct a targeted prompt that measurably reduces bias. This research provides a crucial toolset for building more transparent, trustworthy, and domain-aware summarization systems.

\section{Limitations}
While our framework systematically detects biases in LLM-generated summaries, it does not evaluate the harmfulness, user impact, or real-world consequences of these biases. The current metrics, Jensen-Shannon Divergence and Coverage, quantify distributional misalignments but do not capture how these biases affect user trust, business decisions, or fairness in downstream applications.

Our analysis is constrained to English-language contact center transcripts. Consequently, the framework's applicability to multilingual contexts remains untested.

Finally, while the use of LLMs as zero-shot LLM labeler enables scalability, it introduces potential propagation of existing model biases, especially for subjective dimensions like \textit{politeness}, into the annotations themselves, a limitation inherent in LLM-based evaluation pipelines.

\section{Ethics Statement}

This work focuses on identifying and quantifying biases in LLM-generated summaries of contact center transcripts. Our dataset consists of anonymized, real-world transcripts that do not contain personally identifiable information. All experiments were conducted using publicly available LLMs and datasets under appropriate usage terms.

Our goal is to improve transparency and accountability in language model behavior, not to assign blame to any specific model or provider. However, we acknowledge that exposing model biases, especially across dimensions like sentiment, speaker prominence, or topic selection—may influence deployment decisions and perceptions of fairness. We urge practitioners to interpret our findings within the methodological scope of this study and avoid overgeneralizing results beyond contact center summarization.

No human annotators were employed for labeling tasks; all labels were produced by LLMs, with validation on a small human-rated subset. There was no involvement of vulnerable populations. We believe our findings contribute positively to the responsible development and evaluation of language technologies.

\bibliography{custom}

\clearpage

\appendix

\section{Taxonomy of Bias Dimensions}
\label{appendix:taxonomy}

This appendix provides a comprehensive description of the 15 dimensions of bias evaluated in our study. The taxonomy is organized into five classes, each targeting a distinct aspect of summary fidelity. For each dimension, we provide its rationale, a description of its labels, and its operational significance.

\subsection{Rationale for Bias Classes}
\label{subsec:class_rationale}

The five classes provide a structured approach to understanding different facets of summary quality and potential bias.

\paragraph{1. Class: Content \& Information Fidelity}
\textbf{Core Purpose:} To ensure the summary is a factually accurate and actionable record of the conversation's substance. Biases in this class directly compromise the summary's primary function as a reliable source of truth. \\
\textbf{Dimensions within this Class:}
\begin{itemize}
    \item \textbf{Entity Type:} This dimension tracks the presence of key named entities. Its operational importance is paramount; the omission of a single key identifier such as a case number, product ID, or callback number, can render a summary useless for follow-up actions and break continuity in the customer journey.
    \item \textbf{Topic:} This dimension ensures the summary reflects the primary purpose and subject matter of the call. A summary with topic bias might over-represent a brief mention of a billing issue in a call that was primarily about technical support, leading to mis-categorization and flawed business intelligence.
    \item \textbf{Solution:} This dimension is crucial for accurately tracking resolution success and agent effectiveness. Misrepresenting a partial fix as a full resolution directly inflates metrics like First Call Resolution (FCR). Furthermore, providing flawed data about which solutions work (or don't work) undermines product and service improvement efforts.
    \item \textbf{Information Repetition:} This dimension captures the nuanced handling of repeated statements. Repetition in a dialogue is not redundant; it is a rich signal often lost in summarization. We identify several key patterns:
        \begin{itemize}
        \item Customer Self-Repetition: A customer repeating their issue multiple times is a strong indicator of rising frustration, a feeling of not being heard, or confusion about the agent's response.
        \item Agent Repeating Customer: An agent paraphrasing or repeating a customer's statement is a standard technique for active listening and confirming understanding. Capturing this is vital for evaluating agent soft skills.
        \item Customer Repeating Agent: A customer repeating an agent's instructions or confirmation number indicates their attempt to verify information, which is a critical part of the interaction.
        \item Agent Self-Repetition: An agent repeating a compliance script or a key piece of information is often a matter of procedural record and must be documented.
        \end{itemize}
    A summary that simply collapses these repeated instances into a single mention loses this critical interactional context. Furthermore, due to ASR (Automatic Speech Recognition) errors, repeated content can sometimes appear contradictory or slightly different in the transcript. How a model handles these near-duplicates, whether it omits them, averages them, or correctly identifies the most likely intent is a key test of its robustness.
        
\end{itemize}

\paragraph{2. Class: Conversational Structure \& Flow}
\textbf{Core Purpose:} To assess the summary's narrative integrity, ensuring the chronological and causal sequence of events is preserved. The ``story'' of the call is often as important as its individual facts. \\
\textbf{Dimensions within this Class:}
\begin{itemize}
    \item \textbf{Position:} This dimension addresses the well-known ``lead bias,'' where models favor information from the beginning of a text. In a contact center context, this is operationally dangerous because crucial resolution steps, escalation decisions, and final confirmations are typically found in the middle and late stages of a conversation and are thus prone to omission.
    \item \textbf{Turn Length} This dimension measures how summary fidelity varies based on the length and complexity of individual turns. Conversations are composed of a mix of utterance types: short, functional turns (e.g., ``Yes,'' ``Okay,'' a case number) and long, narrative turns (e.g., a customer explaining their entire problem history). A key challenge for summarization is to correctly weigh the importance of these different turn types. A model might over-represent short, declarative turns while failing to extract the crucial details embedded within a single long, information-dense monologue. This dimension, therefore, measures the model's robustness in handling turns of varying complexity and its ability to avoid being biased towards either terse or verbose utterances.
    \item \textbf{Temporal Sequence:} This dimension measures whether the chronology of key events is preserved. A summary that misorders events, for example, by placing a customer's expression of frustration \textit{after} a proposed solution, fundamentally breaks the cause-and-effect narrative and can lead to unfair assessments of agent performance.
\end{itemize}

\paragraph{3. Class: Speaker \& Role Representation}
\textbf{Core Purpose:} To focus on the fair and accurate attribution of utterances and actions to the conversational participants. This is essential for accountability and performance evaluation. \\
\textbf{Dimensions within this Class:}
\begin{itemize}
    \item \textbf{Speaker:} This dimension reflects the balance in prominence between the customer and agent voices. A summary with speaker bias might over-represent the agent's turns, making them seem domineering, or under-represent them, making them appear passive. Both scenarios lead to a distorted picture of the interaction.
    \item \textbf{Agent Action:} This dimension tracks whether key agent behaviors are captured. QA scorecards are built around discrete agent actions like \textit{questioning}, \textit{informing}, \textit{empathizing}, and \textit{building rapport}. A summary that omits these actions provides an incomplete record for performance assessment and coaching. (Note: Customer activity is not separately modeled, as customer turns are typically reactive and lack the standardized operational roles of an agent).
\end{itemize}

\paragraph{4. Class: Linguistic \& Stylistic Dimensions}
\textbf{Core Purpose:} To target distortions in the \textit{manner} and \textit{tone} of the conversation. These stylistic features carry significant diagnostic information about the customer experience and agent professionalism that is lost if a summary only reports literal content. \\
\textbf{Dimensions within this Class:}
\begin{itemize}
    \item \textbf{Language Complexity:} This dimension addresses the simplification or complication of language. A summary that over-simplifies technical language may fail to document an agent's expertise. Conversely, a summary that fails to capture the simplicity of an agent's explanation may miss an example of excellent customer communication.
    \item \textbf{Disfluency:} This dimension tracks the presence of hesitations, false starts, and repetitions. While often considered ``noise,'' disfluencies are a rich source of information. Removing a customer's hesitations can erase crucial evidence of their uncertainty or confusion, misrepresenting the true customer experience and an agent's effectiveness in providing clarity.
    \item \textbf{Politeness:} This dimension measures the representation of social niceties. An agent's demeanor is a core metric for QA. A summary that ``sanitizes'' a rude interaction or makes a professional agent seem curt eliminates vital data for performance reviews and coaching.
\end{itemize}

\paragraph{5. Class: Affective \& Pragmatic Interpretation}
\textbf{Core Purpose:} To address the emotional and intentional subtext of the conversation, which is often more critical for business outcomes than the raw facts.\\
\textbf{Dimensions within this Class:}
\begin{itemize}
    \item \textbf{Sentiment:} This dimension captures the emotional valence of the interaction. Its importance for risk management cannot be overstated. A summary that minimizes genuine customer frustration by labeling it as neutral ``unhappiness'' or ``dissatisfaction'' can cause a high-priority churn risk to be overlooked by downstream systems and human reviewers.
    \item \textbf{Emotion Shift:} This dimension identifies more nuanced changes in emotional representation, such as amplification (making a neutral comment sound negative) or attenuation (weakening a strong emotion). These shifts affect the perceived severity of an issue and can lead to misprioritization in customer retention workflows.
    \item \textbf{Urgency:} This dimension measures the representation of time-sensitivity. Failing to flag a high-urgency request---such as ``I need to cancel this fraudulent transaction \textit{right now}!''---represents a direct and immediate failure in customer service with potentially significant financial and reputational consequences.
\end{itemize}

\subsection{Detailed Descriptions of Bias Dimensions and their Labels}

The following table \ref{tab:bias_dimensions_full} provides a complete list of the 15 bias dimensions, their corresponding labels used for classification, and a brief description. The source of the annotation (LLM-annotated, computed, or derived) is also indicated. Dimensions marked with (Multiselect) allow for the assignment of multiple labels per turn or proposition.

\begin{table*}[!hbt]
\centering
\small
\setlength{\tabcolsep}{5pt}
\renewcommand{\arraystretch}{1.2}
\begin{tabular}{p{0.22\linewidth} p{0.38\linewidth} p{0.34\linewidth}}
\toprule
\textbf{Bias Dimension} & \textbf{Labels} & \textbf{Operational Significance}  \\
\midrule
\multicolumn{3}{l}{\textit{\textbf{1. Content \& Information Fidelity}}} \\
\textbf{Entity Type} & People, Identifiers, Phone Number, Email, Time Info, Date, Location Info, Products/Services, Monetary, Company/Organization, Other & Over/underrepresentation of key factual data required for action. \\
\textbf{Topic} & Greeting/Introductions, Identity Verification, Issue, Information Gathering, Product/Service Inquiry, Diagnosis/Troubleshooting, Solution, Action, Transaction, Offers/Upgrades, Sales, Resolution Confirmation, Next Steps, Closure, Empathy, Complaint, Policy Explanation, Feedback, Scheduling, Billing, Compliance, Miscellaneous & Over-focus or neglect of certain topical segments, skewing the perceived purpose of the call. \\
\textbf{Solution} & Diagnosis, Advisory, Root Cause, Directive/Command, Preventive Measure, Escalate, Self-Help, Partial Fix, Rejected Fix, Follow-up, Set Expectation, Reassure, No Solution & Omission or distortion of resolutions, impacting FCR and product insights. \\
\textbf{Information Repetition} & No Repetition, Customer Self-Repetition, Agent Self-Repetition, Customer Repeats Agent, Agent Repeats Customer & Loss of context regarding participant frustration or confirmation loops. \\
\midrule
\multicolumn{3}{l}{\textit{\textbf{2. Conversational Structure \& Flow}}} \\
\textbf{Position} (computed) & Very Early, Early, Mid, Late, Very Late & Preference for information from specific segments of the conversation. \\
\textbf{Turn Length} (computed) & Very Short, Short, Mid, Long, Very Long & Variation in summary fidelity across dialogues of different length. \\
\textbf{Temporal Sequence} (derived) & In-order, Early-shift, Late-shift, Omitted, Added & Distortion of the chronological order of events, breaking causal chains. \\
\midrule
\multicolumn{3}{l}{\textit{\textbf{3. Speaker \& Role Representation}}} \\
\textbf{Speaker} (computed) & Agent, Customer & Unequal representation of agent vs. customer voice and contribution. \\
\textbf{Agent Action} & Request Information, Provide Information, Confirm Understanding, Build Rapport, Acknowledge, Escalate, Compliance, Other & Misrepresentation of agent actions, impacting performance evaluation. \\
\midrule
\multicolumn{3}{l}{\textit{\textbf{4. Linguistic \& Stylistic Dimensions}}} \\
\textbf{Language Complexity} (Multiselect) & Simple/Clear, Declarative, Long/Multi-Clause, Technical, Jargon, Abbreviations, Dense, Wordy/Vague, Formal, Informal, Empathic, Blunt, Slang, Passive Voice & Disproportionate simplification or complication of the original language style. \\
\textbf{Disfluency} (Multiselect) & Filled Pause, Repetition, False Start, Repair, Prolongation, Stutter, Discourse Marker, Interjection, Cutoff & Selective omission of speech imperfections that signal user confusion. \\
\textbf{Politeness} & Impolite, Standard, Minimal, Elevated & Neutralization or exaggeration of politeness, masking agent/customer demeanor. \\
\midrule
\multicolumn{3}{l}{\textit{\textbf{5. Affective \& Pragmatic Interpretation}}} \\
\textbf{Sentiment} & Very Positive, Positive, Neutral, Negative, Very Negative & Divergence in emotional tone, masking customer satisfaction or churn risk. \\
\textbf{Emotion Shift} (derived) & Balanced, Amplified, Attenuated, Inverted, Spurious & How the summary distorts, drops, or fabricates emotional nuance. \\
\textbf{Urgency} & None, Low, Moderate, High, Critical & Failure to represent time-sensitive requests, leading to service failures. \\
\bottomrule
\end{tabular}
\caption{The full taxonomy of 15 bias dimensions, organized by class. For each dimension, we provide its corresponding labels and operational significance. Labeling sources are noted in parentheses.}
\label{tab:bias_dimensions_full}
\end{table*}

\begin{table*}[ht]
    \centering
    \resizebox{2\columnwidth}{!}{%
    \begin{tabular}{|l|l|p{8cm}|}
    \hline
    \textbf{Code} & \textbf{Label} & \textbf{Description} \\ 
    \hline
    very\_early  & Very Early  & Tokens in the first 20\% of the transcript. \\ 
    \hline
    early        & Early       & Tokens in the next 20\% (20\%–40\%). \\ 
    \hline
    mid          & Mid         & Tokens in the middle segment (40\%–60\%). \\ 
    \hline
    late         & Late        & Tokens in the following 20\% (60\%–80\%). \\ 
    \hline
    very\_late   & Very Late   & Tokens in the final 20\% of the transcript (80\%–100\%). \\ 
    \hline
    \end{tabular}
    }
    \caption{The label set for the Position bias dimension, including descriptions and short codes used for labeling.}
    \label{tab:position_bias}
\end{table*}

\begin{table*}[ht]
    \centering
    \begin{tabular}{|l|l|p{8cm}|}
    \hline
    \textbf{Code}   & \textbf{Label} & \textbf{Description} \\ 
    \hline
    agent           & Agent            & Utterances spoken by the service agent. \\ 
    \hline
    customer        & Customer         & Utterances spoken by the customer. \\ 
    \hline
    \end{tabular}
    \caption{The label set for the Speaker bias dimension, including descriptions and short codes used for labeling.}
    \label{tab:speaker_bias}
\end{table*}

\begin{table*}[ht]
    \centering
    \begin{tabular}{|l|l|p{8cm}|}
    \hline
    \textbf{Code} & \textbf{Label} & \textbf{Description} \\ 
    \hline
    people        & People         & Named individuals. \\ 
    \hline
    identifiers   & Identifiers    & IDs like account numbers. \\ 
    \hline
    phone\_number         & Phone Number   & Telephone numbers. \\ 
    \hline
    email         & Email          & Email addresses. \\ 
    \hline
    time\_info          & Time Info      & Time-related entities (e.g., 3 PM). \\ 
    \hline
    date          & Date           & Dates and calendar references. \\ 
    \hline
    location\_info      & Location Info  & Geographical references. \\ 
    \hline
    product       & Products/Services & Product or service mentions. \\ 
    \hline
    monetary         & Monetary       & Currency and financial references. \\ 
    \hline
    company\_organization       & Company/Organization & Business or organization names. \\ 
    \hline
    other         & Others         & Named entities not in predefined types. \\ 
    \hline
    \end{tabular}
    \caption{The label set for the Entity Type bias dimension, including descriptions and short codes used for labeling.}
    \label{tab:entity_type_bias}
\end{table*}

\begin{table*}[ht]
    \centering
    \begin{tabular}{|l|l|p{12cm}|}
    \hline
    \textbf{Code} & \textbf{Label} & \textbf{Description} \\ 
    \hline
    very\_pos  & Very Positive  & Strongly positive tone \\ 
    \hline
    pos        & Positive       & Moderately positive tone \\ 
    \hline
    neg        & Negative       & Moderately negative tone \\ 
    \hline
    very\_neg  & Very Negative  & Strongly negative tone \\ 
    \hline
    info       & Informational  & Information content or presence of factual tokens (dates, names, IDs) — high priority over 'neutral' \\ 
    \hline
    neutral    & Neutral        & Does not have information and contains explicit neutral-emotion cues (e.g., “okay,” “fine,” “so-so,” “not sure”) \\ 
    \hline
    \end{tabular}
    \caption{The label set for the Sentiment bias dimension, including descriptions and short codes used for labeling.}
    \label{tab:sentiment_bias}
\end{table*}

\begin{table*}[ht]
    \centering
    \begin{tabular}{|l|l|p{8cm}|}
    \hline
    \textbf{Code} & \textbf{Label} & \textbf{Description} \\ 
    \hline
    greet        & Greetings/Introductions       & Greetings, introductions \\ 
    \hline
    id\_verif     & ID Verification               & ID or account verification \\ 
    \hline
    issue        & Issue/Problem Statement       & Customer's reason for contact \\ 
    \hline
    info\_gath    & Information Gathering         & Agent probing/investigating \\ 
    \hline
    prod\_inq     & Product Inquiry               & Product or service questions \\ 
    \hline
     diag         & Diagnosis                     & Diagnosis or troubleshooting \\ 
    \hline
    soln         & Solution                      & Proposing a solution \\ 
    \hline
    action       & Action                        & Performing an action \\ 
    \hline
    transact     & Transaction                   & Payments, refunds, orders \\ 
    \hline
    offers       & Offers                        & Service offers or upgrades \\ 
    \hline
    sales        & Sales                         & Sales, upselling, persuasion \\ 
    \hline
    resolve\_conf & Resolution Confirmation       & Confirming issue is resolved \\ 
    \hline
    next         & Next Steps                    & Next steps, follow-ups \\ 
    \hline
    close        & Closure                       & Farewell, call closure \\ 
    \hline
    empathy      & Empathy                       & Expressing care or rapport \\ 
    \hline
    complaint    & Complaint Handling            & Handling complaints/escalation \\ 
    \hline
    policy       & Policy Explanation            & Explaining rules or terms \\ 
    \hline
    feedback     & Feedback Request              & Requesting feedback or surveys \\ 
    \hline
    sched        & Scheduling                    & Appointments, scheduling \\ 
    \hline
    billing      & Billing Issues                & Billing/payment issues \\ 
    \hline
    compliance   & Compliance                    & Compliance or regulations \\ 
    \hline
    misc         & Miscellaneous                 & Miscellaneous \\ 
    \hline
    \end{tabular}

    \caption{The label set for the Topic bias dimension, including descriptions and short codes used for labeling.}
    \label{tab:topic_bias}
\end{table*}

\begin{table*}[ht]
    \centering
    \begin{tabular}{|l|l|p{8cm}|}
    \hline
    \textbf{Code} & \textbf{Label} & \textbf{Description} \\ 
    \hline
    filled      & Filled Pause        & "uh", "um", etc. \\ 
    \hline
    silent      & Silent Pause        & Silent pauses \\ 
    \hline
    repeat      & Repetition          & Word/phrase repetition \\ 
    \hline
    false\_start & False Start         & Incomplete start \\ 
    \hline
    repair      & Repair              & Self-correction \\ 
    \hline
    prolong     & Prolongation        & Stretched sounds \\ 
    \hline
    stutter     & Stutter             & Repeated syllables \\ 
    \hline
    marker      & Discourse Marker    & Discourse filler ("like", "you know") \\ 
    \hline
    interject   & Interjection        & "oh!", "hmm" \\ 
    \hline
    cutoff      & Cutoff              & Abandoned utterance \\ 
    \hline
    placeholder & Placeholder         & "sort of", "you know what I mean" \\ 
    \hline
    overlap     & Overlap             & Overlapping talk \\ 
    \hline
    \end{tabular}
    \caption{The label set for the Disfluency bias dimension, including descriptions and short codes used for labeling.}
    \label{tab:disfluency_bias}
\end{table*}

\begin{table*}[ht]
    \centering
    \begin{tabular}{|l|l|p{8cm}|}
    \hline
    \textbf{Code} & \textbf{Label} & \textbf{Description} \\ 
    \hline
    ask\_info     & Request Information        & Asking for details or clarification (e.g., "Could you confirm your order number?"). \\
    \hline
    give\_info    & Provide Information        & Supplying facts, context, or background not tied to a solution. \\
    \hline
    check\_under  & Confirm Understanding      & Verifying if the other party comprehends or observes the same thing (e.g., "Do you see the change on your end?"). \\
    \hline
    rapport       & Build Rapport              & Expressions of empathy, politeness, friendliness, or gratitude. \\
    \hline
    backchannel   & Acknowledgement / Cue      & Verbal cues like "Uh-huh", "Okay", or "Got it" to show active listening. \\
    \hline
    escalate      & Escalate / Transfer Action & Referring or handing over to another party or department. \\
    \hline
    compliance    & Compliance / Verification  & Fulfilling identity, legal, or policy requirements. \\
    \hline
    idle          & Passive / No-Op Response   & Moments of silence or minimal interaction without progress. \\
    \hline
    other         & Other Conversational Act   & Any conversational act not covered above, such as small talk. \\
    \hline
    \end{tabular}
    \caption{The label set for the Agent Action bias dimension, including descriptions and short codes used for labeling.}
    \label{tab:conversational_acts}
\end{table*}

\begin{table*}[ht]
    \centering
    \begin{tabular}{|l|l|p{8cm}|}
    \hline
    \textbf{Code} & \textbf{Label} & \textbf{Description} \\ 
    \hline
    very\_short  & Very Short & Dialogues with 0-5 tokens \\ 
    \hline
    short        & Short & Dialogues with 5-15 tokens\\ 
    \hline
    mid          & Mid & Dialogues with 15-50 tokens \\ 
    \hline
    long         & Long & Dialogues with 50-100 tokens \\ 
    \hline
    very\_long   & Very Long & Dialogues with more than 100 tokens \\ 
    \hline
    \end{tabular}
    \caption{The label set for the Turn Length bias dimension, including descriptions and short codes used for labeling.}
    \label{tab:dialogue_length_bias}
\end{table*}

\begin{table*}[ht]
    \centering
    \begin{tabular}{|l|p{4cm}|p{8cm}|}
    \hline
    \textbf{Code} & \textbf{Label} & \textbf{Description} \\
    \hline
    inorder      & Correct Order       & Events appear in the same order as in the original call. \\
    \hline
    early-shift  & Shifted Earlier     & An event appears earlier in the summary than in the original call. \\
    \hline
    late-shift   & Shifted Later       & An event appears later in the summary than in the original call. \\
    \hline
    omitted      & Omitted Event       & A key event from the original call is missing in the summary. \\
    \hline
    added        & Added Event         & The summary introduces an event not present in the original call. \\
    \hline
    \end{tabular}
    \caption{The label set for the Temporal Sequence bias dimension, including descriptions and short codes used for labeling.}
    \label{tab:temporal_sequence_bias}
\end{table*}

\begin{table*}[ht]
    \centering
    \begin{tabular}{|l|l|p{8cm}|}
    \hline
    \textbf{Code} & \textbf{Label} & \textbf{Description} \\
    \hline
    balanced   & Emotion Preserved & Summary preserves the exact sentiment(s) and intensity(s) of the transcript. \\
    \hline
    amplified  & Emotion Amplified & Summary intensifies existing sentiment(s): stronger valence or added emphasis beyond transcript. \\
    \hline
    attenuated & Emotion Attenuated & Summary weakens or omits sentiment: reduces intensity or drops emotion to neutral/informational. \\
    \hline
    inverted   & Emotion Inverted & Summary flips polarity: presents the opposite emotion to what the transcript expressed. \\
    \hline
    spurious   & Emotion Introduced & Summary introduces emotion where transcript was purely factual or neutral. \\
    \hline
    focused    & Emotion Narrowed & Transcript had multiple distinct emotions but summary reports only one (loss of nuance). \\
    \hline
    \end{tabular}
    \caption{The label set for the Emotion Shift bias dimension, including descriptions and short codes used for labeling.}
    \label{tab:emotion_shift_bias}
\end{table*}

\begin{table*}[ht]
    \centering
    \begin{tabular}{|l|l|p{8cm}|}
    \hline
    \textbf{Code} & \textbf{Label} & \textbf{Description} \\ 
    \hline
    no\_rep            & No Repetition & No repetition occurred. \\ 
    \hline
    cust\_self    & Customer Self-Repetition & Customer repeats their own words. \\ 
    \hline
    agent\_self       & Agent Self-Repetition & Agent repeats themselves. \\ 
    \hline
    cust\_echo & Customer Repeats Agent & Customer echoes agent. \\ 
    \hline
    agent\_echo    & Agent Repeats Customer & Agent echoes customer. \\ 
    \hline
    \end{tabular}
    \caption{The label set for the Information Repetition bias dimension, including descriptions and short codes used for labeling.}
    \label{tab:repetition_bias}
\end{table*}

\begin{table*}[ht]
    \centering
    \begin{tabular}{|l|l|p{9cm}|}
    \hline
    \textbf{Code} & \textbf{Label} & \textbf{Description} \\
    \hline
    standard\_clear         & Clear                   & Clear, direct, and easily understood language. \\
    \hline
    simple\_syntax          & Simple Syntax           & Predominantly short, declarative sentences. \\
    \hline
    complex\_syntax         & Complex Syntax          & Long, multi-clause, or convoluted sentences. \\
    \hline
    technical\_terms        & Technical Terms         & Specialized terms related to a specific domain. \\
    \hline
    industry\_jargon        & Industry Jargon         & Terms/phrases specific to an industry or company. \\
    \hline
    acronyms\_abbreviations & Abbreviations           & Use of shortened forms of words or phrases. \\
    \hline
    info\_dense             & Information Dense       & Highly concise; packed with specific information. \\
    \hline
    verbose\_hedging        & Verbose / Hedging       & Wordy, uses fillers, qualifiers, or vague language. \\
    \hline
    formal\_register        & Formal Register         & Polished, professional, and structured tone. \\
    \hline
    informal\_colloquial    & Informal / Colloquial   & Conversational, casual, everyday language. \\
    \hline
    empathetic\_softening   & Empathetic              & Language used to show understanding or soften news. \\
    \hline
    abrupt\_blunt           & Blunt                   & Overly direct, lacking typical softeners or politeness. \\
    \hline
    idioms\_slang           & Idioms / Slang          & Figurative expressions or informal slang. \\
    \hline
    passive\_voice\_prominent& Passive Voice           & Significant use of passive voice constructions. \\
    \hline
    \end{tabular}
    \caption{The label set for the Language Complexity bias dimension, including descriptions and short codes used for labeling.}
    \label{tab:language_complexity_bias}
\end{table*}

\begin{table*}[ht]
    \centering
    \begin{tabular}{|l|l|p{8cm}|}
    \hline
    \textbf{Code} & \textbf{Label} & \textbf{Description} \\ 
    \hline
    diag\_expl     & Diagnostic Explanation & Identifying the nature of the issue. \\
    \hline
    advisory       & General Advice         & Offering advice or suggestions. \\
    \hline
    root\_cause    & Root Cause             & Explaining the underlying reason for the issue. \\
    \hline
    directive      & Directive / Commands   & Concrete steps or commands to take. \\
    \hline
    preventive     & Preventive             & Preventing future issues from occurring. \\
    \hline
    escalate       & Escalation             & Escalation or transfer to another team. \\
    \hline
    self\_help     & Self-Help              & Do-it-yourself instructions. \\
    \hline
    partial        & Partial Fix            & Incomplete or partial resolution. \\
    \hline
    rejected       & Rejected               & Solution was offered but not applied. \\
    \hline
    followup       & Follow-Up              & Future action or check-in is promised. \\
    \hline
    expect         & Set Expectations       & Sets realistic timelines or expectations. \\
    \hline
    reassure       & Reassurance            & Provides emotional closure or comfort. \\
    \hline
    no\_soln       & No Solution            & No resolution was provided. \\
    \hline
    \end{tabular}
    \caption{The label set for the Solution bias dimension, including descriptions and short codes used for labeling.}
    \label{tab:solution_bias}
\end{table*}

\begin{table*}[ht]
    \centering
    \begin{tabular}{|l|l|p{10cm}|}
    \hline
    \textbf{Code} & \textbf{Label} & \textbf{Description} \\ 
    \hline
    none      & None      & No politeness cues (no please/thank you/etc.) \\ 
    \hline
    minimal   & Minimal   & One-off courtesy (“thank you”, “please”) \\ 
    \hline
    standard  & Standard  & Expected level (“please let me know”, “thanks for waiting”) \\ 
    \hline
    elevated  & Elevated  & Multiple markers + honorifics (“sir/madam”, “kindly”) \\ 
    \hline
    impolite & Impolite & Impoliteness cues \\ 
    \hline
    \end{tabular}
    \caption{The label set for the Politeness bias dimension, including descriptions and short codes used for labeling.}
    \label{tab:politeness_bias}
\end{table*}

\begin{table*}[ht]
    \centering
    \begin{tabular}{|l|l|p{11cm}|}
    \hline
    \textbf{Code} & \textbf{Label} & \textbf{Description} \\ 
    \hline
    none      & None      & No urgency language \\ 
    \hline
    low       & Low       & Mild timeframe hints (“when you can”, “at your convenience”) \\ 
    \hline
    moderate  & Moderate  & Moderate urgency (“soon”, “shortly”) \\ 
    \hline
    high      & High      & Strong urgency (“ASAP”, “urgent”) \\ 
    \hline
    critical  & Critical  & Extreme immediacy (“immediately”, “right now”, “without delay”) \\ 
    \hline
    \end{tabular}
    \caption{The label set for the Urgency bias dimension, including descriptions and short codes used for labeling.}
    \label{tab:urgency_bias}
\end{table*}

\clearpage

\section{Framework Methodology and Implementation}
\label{appendix:methodology}

This section provides a detailed description of the \textit{BlindSpot} framework's methodology, including the end-to-end data processing workflow, the validation of our LLM Labeler, and a guide to interpreting the final bias metrics.

\subsection{Detailed Workflow of \textit{BlindSpot}}
\label{subsec:workflow}

The framework's core function is to quantify bias by comparing the distributional properties of a source transcript and its generated summary. The process, illustrated in Figure~\ref{fig:experiments} of the main paper, is composed of three main stages: (1) creating a reference distribution from the transcript, (2) creating a summary distribution, and (3) calculating bias metrics.

\subsubsection{Stage 1: Transcript Pipeline (Generating Reference Distribution \(P_d\))}
The objective of this pipeline is to establish a reference label distribution, \(P_d\), for each of the 15 bias dimensions.

\paragraph{1. Transcript Segmentation} To manage long contexts and ensure consistent JSON output from the LLM Labeler, each transcript \(T\) is first partitioned into sequential, non-overlapping segments $\{S_1, \dots, S_k\}$ of 50 turns each. This segmentation mitigates potential performance degradation and out-of-spec responses when processing very long transcripts in a single pass.

\paragraph{2. Turn-level Annotation} We employ a hybrid approach to annotate every turn in the transcript across all bias dimensions. The annotation source depends on the nature of the dimension:
\begin{itemize}
    \item \textbf{LLM-Annotated (Semantic Dimensions):} For dimensions requiring semantic understanding, we use our LLM Labeler ($\mathcal{L}$) to process each segment and assign labels. These include \textit{Sentiment, Topic, Solution, Information Repetition, Language Complexity, Disfluency, Politeness, Urgency, Entity Type,} and \textit{Agent Action}.
    \item \textbf{Computed (Structural Dimensions):} For dimensions based on the transcript's structure, labels are computed algorithmically. \textit{Speaker} is extracted directly from conversation metadata. \textit{Position} is calculated by normalizing a turn's index into one of five quintiles (`Very Early`, `Early`, `Mid`, `Late`, `Very Late`). \textit{Turn Length} is determined by the token count of the turn, categorized into discrete length buckets.
    \item \textbf{Derived (Relational Dimensions):} Two dimensions are inferred from the primary labels. \textit{Emotion Shift} is derived by comparing the sentiment of a proposition to its source turns, and \textit{Temporal Sequence} is derived from the mapping between chronologically ordered summary propositions and their source turn indices.
\end{itemize}

\paragraph{3. Reference Distribution (\(P_d\)) Generation} The turn-level annotations are aggregated across the entire transcript to form a normalized categorical distribution \(P_d\) for each dimension \(d\), which serves as our reference or ``ground truth.''

\subsubsection{Stage 2: Summary Pipeline (Generating Summary Distribution \(Q_d\))}
This pipeline generates a corresponding distribution, \(Q_d\), from the LLM-generated summary.

\paragraph{1. Summary Generation} The summarization model under evaluation, $\mathcal{M}$, generates an abstractive summary \(S\) from the full, unsegmented transcript \(T\). This mirrors real-world usage where the model processes the entire conversation at once.

\paragraph{2. Proposition Extraction} To enable fine-grained, sentence-level analysis, the generated summary \(S\) is decomposed into a set of minimal semantic units, or \textbf{propositions} $\{p_1, \dots, p_m\}$. This is performed by an LLM instructed to isolate each atomic fact or claim, creating a standardized unit of analysis.

\paragraph{3. Proposition Labeling and Mapping} Each proposition is then labeled using the same hybrid methodology as the transcript turns. For turn-dependent dimensions like \textit{Position} or \textit{Urgency}, a crucial mapping step is performed where the LLM Labeler identifies the set of source turn indices that each proposition summarizes (a one-to-many mapping). The labels from these source turns are then projected onto the proposition.

\paragraph{4. Summary Distribution (\(Q_d\)) Generation} The proposition-level labels are aggregated to form the summary's categorical distribution \(Q_d\) for each dimension \(d\).

\subsubsection{Stage 3: Bias Quantification and Interpretation}

\paragraph{1. Metric Calculation} With both \(P_d\) and \(Q_d\) computed, we quantify bias using two complementary metrics:
\begin{itemize}
    \item \textbf{Fidelity Gap:} We use Jensen-Shannon (JS) Divergence between \(P_d\) and \(Q_d\) to measure the overall distributional distortion. A score of 0 indicates identical distributions, while higher values indicate greater divergence.
    \item \textbf{Coverage:} We calculate the percentage of labels present in the transcript (where \(P_d(c) > 0\)) that are also present in the summary (where \(Q_d(c) > 0\)). This directly measures the omission of information.
\end{itemize}
For derived dimensions like \textit{Temporal Sequence}, the reference distribution \(P_d\) is a one-hot distribution representing the ideal chronological order, allowing JSD to directly measure any reordering.

\paragraph{2. Interpreting Results} The combination of our two metrics provides a nuanced view of a summary's faithfulness. For each dimension, we interpret the pair as follows:
\begin{itemize}
  \item \textbf{Low Fidelity Gap \& High Coverage:} A faithful summary that retains nearly all source labels and preserves their original proportions.
  \item \textbf{Low Fidelity Gap \& Low Coverage:} A selectively faithful summary that accurately represents the distribution of the labels it includes but omits other labels entirely.
  \item \textbf{High Fidelity Gap \& High Coverage:} A distorting summary that mentions information from all source labels but skews their relative importance, leading to misrepresentation.
  \item \textbf{High Fidelity Gap \& Low Coverage:} The worst-case scenario; a summary that both ignores entire sets of labels and misrepresents those it chooses to include.
\end{itemize}

\subsection{LLM Labeler Validation}
\label{subsec:llm_validation}

The integrity of our framework hinges on the reliability of our LLM Labeler ($\mathcal{L}$), GPT-4o. To validate its performance, we conducted a rigorous human annotation study.

\begin{enumerate}
    \item \textbf{Dataset Creation:} We randomly sampled 1,000 turn-proposition pairs from our dataset, ensuring coverage across all 15 bias dimensions. A human annotator trained in contact center analytics and familiar with the operational context, independently validated each label assigned by the LLM Labeler according to detailed annotation guidelines.
    \item \textbf{Evaluation:} The LLM Labeler ($\mathcal{L}$) achieved an accuracy of 93.68\% against human annotation. As expected, performance varied slightly by dimension, with higher accuracy on objective dimensions like \textit{Entity Type} and slightly lower, yet still high, accuracy on more subjective dimensions like \textit{Politeness}. This result gave us confidence in using the LLM as a scalable and reliable tool for our large-scale analysis.
\end{enumerate}

\subsection{LLM-Judge for Holistic Quality Assessment}
\label{subsec:llm_judge}

To contextualize our fine-grained bias findings, we also measure the overall, holistic quality of each summary using an ``LLM-as-a-Judge'' approach. The goal is to establish a baseline quality score against which we can compare our bias metrics. This allows us to investigate a central question of our work: can summaries that are perceived as high-quality by a powerful LLM still harbor  operational biases?

\paragraph{Implementation Details}
For each of the 50,000 transcript-summary pairs (2500 transcripts \(\times\) 20 models), we prompt a powerful arbitrator LLM (GPT-4o) to act as an impartial judge. As detailed in Box~\ref{box:prompt_llm_judge}, the judge is tasked with assigning an integer score from 1 (poor) to 5 (excellent) based on three explicit criteria derived from standard summarization quality dimensions:
\begin{enumerate}
    \item \textbf{Factual Consistency:} This criterion ensures that all claims, facts, and events mentioned in the summary are factually supported by the source transcript. It penalizes any hallucinations or contradictions.
    \item \textbf{Completeness:} This assesses whether the summary includes all critical information from the conversation without significant omissions of key events, decisions, or outcomes.
    \item \textbf{Succinctness and Relevance:} This criterion, framed in the prompt as ``Presence of irrelevant information,'' penalizes summaries that include extraneous details, conversational filler, or other information not directly relevant to the core purpose of the interaction.
\end{enumerate}
The judge is instructed to output both the numerical score and a brief textual justification for its reasoning. For our quantitative analysis, we use the numerical score, which we refer to as the \textbf{LLM Judge Score}.

\begin{tcolorbox}[
    breakable, 
    colframe=orange!85!black, 
    colback=orange!10!white, 
    title=Prompt for LLM Judge Score,
    label={box:prompt_llm_judge}
]
You are provided with an input call transcript and its abstractive summary. Your task is to evaluate the quality of the summary according to the transcript.

Assign an integer score between 1 and 5 (higher the score, better the response quality).

Evaluate the response using the following criteria:
\begin{enumerate}
    \item \textbf{Factual Consistency} - Are the facts and claims in the summary correct?
    \item \textbf{Completeness} - Is all necessary information included?
    \item \textbf{Presence of irrelevant information} - Does the summary stay focused on the task?
\end{enumerate}

\textbf{Output Format}:
Score: [1-5]
Reason: [Feedback on prompt]

\end{tcolorbox}

\paragraph{Acknowledged Limitations}
While scalable and effective for capturing general quality, we acknowledge the known limitations of the LLM-as-a-Judge paradigm. These include potential agreement bias (a tendency to favor summaries stylistically similar to its own training data), positional bias (over-weighting information at the beginning or end of the summary), and an inability to detect subtle but operationally critical omissions that our \textit{BlindSpot} framework is designed to find. Therefore, we use this score not as an absolute measure of truth, but as a proxy for a summary's perceived holistic quality. The potential for this high-level score to mask the fine-grained biases we investigate is a central motivation for our work.

\clearpage
\section{Experimental Configuration}
\label{appendix:experimental_setup}

This section provides a detailed overview of the experimental configuration used in our study, including model generation parameters, dataset statistics, and the full list of evaluated models.

\subsection{Generation Parameters}

To ensure a fair and reproducible comparison, we employed a standardized set of generation parameters for all summarization tasks. The specific settings were chosen to elicit factual and deterministic outputs while accommodating different model types.
For the majority of models, we set the temperature to 0 to minimize randomness and produce the most likely, consistent summary for a given transcript. For reasoning models, we used a temperature of 1 and set reasoning\_effort to low. Other key parameters, such as top\_p, frequency\_penalty, and presence\_penalty, were set to neutral values to avoid confounding the results and to observe the models' inherent summarization behaviors. The maximum output length was capped at 1000 tokens, which was sufficient for all summaries in our corpus.

\begin{table}[!hbt]
\centering
\small
\begin{tabular}{l r}
\toprule
\textbf{Parameter} & \textbf{Value} \\
\midrule
Temperature (non-reasoning models) & 0 \\
Temperature (reasoning models) & 1 \\
Top-p & 1.0 \\
Max Tokens & 1000 \\
Frequency Penalty & 0.0 \\
Presence Penalty & 0.0 \\
Stop & None \\
Seed & None \\
Reasoning Effort (reasoning models) & low \\
\bottomrule
\end{tabular}
\caption{LLM generation parameters for summarization.}
\label{tab:generation_params_summarizer}
\end{table}

For the LLM Labeler, which performs the labeling tasks in our framework, we used GPT-4o with slightly different parameters to balance consistency with nuanced classification. A low temperature of 0.1 was chosen to ensure high reproducibility and determinism while allowing for minimal flexibility.

\begin{table}[!hbt]
\centering
\small
\begin{tabular}{l r}
\toprule
\textbf{Parameter} & \textbf{Value} \\
\midrule
LLM & GPT-4o \\
Temperature & 0.1 \\
Top-p & 1.0 \\
Max Tokens & None \\
Frequency Penalty & 0.0 \\
Presence Penalty & 0.0 \\
Stop & None \\
Seed & None \\
\bottomrule
\end{tabular}
\caption{LLM generation parameters for LLM Labeler.}
\label{tab:generation_params_labeler}
\end{table}

\subsection{Models Evaluated}
To conduct a comprehensive audit of bias, we selected a diverse set of 20 large language models. Our selection spans multiple major model providers and open-source families, including Meta (Llama), Amazon (Nova), Anthropic (Claude), Google (Gemini), and OpenAI (GPT). Furthermore, we intentionally included models of varying scales within the same family (e.g., Llama-3.2 1B vs. Llama-3.3 70B; GPT-4.1-nano vs. GPT-4.1). This approach allows us to analyze the influence of both model architecture and parameter scale on the prevalence and nature of biases. For full transparency and reproducibility, the specific model identifiers used in our experiments are listed in Table \ref{tab:llm_versions}.

\begin{table}[!hbt]
\centering
\resizebox{\columnwidth}{!}{%
\begin{tabular}{l l}
\toprule
\textbf{Short Name} & \textbf{Model ID} \\
\midrule
\texttt{Llama-3.2-1B}        & \texttt{meta/llama3-2-1b-instruct-v1} \\
\texttt{Llama-3.2-3B}        & \texttt{meta/llama3-2-3b-instruct-v1} \\
\texttt{Llama-3.3-70B}       & \texttt{meta/llama3-3-70b-instruct-v1} \\
\texttt{Llama-4-Maverick}    & \texttt{meta/llama4-maverick-17b-instruct-v1} \\
\texttt{Nova Micro}          & \texttt{amazon/nova-micro-v11} \\
\texttt{Nova Lite}           & \texttt{amazon/nova-lite-v1} \\
\texttt{Nova Pro}            & \texttt{amazon/nova-pro-v1} \\
\texttt{Claude-3.5-Haiku}    & \texttt{anthropic/claude-3-5-haiku-20241022-v1} \\
\texttt{Claude-3.7-Sonnet}   & \texttt{anthropic/claude-3-7-sonnet-20250219-v1} \\
\texttt{Claude-4-Sonnet}     & \texttt{anthropic/claude-sonnet-4-20250514-v1} \\
\texttt{Deepseek-R1}         & \texttt{deepseek/r1-v1} \\
\texttt{Gemini-2.0-Flash}    & \texttt{google/gemini-2.0-flash} \\
\texttt{Gemini-2.0-Flash-lite} & \texttt{google/gemini-2.0-flash-lite} \\
\texttt{GPT-4o-mini}         & \texttt{openai/gpt-4o-mini-2024-07-18} \\
\texttt{GPT-4o}              & \texttt{openai/gpt-4o-2024-08-06} \\
\texttt{GPT-4.1-nano}        & \texttt{openai/gpt-4.1-nano-2025-04-14} \\
\texttt{GPT-4.1-mini}        & \texttt{openai/gpt-4.1-mini-2025-04-14} \\
\texttt{GPT-4.1}             & \texttt{openai/gpt-4.1-2025-04-14} \\
\texttt{o3-mini}             & \texttt{openai/o3-mini-2025-01-31} \\
\texttt{o4-mini}             & \texttt{openai/o4-mini-2025-04-16} \\
\bottomrule
\end{tabular}
}
\caption{Identifiers of LLMs used in evaluation.}
\label{tab:llm_versions}
\end{table}

\subsection{Transcript Statistics}
Our evaluation was conducted on a corpus of real-world, anonymized contact center transcripts from 12 distinct domains. As shown in Table \ref{tab:transcript_statistics}, the conversations are substantial and highly variable in length. The average transcript contains approximately 317 turns and over 5,000 tokens, with the longest conversation extending to 548 turns and over 11,000 tokens. This significant variation in length and content provides a robust testbed for evaluating the models' summarization capabilities across a range of complexities, from brief, straightforward interactions to long, multi-issue dialogues. The distribution is slightly right-skewed, with the median length (290 turns) and call duration (37 minutes and 31 seconds) being lower than the mean, which is typical for such datasets.

\begin{table}[!hbt]
\centering
\resizebox{\columnwidth}{!}{%
\begin{tabular}{lcrc}
\toprule
\textbf{Statistic} & \textbf{num\_turns} & \textbf{token\_count} & \textbf{call\_duration (mm:ss)}\\
\midrule
Count & 549 & 605 & 09:09 \\
Mean  & 317 & 5110 & 36:58\\
Std   & 128 & 2180 & 13:52\\
Min   & 55  & 244 & 10:01\\
25\%  & 214 & 3003 & 28:20 \\
50\% & 290 & 5048 & 37:31\\
75\%  & 429 & 6840 & 44:32\\
Max   & 548 & 11348 & 92:32\\
\bottomrule
\end{tabular}
}
\caption{Summary statistics of number of turns and token counts across transcripts.}
\label{tab:transcript_statistics}
\end{table}

\clearpage

\section{Supplemental Results and Analysis}

This appendix provides additional results and analyses that complement the findings presented in the main paper. It includes a comprehensive breakdown of model performance across all bias dimensions, results using alternative divergence metrics, an analysis of how performance varies with transcript length, and a deeper look at model-level representation biases.

\subsection{Model Performance with Standard Deviation}
Table~\ref{table:main_results_with_deviation} presents the complete evaluation results for all 20 LLMs across the 15 bias dimensions, including both the mean and standard deviation for each metric. These detailed results support the main paper's claim that bias is a systemic issue, with most models clustering within a narrow performance band for many dimensions. The standard deviation values indicate the consistency of a model's performance across the 2500 transcripts. 

\begin{table*}[!hbt]
  \centering
  \small
  \renewcommand{\arraystretch}{1.2}
  \resizebox{\textwidth}{!}{%
    \begin{tabular}{@{}l *{21}{>{\centering\arraybackslash}p{1.2cm}} @{}}

      \textbf{Metric / Bias} 
        & \rotatebox{90}{\texttt{llama-3.2-1b}} 
        & \rotatebox{90}{\texttt{llama-3.2-3b}} 
        & \rotatebox{90}{\texttt{llama-3.3-70b}} 
        & \rotatebox{90}{\texttt{llama-4-maverick}} 
        & \rotatebox{90}{\texttt{nova-micro}} 
        & \rotatebox{90}{\texttt{nova-lite}} 
        & \rotatebox{90}{\texttt{nova-pro}} 
        & \rotatebox{90}{\texttt{claude-3.5-haiku}} 
        & \rotatebox{90}{\texttt{claude-3.7-sonnet}} 
        & \rotatebox{90}{\texttt{claude-4-sonnet}} 
        & \rotatebox{90}{\texttt{deepseek-r1}} 
        & \rotatebox{90}{\texttt{gemini-2.0-flash}} 
        & \rotatebox{90}{\texttt{gemini-2.0-flash-lite}} 
        & \rotatebox{90}{\texttt{gpt-4o-mini}} 
        & \rotatebox{90}{\texttt{gpt-4o}} 
        & \rotatebox{90}{\texttt{gpt-4.1-nano}} 
        & \rotatebox{90}{\texttt{gpt-4.1-mini}} 
        & \rotatebox{90}{\texttt{gpt-4.1}} 
        & \rotatebox{90}{\texttt{o3-mini}} 
        & \rotatebox{90}{\texttt{o4-mini}}
        & \rotatebox{90}
        {\textbf{Average}}\\
      \midrule
      \multicolumn{21}{@{}l}{\textit{\textbf{JS Divergence (JSD)}} ($\downarrow$ better)} \\
      Position       & $0.026 \pm 0.024$ & $0.019 \pm 0.015$ & $0.016 \pm 0.012$ & $0.017 \pm 0.014$ & $0.017 \pm 0.014$ & $0.016 \pm 0.014$ & $0.019 \pm 0.014$ & $0.016 \pm 0.013$ & $0.017 \pm 0.013$ & $0.017 \pm 0.013$ & $0.017 \pm 0.018$ & $0.077 \pm 0.121$ & $0.076 \pm 0.117$ & $0.017 \pm 0.014$ & $0.017 \pm 0.014$ & $0.014 \pm 0.012$ & $0.015 \pm 0.011$ & $0.016 \pm 0.013$ & $0.015 \pm 0.012$ & $0.017 \pm 0.013$ & $0.023 \pm 0.022$ \\
      Speaker        & $0.016 \pm 0.023$ & $0.016 \pm 0.024$ & $0.014 \pm 0.018$ & $0.014 \pm 0.021$ & $0.018 \pm 0.024$ & $0.016 \pm 0.022$ & $0.016 \pm 0.022$ & $0.013 \pm 0.019$ & $0.012 \pm 0.018$ & $0.011 \pm 0.015$ & $0.014 \pm 0.020$ & $0.048 \pm 0.075$ & $0.048 \pm 0.073$ & $0.012 \pm 0.019$ & $0.014 \pm 0.020$ & $0.015 \pm 0.022$ & $0.013 \pm 0.017$ & $0.013 \pm 0.018$ & $0.015 \pm 0.021$ & $0.014 \pm 0.019$ & $0.018 \pm 0.024$ \\
      Sentiment      & $0.041 \pm 0.034$ & $0.041 \pm 0.034$ & $0.038 \pm 0.031$ & $0.040 \pm 0.031$ & $0.039 \pm 0.034$ & $0.040 \pm 0.032$ & $0.040 \pm 0.032$ & $0.043 \pm 0.034$ & $0.046 \pm 0.036$ & $0.048 \pm 0.036$ & $0.046 \pm 0.037$ & $0.069 \pm 0.100$ & $0.068 \pm 0.103$ & $0.038 \pm 0.032$ & $0.040 \pm 0.033$ & $0.036 \pm 0.032$ & $0.040 \pm 0.032$ & $0.039 \pm 0.032$ & $0.036 \pm 0.031$ & $0.043 \pm 0.032$ & $0.044 \pm 0.036$ \\
      Topic          & $0.058 \pm 0.031$ & $0.050 \pm 0.027$ & $0.047 \pm 0.025$ & $0.048 \pm 0.027$ & $0.052 \pm 0.029$ & $0.050 \pm 0.028$ & $0.054 \pm 0.029$ & $0.054 \pm 0.029$ & $0.057 \pm 0.030$ & $0.058 \pm 0.030$ & $0.057 \pm 0.029$ & $0.128 \pm 0.140$ & $0.121 \pm 0.136$ & $0.045 \pm 0.025$ & $0.050 \pm 0.027$ & $0.047 \pm 0.029$ & $0.048 \pm 0.026$ & $0.047 \pm 0.027$ & $0.046 \pm 0.026$ & $0.053 \pm 0.029$ & $0.060 \pm 0.035$ \\
      Agent Action   & $0.180 \pm 0.062$ & $0.178 \pm 0.058$ & $0.174 \pm 0.058$ & $0.178 \pm 0.058$ & $0.182 \pm 0.059$ & $0.182 \pm 0.058$ & $0.184 \pm 0.061$ & $0.182 \pm 0.060$ & $0.188 \pm 0.058$ & $0.188 \pm 0.058$ & $0.189 \pm 0.060$ & $0.215 \pm 0.079$ & $0.213 \pm 0.081$ & $0.175 \pm 0.061$ & $0.178 \pm 0.059$ & $0.176 \pm 0.060$ & $0.180 \pm 0.058$ & $0.178 \pm 0.058$ & $0.177 \pm 0.059$ & $0.185 \pm 0.059$ & $0.184 \pm 0.061$ \\
      Solution       & $0.046 \pm 0.067$ & $0.030 \pm 0.042$ & $0.029 \pm 0.044$ & $0.029 \pm 0.045$ & $0.028 \pm 0.046$ & $0.027 \pm 0.044$ & $0.032 \pm 0.049$ & $0.031 \pm 0.045$ & $0.035 \pm 0.049$ & $0.035 \pm 0.050$ & $0.032 \pm 0.050$ & $0.073 \pm 0.110$ & $0.068 \pm 0.107$ & $0.028 \pm 0.043$ & $0.027 \pm 0.040$ & $0.023 \pm 0.034$ & $0.027 \pm 0.048$ & $0.025 \pm 0.038$ & $0.024 \pm 0.037$ & $0.027 \pm 0.037$ & $0.034 \pm 0.049$ \\
      Politeness     & $0.036 \pm 0.035$ & $0.038 \pm 0.036$ & $0.035 \pm 0.034$ & $0.035 \pm 0.034$ & $0.038 \pm 0.038$ & $0.037 \pm 0.035$ & $0.037 \pm 0.034$ & $0.033 \pm 0.032$ & $0.032 \pm 0.032$ & $0.031 \pm 0.032$ & $0.035 \pm 0.036$ & $0.066 \pm 0.100$ & $0.063 \pm 0.094$ & $0.034 \pm 0.034$ & $0.035 \pm 0.036$ & $0.031 \pm 0.031$ & $0.033 \pm 0.032$ & $0.033 \pm 0.031$ & $0.031 \pm 0.031$ & $0.035 \pm 0.035$ & $0.037 \pm 0.037$ \\
      Urgency        & $0.025 \pm 0.042$ & $0.023 \pm 0.040$ & $0.023 \pm 0.042$ & $0.023 \pm 0.037$ & $0.024 \pm 0.042$ & $0.023 \pm 0.041$ & $0.024 \pm 0.041$ & $0.025 \pm 0.045$ & $0.025 \pm 0.041$ & $0.027 \pm 0.047$ & $0.026 \pm 0.042$ & $0.049 \pm 0.097$ & $0.045 \pm 0.090$ & $0.022 \pm 0.039$ & $0.024 \pm 0.039$ & $0.022 \pm 0.039$ & $0.023 \pm 0.041$ & $0.024 \pm 0.042$ & $0.022 \pm 0.040$ & $0.024 \pm 0.043$ & $0.026 \pm 0.043$ \\
      Order          & $0.394 \pm 0.076$ & $0.358 \pm 0.077$ & $0.337 \pm 0.080$ & $0.356 \pm 0.091$ & $0.382 \pm 0.085$ & $0.370 \pm 0.084$ & $0.387 \pm 0.093$ & $0.362 \pm 0.080$ & $0.358 \pm 0.081$ & $0.348 \pm 0.084$ & $0.347 \pm 0.082$ & $0.467 \pm 0.147$ & $0.467 \pm 0.149$ & $0.380 \pm 0.080$ & $0.385 \pm 0.084$ & $0.351 \pm 0.084$ & $0.326 \pm 0.081$ & $0.333 \pm 0.083$ & $0.353 \pm 0.081$ & $0.349 \pm 0.089$ & $0.370 \pm 0.088$ \\
      Emotion        & $0.116 \pm 0.076$ & $0.144 \pm 0.075$ & $0.138 \pm 0.068$ & $0.129 \pm 0.069$ & $0.140 \pm 0.071$ & $0.137 \pm 0.072$ & $0.132 \pm 0.073$ & $0.131 \pm 0.074$ & $0.116 \pm 0.072$ & $0.119 \pm 0.074$ & $0.128 \pm 0.080$ & $0.119 \pm 0.127$ & $0.112 \pm 0.119$ & $0.149 \pm 0.070$ & $0.137 \pm 0.071$ & $0.137 \pm 0.079$ & $0.129 \pm 0.074$ & $0.125 \pm 0.071$ & $0.122 \pm 0.077$ & $0.107 \pm 0.074$ & $0.128 \pm 0.077$ \\
      Repetition     & $0.091 \pm 0.112$ & $0.093 \pm 0.122$ & $0.084 \pm 0.109$ & $0.080 \pm 0.108$ & $0.086 \pm 0.112$ & $0.090 \pm 0.115$ & $0.087 \pm 0.106$ & $0.084 \pm 0.109$ & $0.089 \pm 0.114$ & $0.086 \pm 0.113$ & $0.087 \pm 0.111$ & $0.100 \pm 0.121$ & $0.100 \pm 0.121$ & $0.078 \pm 0.102$ & $0.089 \pm 0.113$ & $0.078 \pm 0.108$ & $0.079 \pm 0.104$ & $0.082 \pm 0.107$ & $0.075 \pm 0.105$ & $0.085 \pm 0.111$ & $0.087 \pm 0.110$ \\
      Disfluency     & $0.055 \pm 0.047$ & $0.052 \pm 0.045$ & $0.050 \pm 0.045$ & $0.051 \pm 0.044$ & $0.050 \pm 0.042$ & $0.052 \pm 0.046$ & $0.054 \pm 0.049$ & $0.051 \pm 0.045$ & $0.052 \pm 0.046$ & $0.053 \pm 0.048$ & $0.053 \pm 0.046$ & $0.076 \pm 0.077$ & $0.075 \pm 0.074$ & $0.049 \pm 0.045$ & $0.051 \pm 0.044$ & $0.048 \pm 0.044$ & $0.051 \pm 0.047$ & $0.051 \pm 0.049$ & $0.049 \pm 0.047$ & $0.054 \pm 0.048$ & $0.054 \pm 0.048$ \\
      Length         & $0.016 \pm 0.014$ & $0.014 \pm 0.013$ & $0.013 \pm 0.011$ & $0.015 \pm 0.012$ & $0.015 \pm 0.012$ & $0.014 \pm 0.012$ & $0.015 \pm 0.012$ & $0.014 \pm 0.012$ & $0.015 \pm 0.012$ & $0.014 \pm 0.012$ & $0.015 \pm 0.014$ & $0.048 \pm 0.080$ & $0.048 \pm 0.085$ & $0.013 \pm 0.011$ & $0.014 \pm 0.011$ & $0.013 \pm 0.011$ & $0.013 \pm 0.011$ & $0.013 \pm 0.011$ & $0.013 \pm 0.012$ & $0.015 \pm 0.012$ & $0.017 \pm 0.018$ \\
      Language       & $0.041 \pm 0.029$ & $0.038 \pm 0.028$ & $0.035 \pm 0.025$ & $0.037 \pm 0.027$ & $0.038 \pm 0.027$ & $0.036 \pm 0.026$ & $0.039 \pm 0.028$ & $0.036 \pm 0.026$ & $0.036 \pm 0.026$ & $0.036 \pm 0.027$ & $0.039 \pm 0.029$ & $0.081 \pm 0.105$ & $0.083 \pm 0.110$ & $0.034 \pm 0.025$ & $0.035 \pm 0.025$ & $0.034 \pm 0.025$ & $0.035 \pm 0.024$ & $0.035 \pm 0.026$ & $0.033 \pm 0.024$ & $0.038 \pm 0.026$ & $0.041 \pm 0.032$ \\
      Entity         & $0.170 \pm 0.082$ & $0.158 \pm 0.079$ & $0.147 \pm 0.082$ & $0.136 \pm 0.077$ & $0.180 \pm 0.087$ & $0.173 \pm 0.085$ & $0.176 \pm 0.094$ & $0.116 \pm 0.069$ & $0.096 \pm 0.065$ & $0.086 \pm 0.058$ & $0.120 \pm 0.071$ & $0.169 \pm 0.090$ & $0.190 \pm 0.094$ & $0.181 \pm 0.086$ & $0.169 \pm 0.086$ & $0.190 \pm 0.094$ & $0.146 \pm 0.084$ & $0.149 \pm 0.081$ & $0.173 \pm 0.097$ & $0.111 \pm 0.070$ & $0.152 \pm 0.082$ \\
      \midrule
    \textbf{Average} & 0.087 & 0.084 & 0.079 & 0.079 & 0.086 & 0.084 & 0.086 & 0.079 & 0.078 & 0.077 & 0.080 & 0.119 & 0.119 & 0.084 & 0.084 & 0.081 & 0.077 & 0.078 & 0.079 & 0.077 & \textbf{--} \\
    
      \midrule
      \multicolumn{21}{@{}l}{\textit{\textbf{Coverage}} (↑ better)} \\
      Position       & $98.79 \pm 9.57$ & $97.77 \pm 14.59$ & $98.17 \pm 13.42$ & $97.50 \pm 14.02$ & $98.07 \pm 13.48$ & $97.93 \pm 14.04$ & $98.03 \pm 13.50$ & $97.93 \pm 14.04$ & $98.03 \pm 13.50$ & $97.67 \pm 14.69$ & $97.66 \pm 14.79$ & $79.38 \pm 34.27$ & $80.57 \pm 33.47$ & $98.23 \pm 12.82$ & $98.40 \pm 12.22$ & $97.80 \pm 14.58$ & $98.00 \pm 14.00$ & $97.77 \pm 14.59$ & $97.97 \pm 14.02$ & $98.03 \pm 13.50$ & $96.18 \pm 16.14$ \\
      Speaker        & $99.16 \pm 9.14$ & $97.83 \pm 14.56$ & $98.17 \pm 13.42$ & $97.67 \pm 14.96$ & $98.17 \pm 14.00$ & $98.00 \pm 14.00$ & $98.17 \pm 14.56$ & $98.00 \pm 14.00$ & $98.17 \pm 13.42$ & $97.83 \pm 14.56$ & $97.83 \pm 14.56$ & $84.50 \pm 30.85$ & $86.08 \pm 29.33$ & $98.33 \pm 12.19$ & $98.50 \pm 12.16$ & $97.83 \pm 14.56$ & $98.00 \pm 14.00$ & $97.83 \pm 14.56$ & $98.00 \pm 14.00$ & $98.17 \pm 13.42$ & $96.81 \pm 15.30$ \\
      Sentiment      & $89.00 \pm 16.52$ & $90.13 \pm 18.29$ & $91.52 \pm 17.28$ & $90.15 \pm 18.71$ & $89.54 \pm 17.87$ & $90.74 \pm 17.74$ & $89.44 \pm 18.02$ & $89.31 \pm 18.60$ & $88.93 \pm 18.40$ & $88.23 \pm 19.01$ & $88.99 \pm 18.81$ & $71.42 \pm 32.44$ & $72.89 \pm 31.73$ & $92.05 \pm 16.48$ & $90.72 \pm 16.66$ & $91.13 \pm 17.67$ & $91.17 \pm 17.65$ & $90.25 \pm 18.50$ & $90.16 \pm 17.94$ & $88.70 \pm 18.81$ & $88.22 \pm 18.47$ \\
      Topic          & $75.54 \pm 14.81$ & $79.11 \pm 16.98$ & $81.03 \pm 15.78$ & $79.44 \pm 16.72$ & $78.12 \pm 16.35$ & $78.83 \pm 16.36$ & $76.58 \pm 15.78$ & $76.04 \pm 16.22$ & $74.55 \pm 15.58$ & $72.85 \pm 15.94$ & $75.72 \pm 16.19$ & $54.42 \pm 30.04$ & $56.96 \pm 30.60$ & $81.53 \pm 15.11$ & $79.59 \pm 15.04$ & $80.37 \pm 16.89$ & $79.12 \pm 15.99$ & $79.62 \pm 16.24$ & $78.96 \pm 15.96$ & $75.20 \pm 16.19$ & $75.68 \pm 16.61$ \\
      Agent Action   & $67.74 \pm 17.60$ & $68.19 \pm 18.72$ & $70.62 \pm 18.54$ & $68.80 \pm 18.46$ & $67.00 \pm 18.70$ & $68.22 \pm 18.27$ & $65.96 \pm 17.98$ & $66.90 \pm 18.71$ & $64.71 \pm 18.18$ & $64.71 \pm 18.69$ & $64.40 \pm 18.44$ & $51.29 \pm 27.03$ & $53.41 \pm 26.78$ & $70.12 \pm 17.13$ & $68.69 \pm 17.53$ & $70.59 \pm 18.59$ & $68.81 \pm 18.16$ & $68.77 \pm 18.10$ & $69.66 \pm 19.29$ & $65.92 \pm 18.05$ & $66.23 \pm 18.36$ \\
      Solution       & $80.32 \pm 23.43$ & $85.02 \pm 21.95$ & $86.44 \pm 20.42$ & $84.87 \pm 21.93$ & $85.36 \pm 21.34$ & $86.54 \pm 20.25$ & $84.45 \pm 21.53$ & $83.74 \pm 21.96$ & $82.61 \pm 21.82$ & $82.91 \pm 21.66$ & $84.00 \pm 21.58$ & $63.07 \pm 36.25$ & $65.50 \pm 36.25$ & $85.42 \pm 20.93$ & $86.11 \pm 19.92$ & $87.33 \pm 21.12$ & $86.42 \pm 20.69$ & $85.99 \pm 21.13$ & $86.96 \pm 20.06$ & $85.28 \pm 20.91$ & $82.92 \pm 21.88$ \\
      Politeness     & $95.15 \pm 14.30$ & $95.42 \pm 16.79$ & $95.82 \pm 15.61$ & $94.90 \pm 17.58$ & $94.69 \pm 16.73$ & $94.76 \pm 16.99$ & $94.96 \pm 16.45$ & $93.68 \pm 17.79$ & $93.22 \pm 17.77$ & $92.90 \pm 18.60$ & $94.00 \pm 17.76$ & $78.53 \pm 33.68$ & $79.88 \pm 33.10$ & $96.01 \pm 15.05$ & $95.46 \pm 15.41$ & $95.11 \pm 16.94$ & $95.13 \pm 16.67$ & $95.12 \pm 16.94$ & $94.78 \pm 17.00$ & $93.61 \pm 17.36$ & $93.16 \pm 17.82$ \\
      Urgency        & $92.09 \pm 19.92$ & $91.93 \pm 21.16$ & $93.73 \pm 18.81$ & $91.86 \pm 21.58$ & $92.57 \pm 20.50$ & $92.17 \pm 21.05$ & $92.21 \pm 20.17$ & $92.26 \pm 20.80$ & $92.61 \pm 20.05$ & $91.21 \pm 21.39$ & $91.60 \pm 22.13$ & $74.53 \pm 37.67$ & $73.78 \pm 38.79$ & $93.02 \pm 19.73$ & $92.96 \pm 19.26$ & $92.82 \pm 20.64$ & $93.16 \pm 20.05$ & $92.41 \pm 20.48$ & $93.63 \pm 19.18$ & $92.18 \pm 20.35$ & $90.64 \pm 21.29$ \\
      Repetition     & $60.83 \pm 34.82$ & $61.83 \pm 35.62$ & $61.52 \pm 35.79$ & $63.04 \pm 35.82$ & $61.57 \pm 35.38$ & $61.84 \pm 35.32$ & $60.43 \pm 35.54$ & $59.83 \pm 35.95$ & $60.34 \pm 36.14$ & $61.64 \pm 35.75$ & $60.47 \pm 35.78$ & $42.91 \pm 39.21$ & $42.15 \pm 39.13$ & $63.61 \pm 35.37$ & $61.23 \pm 34.90$ & $65.60 \pm 34.80$ & $63.84 \pm 34.79$ & $61.84 \pm 35.98$ & $65.85 \pm 35.12$ & $62.49 \pm 35.66$ & $60.14 \pm 35.74$ \\
      Disfluency     & $67.91 \pm 19.69$ & $68.20 \pm 20.99$ & $70.23 \pm 20.40$ & $68.16 \pm 21.48$ & $69.37 \pm 20.17$ & $68.64 \pm 20.93$ & $67.40 \pm 20.60$ & $69.17 \pm 20.55$ & $68.35 \pm 20.61$ & $67.96 \pm 21.41$ & $67.96 \pm 20.62$ & $51.44 \pm 29.65$ & $52.93 \pm 30.07$ & $69.99 \pm 20.16$ & $69.42 \pm 19.39$ & $70.48 \pm 20.81$ & $69.66 \pm 20.31$ & $69.43 \pm 20.77$ & $70.65 \pm 20.91$ & $67.52 \pm 20.53$ & $67.24 \pm 20.77$ \\
      Length         & $87.00 \pm 15.88$ & $86.77 \pm 18.63$ & $87.82 \pm 17.63$ & $86.12 \pm 19.01$ & $86.60 \pm 18.12$ & $86.83 \pm 18.28$ & $85.63 \pm 18.21$ & $85.77 \pm 18.47$ & $85.94 \pm 17.99$ & $85.32 \pm 18.47$ & $85.48 \pm 18.66$ & $69.32 \pm 32.26$ & $71.16 \pm 31.56$ & $87.65 \pm 17.57$ & $87.00 \pm 17.46$ & $87.67 \pm 18.57$ & $87.16 \pm 18.13$ & $87.38 \pm 18.17$ & $87.81 \pm 17.98$ & $85.72 \pm 18.32$ & $85.01 \pm 18.63$ \\
      Language       & $82.51 \pm 16.27$ & $83.30 \pm 18.04$ & $84.56 \pm 17.02$ & $83.27 \pm 17.91$ & $83.37 \pm 17.30$ & $82.93 \pm 17.60$ & $82.10 \pm 17.32$ & $82.81 \pm 17.64$ & $82.91 \pm 17.33$ & $82.75 \pm 17.92$ & $82.96 \pm 17.73$ & $63.13 \pm 31.92$ & $65.01 \pm 31.87$ & $84.60 \pm 16.64$ & $83.92 \pm 16.61$ & $84.46 \pm 17.76$ & $83.10 \pm 17.70$ & $83.72 \pm 17.92$ & $84.40 \pm 17.30$ & $83.19 \pm 17.73$ & $81.45 \pm 17.97$ \\
      Entity         & $50.66 \pm 15.22$ & $52.03 \pm 15.74$ & $54.04 \pm 17.11$ & $56.52 \pm 17.08$ & $47.02 \pm 15.89$ & $48.82 \pm 16.54$ & $48.73 \pm 17.30$ & $60.34 \pm 18.33$ & $67.39 \pm 18.63$ & $70.96 \pm 18.33$ & $60.00 \pm 18.71$ & $34.32 \pm 26.54$ & $32.37 \pm 23.72$ & $46.29 \pm 14.78$ & $49.57 \pm 16.48$ & $44.64 \pm 16.16$ & $54.48 \pm 16.76$ & $53.62 \pm 16.95$ & $48.91 \pm 17.51$ & $63.07 \pm 18.76$ & $52.19 \pm 17.63$ \\
      \midrule
    \textbf{Score} & 80.52 & 81.35 & 82.59 & 81.72 & 80.88 & 81.25 & 80.31 & 81.21 & 81.37 & 81.30 & 80.85 & 62.94 & 64.05 & 82.07 & 81.66 & 81.99 & 82.16 & 81.83 & 82.13 & 81.47 & \textbf{--} \\
    
      \midrule
    \textbf{LLM Judge Score} 
    & $2.07 \pm 1.29$  
    & $4.04 \pm 0.97$  
    & $4.79 \pm 0.32$  
    & $4.87 \pm 0.25$  
    & $4.68 \pm 0.39$  
    & $4.62 \pm 0.50$  
    & $4.85 \pm 0.29$  
    & $4.83 \pm 0.27$  
    & $4.72 \pm 0.34$  
    & $4.81 \pm 0.23$  
    & $4.71 \pm 0.37$  
    & $3.87 \pm 1.56$  
    & $3.96 \pm 1.57$  
    & $4.71 \pm 0.36$  
    & $4.85 \pm 0.25$  
    & $4.72 \pm 0.36$  
    & $4.78 \pm 0.31$  
    & $4.78 \pm 0.29$  
    & $4.74 \pm 0.33$  
    & $4.79 \pm 0.30$  
    & $4.73$ \\

    \textbf{Compression Ratio} & 
    0.135 & 0.064 & 0.07 & 0.059 & 0.045 & 0.05 & 0.041 & 0.053 & 0.062 & 0.068 & 0.056 & 0.025 & 0.024 & 0.045 & 0.047 & 0.042 & 0.056 & 0.064 & 0.06 & 0.056 & \textbf{0.056} \\
    
    \textbf{Compression Factor} & 
    10.98 & 18.83 & 17.23 & 20.75 & 27.44 & 25.29 & 31.2 & 22.86 & 19.05 & 17.29 & 21.87 & 62.01 & 60.78 & 26.37 & 27.73 & 29.19 & 20.84 & 17.68 & 20.13 & 21.25 & \textbf{25.94} \\

      \bottomrule

    \end{tabular}%
    }
  \caption{Detailed evaluation results for all 20 LLMs, showing mean and standard deviation.}
  \label{table:main_results_with_deviation}
\end{table*}

\subsection{Analysis with Alternative Divergence Metrics}
\label{subsec:alternate_metrics}
To ensure that our findings are not an artifact of our chosen divergence metric (JSD), we re-calculated the fidelity gap using three alternative metrics: Wasserstein Distance, Total Variation Distance (TVD), Chi-Square test and Kullback-Leibler (KL) Divergence. As shown in Table~\ref{table:model_performance_tvd_was} and Table~\ref{table:model_performance_kl_chi}, the relative model rankings and the identification of the most challenging bias dimensions (e.g., Temporal Sequence, Entity Type) remain highly consistent across all metrics. This consistency demonstrates the robustness of our core findings.

\begin{table*}[!hbt]
  \centering
  \small
  \renewcommand{\arraystretch}{1.2}
  \resizebox{\textwidth}{!}{%
    \begin{tabular}{@{}l *{21}{>{\centering\arraybackslash}p{1.2cm}} @{}}
      \toprule
      \textbf{Metric / Bias} 
        & \rotatebox{90}{\texttt{llama-3.2-1b}} 
        & \rotatebox{90}{\texttt{llama-3.2-3b}} 
        & \rotatebox{90}{\texttt{llama-3.3-70b}} 
        & \rotatebox{90}{\texttt{llama-4-maverick}} 
        & \rotatebox{90}{\texttt{nova-micro}} 
        & \rotatebox{90}{\texttt{nova-lite}} 
        & \rotatebox{90}{\texttt{nova-pro}} 
        & \rotatebox{90}{\texttt{claude-3.5-haiku}} 
        & \rotatebox{90}{\texttt{claude-3.7-sonnet}} 
        & \rotatebox{90}{\texttt{claude-4-sonnet}} 
        & \rotatebox{90}{\texttt{deepseek-r1}} 
        & \rotatebox{90}{\texttt{gemini-2.0-flash}} 
        & \rotatebox{90}{\texttt{gemini-2.0-flash-lite}} 
        & \rotatebox{90}{\texttt{gpt-4o-mini}} 
        & \rotatebox{90}{\texttt{gpt-4o}} 
        & \rotatebox{90}{\texttt{gpt-4.1-nano}} 
        & \rotatebox{90}{\texttt{gpt-4.1-mini}} 
        & \rotatebox{90}{\texttt{gpt-4.1}} 
        & \rotatebox{90}{\texttt{o3-mini}} 
        & \rotatebox{90}{\texttt{o4-mini}}
        & \rotatebox{90}
        {\textbf{Average}}\\
      \midrule
      \multicolumn{21}{@{}l}{\textit{\textbf{Wasserstein Distance}}} \\Position       & 0.335 & 0.281 & 0.256 & 0.265 & 0.277 & 0.265 & 0.293 & 0.263 & 0.279 & 0.270 & 0.272 & 0.450 & 0.453 & 0.284 & 0.273 & 0.237 & 0.246 & 0.262 & 0.263 & 0.276 & 0.291 \\
      Speaker        & 0.138 & 0.138 & 0.128 & 0.129 & 0.146 & 0.135 & 0.137 & 0.125 & 0.119 & 0.115 & 0.131 & 0.199 & 0.202 & 0.120 & 0.130 & 0.131 & 0.124 & 0.124 & 0.132 & 0.131 & 0.137 \\
      Sentiment      & 0.455 & 0.441 & 0.428 & 0.455 & 0.428 & 0.441 & 0.444 & 0.482 & 0.520 & 0.541 & 0.494 & 0.526 & 0.527 & 0.425 & 0.437 & 0.406 & 0.443 & 0.441 & 0.420 & 0.491 & 0.461 \\
      Topic          & 1.164 & 1.095 & 1.030 & 1.067 & 1.083 & 1.102 & 1.100 & 1.125 & 1.145 & 1.134 & 1.156 & 1.777 & 1.778 & 1.028 & 1.055 & 1.062 & 1.080 & 1.065 & 1.013 & 1.072 & 1.167 \\
      Agent Action & 1.600 & 1.575 & 1.564 & 1.576 & 1.610 & 1.600 & 1.607 & 1.592 & 1.624 & 1.618 & 1.617 & 1.671 & 1.672 & 1.575 & 1.586 & 1.578 & 1.591 & 1.588 & 1.608 & 1.616 & 1.594 \\
      Solution       & 0.698 & 0.590 & 0.561 & 0.571 & 0.544 & 0.548 & 0.571 & 0.568 & 0.643 & 0.617 & 0.581 & 0.939 & 0.910 & 0.552 & 0.546 & 0.479 & 0.534 & 0.509 & 0.523 & 0.542 & 0.588 \\
      Repetition     & 0.365 & 0.370 & 0.345 & 0.338 & 0.354 & 0.362 & 0.358 & 0.339 & 0.359 & 0.354 & 0.357 & 0.382 & 0.386 & 0.340 & 0.359 & 0.337 & 0.338 & 0.329 & 0.321 & 0.351 & 0.352 \\
      Disfluency     & 0.707 & 0.720 & 0.688 & 0.689 & 0.692 & 0.706 & 0.711 & 0.701 & 0.709 & 0.704 & 0.719 & 0.866 & 0.840 & 0.680 & 0.692 & 0.686 & 0.698 & 0.698 & 0.697 & 0.715 & 0.715 \\
      Politeness     & 0.218 & 0.226 & 0.217 & 0.217 & 0.223 & 0.221 & 0.222 & 0.205 & 0.202 & 0.195 & 0.214 & 0.259 & 0.260 & 0.214 & 0.214 & 0.203 & 0.209 & 0.210 & 0.199 & 0.213 & 0.218 \\
      Urgency        & 0.160 & 0.152 & 0.149 & 0.151 & 0.156 & 0.150 & 0.155 & 0.159 & 0.161 & 0.165 & 0.167 & 0.206 & 0.198 & 0.151 & 0.155 & 0.151 & 0.156 & 0.155 & 0.145 & 0.153 & 0.160 \\
      Length         & 0.195 & 0.190 & 0.180 & 0.188 & 0.193 & 0.185 & 0.192 & 0.188 & 0.189 & 0.182 & 0.189 & 0.283 & 0.287 & 0.181 & 0.187 & 0.181 & 0.180 & 0.182 & 0.181 & 0.190 & 0.193 \\
      Language       & 0.763 & 0.747 & 0.705 & 0.732 & 0.743 & 0.713 & 0.750 & 0.720 & 0.743 & 0.734 & 0.764 & 1.085 & 1.065 & 0.679 & 0.707 & 0.686 & 0.706 & 0.681 & 0.694 & 0.750 & 0.748 \\
      Entity Types   & 1.151 & 1.112 & 1.057 & 1.022 & 1.228 & 1.216 & 1.229 & 0.951 & 0.827 & 0.796 & 0.963 & 1.184 & 1.238 & 1.220 & 1.166 & 1.303 & 1.074 & 1.081 & 1.270 & 0.955 & 1.091 \\
      Emotion        & 1.065 & 1.186 & 1.166 & 1.129 & 1.181 & 1.163 & 1.148 & 1.178 & 1.111 & 1.142 & 1.144 & 1.207 & 1.137 & 1.198 & 1.151 & 1.132 & 1.115 & 1.059 & 1.061 & 0.991 & 1.128 \\
      Order          & 1.515 & 1.185 & 1.073 & 1.029 & 0.986 & 1.033 & 0.987 & 1.040 & 0.949 & 1.012 & 1.079 & 1.675 & 1.740 & 0.970 & 0.972 & 1.062 & 0.977 & 0.991 & 0.996 & 0.976 & 1.097 \\
      \midrule
      \textbf{Average} & 0.675 & 0.634 & 0.602 & 0.604 & 0.616 & 0.616 & 0.621 & 0.609 & 0.611 & 0.611 & 0.620 & 0.808 & 0.806 & 0.601 & 0.615 & 0.609 & 0.610 & 0.607 & 0.608 & 0.614 & -- \\
      \midrule
      \multicolumn{21}{@{}l}{\textit{\textbf{Total Variation Distance}}} 
      \\
      Position       & 0.173 & 0.151 & 0.139 & 0.143 & 0.145 & 0.140 & 0.151 & 0.141 & 0.145 & 0.142 & 0.144 & 0.248 & 0.248 & 0.146 & 0.144 & 0.132 & 0.135 & 0.140 & 0.137 & 0.143 & 0.153 \\
      Speaker        & 0.138 & 0.138 & 0.128 & 0.129 & 0.146 & 0.135 & 0.137 & 0.125 & 0.119 & 0.115 & 0.131 & 0.199 & 0.202 & 0.120 & 0.130 & 0.131 & 0.124 & 0.124 & 0.132 & 0.131 & 0.137 \\
      Sentiment      & 0.208 & 0.212 & 0.203 & 0.209 & 0.202 & 0.207 & 0.206 & 0.218 & 0.225 & 0.233 & 0.224 & 0.242 & 0.238 & 0.205 & 0.206 & 0.195 & 0.207 & 0.205 & 0.194 & 0.215 & 0.211 \\
      Topic          & 0.233 & 0.221 & 0.213 & 0.216 & 0.221 & 0.220 & 0.225 & 0.225 & 0.228 & 0.227 & 0.233 & 0.336 & 0.328 & 0.210 & 0.221 & 0.212 & 0.215 & 0.213 & 0.205 & 0.218 & 0.231 \\
      Agent Action & 0.460 & 0.458 & 0.454 & 0.461 & 0.464 & 0.465 & 0.468 & 0.468 & 0.479 & 0.480 & 0.477 & 0.515 & 0.508 & 0.453 & 0.460 & 0.455 & 0.464 & 0.460 & 0.461 & 0.474 & 0.466 \\
      Solution       & 0.163 & 0.133 & 0.126 & 0.128 & 0.123 & 0.121 & 0.133 & 0.128 & 0.141 & 0.139 & 0.133 & 0.210 & 0.201 & 0.126 & 0.125 & 0.109 & 0.118 & 0.115 & 0.114 & 0.120 & 0.132 \\
      Repetition     & 0.236 & 0.239 & 0.224 & 0.215 & 0.227 & 0.234 & 0.230 & 0.219 & 0.230 & 0.225 & 0.229 & 0.250 & 0.251 & 0.214 & 0.228 & 0.212 & 0.214 & 0.217 & 0.206 & 0.226 & 0.226 \\
      Disfluency     & 0.173 & 0.169 & 0.165 & 0.166 & 0.165 & 0.166 & 0.172 & 0.169 & 0.169 & 0.170 & 0.171 & 0.215 & 0.212 & 0.163 & 0.166 & 0.159 & 0.165 & 0.166 & 0.162 & 0.171 & 0.171 \\
      Politeness     & 0.211 & 0.218 & 0.211 & 0.211 & 0.217 & 0.215 & 0.216 & 0.200 & 0.197 & 0.190 & 0.208 & 0.253 & 0.253 & 0.208 & 0.207 & 0.197 & 0.203 & 0.204 & 0.194 & 0.208 & 0.211 \\
      Urgency        & 0.125 & 0.120 & 0.119 & 0.121 & 0.123 & 0.119 & 0.124 & 0.125 & 0.126 & 0.129 & 0.131 & 0.167 & 0.161 & 0.119 & 0.122 & 0.118 & 0.122 & 0.122 & 0.114 & 0.122 & 0.125 \\
      Length         & 0.111 & 0.107 & 0.102 & 0.107 & 0.109 & 0.105 & 0.109 & 0.105 & 0.107 & 0.104 & 0.109 & 0.168 & 0.170 & 0.102 & 0.105 & 0.100 & 0.102 & 0.103 & 0.102 & 0.106 & 0.110 \\
      Language       & 0.193 & 0.186 & 0.176 & 0.185 & 0.187 & 0.182 & 0.190 & 0.180 & 0.181 & 0.180 & 0.187 & 0.257 & 0.260 & 0.172 & 0.178 & 0.172 & 0.175 & 0.175 & 0.172 & 0.188 & 0.182 \\
      Entity Types   & 0.403 & 0.386 & 0.367 & 0.351 & 0.420 & 0.411 & 0.413 & 0.320 & 0.287 & 0.272 & 0.328 & 0.404 & 0.432 & 0.422 & 0.404 & 0.440 & 0.368 & 0.374 & 0.413 & 0.316 & 0.373 \\
      Emotion        & 0.283 & 0.346 & 0.335 & 0.316 & 0.338 & 0.332 & 0.320 & 0.319 & 0.284 & 0.290 & 0.310 & 0.269 & 0.256 & 0.357 & 0.331 & 0.331 & 0.313 & 0.305 & 0.296 & 0.265 & 0.310 \\
      Order          & 0.758 & 0.712 & 0.683 & 0.705 & 0.740 & 0.726 & 0.745 & 0.717 & 0.710 & 0.697 & 0.696 & 0.816 & 0.815 & 0.740 & 0.745 & 0.702 & 0.666 & 0.676 & 0.704 & 0.696 & 0.722 \\
      \midrule
      \textbf{Average} & 0.278 & 0.268 & 0.257 & 0.257 & 0.265 & 0.262 & 0.266 & 0.258 & 0.260 & 0.259 & 0.261 & 0.294 & 0.292 & 0.257 & 0.264 & 0.251 & 0.255 & 0.256 & 0.254 & 0.259 & -- \\
      \bottomrule
    \end{tabular}%
  }
  \caption{Model performance using Wasserstein Distance and Total Variation Distance (TVD) as the fidelity gap metric. The overall performance trends are consistent with those observed using JS Divergence.}
  \label{table:model_performance_tvd_was}
\end{table*}

\begin{table*}[!hbt]
  \centering
  \small
  \renewcommand{\arraystretch}{1.2}
  \resizebox{\textwidth}{!}{%
    \begin{tabular}{@{}l *{21}{>{\centering\arraybackslash}p{1.2cm}} @{}}
      \toprule
      \textbf{Metric / Bias} 
        & \rotatebox{90}{\texttt{llama-3.2-1b}} 
        & \rotatebox{90}{\texttt{llama-3.2-3b}} 
        & \rotatebox{90}{\texttt{llama-3.3-70b}} 
        & \rotatebox{90}{\texttt{llama-4-maverick}} 
        & \rotatebox{90}{\texttt{nova-micro}} 
        & \rotatebox{90}{\texttt{nova-lite}} 
        & \rotatebox{90}{\texttt{nova-pro}} 
        & \rotatebox{90}{\texttt{claude-3.5-haiku}} 
        & \rotatebox{90}{\texttt{claude-3.7-sonnet}} 
        & \rotatebox{90}{\texttt{claude-4-sonnet}} 
        & \rotatebox{90}{\texttt{deepseek-r1}} 
        & \rotatebox{90}{\texttt{gemini-2.0-flash}} 
        & \rotatebox{90}{\texttt{gemini-2.0-flash-lite}} 
        & \rotatebox{90}{\texttt{gpt-4o-mini}} 
        & \rotatebox{90}{\texttt{gpt-4o}} 
        & \rotatebox{90}{\texttt{gpt-4.1-nano}} 
        & \rotatebox{90}{\texttt{gpt-4.1-mini}} 
        & \rotatebox{90}{\texttt{gpt-4.1}} 
        & \rotatebox{90}{\texttt{o3-mini}} 
        & \rotatebox{90}{\texttt{o4-mini}}
        & \rotatebox{90}
        {\textbf{Average}}\\
      \midrule
      \multicolumn{21}{@{}l}{\textit{\textbf{ KL-Divergence}}} \\
            Position       & 0.156 & 0.078 & 0.066 & 0.087 & 0.074 & 0.069 & 0.086 & 0.074 & 0.077 & 0.076 & 0.094 & 2.754 & 2.660 & 0.079 & 0.074 & 0.060 & 0.061 & 0.077 & 0.062 & 0.085 & 0.345 \\
      Speaker        & 0.072 & 0.072 & 0.058 & 0.063 & 0.080 & 0.069 & 0.070 & 0.059 & 0.053 & 0.046 & 0.063 & 1.757 & 1.650 & 0.054 & 0.062 & 0.067 & 0.055 & 0.056 & 0.065 & 0.061 & 0.240 \\
      Sentiment      & 0.261 & 0.214 & 0.194 & 0.212 & 0.228 & 0.213 & 0.235 & 0.245 & 0.274 & 0.285 & 0.262 & 1.318 & 1.372 & 0.186 & 0.219 & 0.197 & 0.213 & 0.209 & 0.206 & 0.244 & 0.348 \\
      Topic          & 1.425 & 1.055 & 0.974 & 1.013 & 1.184 & 1.088 & 1.249 & 1.278 & 1.476 & 1.566 & 1.365 & 4.229 & 3.965 & 0.867 & 1.052 & 1.037 & 1.069 & 1.048 & 1.079 & 1.335 & 1.475 \\
      Agent Action & 5.685 & 5.397 & 5.076 & 5.323 & 5.686 & 5.615 & 5.937 & 5.532 & 5.775 & 5.489 & 6.073 & 7.543 & 7.355 & 5.074 & 5.465 & 4.961 & 5.252 & 5.364 & 5.079 & 5.671 & 5.617 \\
      Solution       & 1.803 & 1.075 & 1.067 & 1.133 & 1.057 & 1.024 & 1.230 & 1.163 & 1.361 & 1.368 & 1.242 & 2.942 & 2.719 & 1.065 & 1.004 & 0.836 & 1.014 & 0.965 & 0.932 & 1.008 & 1.304 \\
      Repetition     & 3.995 & 4.013 & 3.671 & 3.461 & 3.769 & 3.928 & 3.861 & 3.745 & 3.891 & 3.781 & 3.821 & 4.353 & 4.421 & 3.423 & 3.964 & 3.410 & 3.484 & 3.596 & 3.264 & 3.662 & 3.749 \\
      Disfluency     & 2.247 & 2.127 & 2.018 & 2.066 & 2.030 & 2.186 & 2.229 & 2.103 & 2.149 & 2.188 & 2.201 & 3.173 & 3.198 & 2.039 & 2.078 & 1.988 & 2.088 & 2.107 & 2.029 & 2.263 & 2.255 \\
      Politeness     & 0.210 & 0.178 & 0.183 & 0.193 & 0.178 & 0.199 & 0.184 & 0.179 & 0.162 & 0.175 & 0.190 & 1.608 & 1.450 & 0.165 & 0.190 & 0.142 & 0.163 & 0.147 & 0.165 & 0.190 & 0.289 \\
      Urgency        & 0.389 & 0.357 & 0.329 & 0.331 & 0.356 & 0.338 & 0.378 & 0.395 & 0.357 & 0.469 & 0.392 & 1.372 & 1.233 & 0.319 & 0.339 & 0.325 & 0.350 & 0.393 & 0.323 & 0.412 & 0.449 \\
      Length         & 0.323 & 0.270 & 0.251 & 0.283 & 0.286 & 0.274 & 0.310 & 0.295 & 0.303 & 0.299 & 0.305 & 1.801 & 1.760 & 0.260 & 0.273 & 0.239 & 0.250 & 0.231 & 0.250 & 0.307 & 0.418 \\
      Language       & 0.698 & 0.603 & 0.548 & 0.592 & 0.606 & 0.567 & 0.630 & 0.602 & 0.587 & 0.581 & 0.634 & 2.278 & 2.348 & 0.549 & 0.545 & 0.583 & 0.574 & 0.573 & 0.550 & 0.584 & 0.696 \\
      Entity Types   & 6.594 & 6.132 & 5.616 & 5.040 & 7.178 & 6.792 & 6.893 & 4.185 & 3.165 & 2.691 & 4.322 & 6.570 & 7.577 & 7.254 & 6.628 & 7.476 & 5.511 & 5.697 & 6.701 & 3.830 & 5.859 \\
      Emotion        & 0.362 & 0.456 & 0.434 & 0.406 & 0.440 & 0.432 & 0.414 & 0.413 & 0.361 & 0.372 & 0.407 & 0.828 & 0.674 & 0.470 & 0.430 & 0.436 & 0.406 & 0.391 & 0.383 & 0.335 & 0.427 \\
      Order          & 1.609 & 1.476 & 1.324 & 1.534 & 1.550 & 1.545 & 1.592 & 1.445 & 1.368 & 1.437 & 1.480 & 6.550 & 6.881 & 1.430 & 1.569 & 1.451 & 1.271 & 1.306 & 1.399 & 1.389 & 1.950 \\
      \midrule
      \textbf{Average} & 1.655 & 1.567 & 1.454 & 1.503 & 1.647 & 1.556 & 1.642 & 1.540 & 1.559 & 1.575 & 1.603 & 3.211 & 3.149 & 1.522 & 1.553 & 1.474 & 1.461 & 1.450 & 1.482 & 1.551 & -- \\
      \midrule
      \multicolumn{21}{@{}l}{\textit{\textbf{Chi-Squared Value}}} 
      \\
            Position       & 0.218 & 0.156 & 0.134 & 0.140 & 0.145 & 0.136 & 0.155 & 0.134 & 0.141 & 0.137 & 0.144 & 0.597 & 0.579 & 0.144 & 0.144 & 0.118 & 0.123 & 0.134 & 0.127 & 0.139 & 0.180 \\
      Speaker        & 0.116 & 0.114 & 0.099 & 0.103 & 0.128 & 0.112 & 0.114 & 0.098 & 0.090 & 0.081 & 0.105 & 0.259 & 0.263 & 0.091 & 0.103 & 0.108 & 0.093 & 0.094 & 0.107 & 0.103 & 0.117 \\
      Sentiment      & 0.592 & 0.577 & 0.521 & 0.559 & 0.549 & 0.558 & 0.566 & 0.609 & 0.665 & 0.700 & 0.656 & 1.708 & 1.785 & 0.536 & 0.552 & 0.501 & 0.550 & 0.536 & 0.498 & 0.605 & 0.663 \\
      Topic          & 0.437 & 0.380 & 0.345 & 0.364 & 0.375 & 0.370 & 0.398 & 0.388 & 0.403 & 0.403 & 0.416 & 2.749 & 2.338 & 0.334 & 0.373 & 0.337 & 0.351 & 0.344 & 0.317 & 0.373 & 0.586 \\
      Agent Action & 1.059 & 1.034 & 1.022 & 1.049 & 1.062 & 1.066 & 1.084 & 1.086 & 1.137 & 1.151 & 1.134 & 1.537 & 1.631 & 1.011 & 1.055 & 1.023 & 1.062 & 1.043 & 1.031 & 1.108 & 1.088 \\
      Solution       & 0.281 & 0.168 & 0.163 & 0.163 & 0.158 & 0.160 & 0.187 & 0.169 & 0.195 & 0.191 & 0.194 & 0.657 & 0.595 & 0.160 & 0.147 & 0.120 & 0.154 & 0.134 & 0.122 & 0.146 & 0.203 \\
      Repetition     & 0.611 & 0.738 & 0.580 & 0.581 & 0.609 & 0.642 & 0.561 & 0.560 & 0.634 & 0.661 & 0.601 & 0.708 & 0.727 & 0.510 & 0.605 & 0.546 & 0.515 & 0.539 & 0.522 & 0.590 & 0.597 \\
      Disfluency     & 0.291 & 0.272 & 0.259 & 0.263 & 0.257 & 0.265 & 0.298 & 0.273 & 0.278 & 0.277 & 0.277 & 0.517 & 0.455 & 0.249 & 0.263 & 0.240 & 0.263 & 0.273 & 0.245 & 0.279 & 0.288 \\
      Politeness     & 0.391 & 0.411 & 0.374 & 0.381 & 0.402 & 0.387 & 0.403 & 0.346 & 0.344 & 0.323 & 0.379 & 0.692 & 0.681 & 0.364 & 0.372 & 0.335 & 0.348 & 0.353 & 0.326 & 0.371 & 0.394 \\
      Urgency        & 0.297 & 0.264 & 0.262 & 0.257 & 0.268 & 0.258 & 0.276 & 0.288 & 0.287 & 0.300 & 0.309 & 0.746 & 0.653 & 0.249 & 0.259 & 0.242 & 0.256 & 0.258 & 0.234 & 0.271 & 0.315 \\
      Length         & 0.113 & 0.104 & 0.092 & 0.102 & 0.103 & 0.101 & 0.105 & 0.097 & 0.102 & 0.098 & 0.106 & 0.381 & 0.493 & 0.093 & 0.097 & 0.086 & 0.093 & 0.095 & 0.091 & 0.103 & 0.128 \\
      Language       & 0.369 & 0.336 & 0.302 & 0.331 & 0.329 & 0.319 & 0.353 & 0.317 & 0.326 & 0.323 & 0.395 & 1.220 & 1.331 & 0.288 & 0.314 & 0.274 & 0.300 & 0.297 & 0.275 & 0.338 & 0.409 \\
      Entity Types   & 54.6M & 20.5M & 34.5M & 33.0M & 26.6M & 23.5M & 25.0M & 24.2M & 24.5M & 25.8M & 23.3M & 32.1M & 30.8M & 22.0M & 20.7M & 25.3M & 22.6M & 19.4M & 26.9M & 24.0M & 27.4M \\
      Emotion        & 435.6M & 547.9M & 501.9M & 459.8M & 532.8M & 517.3M & 474.6M & 504.0M & 426.3M & 479.0M & 479.1M & 848.1M & 744.5M & 541.3M & 486.6M & 507.5M & 484.1M & 433.5M & 441.1M & 384.9M & 512.5M \\
      Order          & 2737.2M & 2669.9M & 2636.1M & 3101.9M & 3679.5M & 3295.7M & 3793.3M & 3050.9M & 3353.2M & 2958.5M & 2789.9M & 5403.4M & 5292.7M & 3620.3M & 3824.0M & 2779.0M & 2661.8M & 2763.7M & 3068.8M & 3109.2M & 3298.9M \\
      \midrule
      \textbf{Average} & 182.6 & 178.8 & 172.8 & 202.6 & 239.0 & 214.0 & 246.0 & 198.9 & 218.0 & 193.0 & 181.5 & 352.9 & 345.2 & 236.6 & 249.6 & 181.1 & 173.7 & 180.7 & 200.0 & 203.0 & -- \\
      \bottomrule
    \end{tabular}%
  }
  \caption{Model performance using KL-Divergence and Chi-Squared values. The relative model and dimension rankings, however, remain stable.}
  \label{table:model_performance_kl_chi}
\end{table*}

\subsection{Impact of Transcript Length on Bias}
To investigate how conversational complexity affects summarization bias, we segmented our dataset into three buckets based on transcript token count: short (<3000 tokens), medium (3000-6000 tokens), and long (>6000 tokens). Tables \ref{table:model_performance_1}, \ref{table:model_performance_2}, and \ref{table:model_performance_3} show model performance for each bucket. While performance is generally stable, we observe a slight trend where bias (JSD) increases and coverage decreases as transcripts become longer and more complex, particularly for dimensions like \textit{Entity Type} and \textit{Topic}. This suggests that models struggle more with information retention and faithful representation as the input context grows.

\begin{table*}[!h]
  \centering
  \small
  \renewcommand{\arraystretch}{1.2}
  \resizebox{\textwidth}{!}{%
    \begin{tabular}{@{}l *{21}{>{\centering\arraybackslash}p{1.2cm}} @{}}
      \toprule
      \textbf{Metric / Bias} 
        & \rotatebox{90}{\texttt{llama-3.2-1b}} 
        & \rotatebox{90}{\texttt{llama-3.2-3b}} 
        & \rotatebox{90}{\texttt{llama-3.3-70b}} 
        & \rotatebox{90}{\texttt{llama-4-maverick}} 
        & \rotatebox{90}{\texttt{nova-micro}} 
        & \rotatebox{90}{\texttt{nova-lite}} 
        & \rotatebox{90}{\texttt{nova-pro}} 
        & \rotatebox{90}{\texttt{claude-3.5-haiku}} 
        & \rotatebox{90}{\texttt{claude-3.7-sonnet}} 
        & \rotatebox{90}{\texttt{claude-4-sonnet}} 
        & \rotatebox{90}{\texttt{deepseek-r1}} 
        & \rotatebox{90}{\texttt{gemini-2.0-flash}} 
        & \rotatebox{90}{\texttt{gemini-2.0-flash-lite}} 
        & \rotatebox{90}{\texttt{gpt-4o-mini}} 
        & \rotatebox{90}{\texttt{gpt-4o}} 
        & \rotatebox{90}{\texttt{gpt-4.1-nano}} 
        & \rotatebox{90}{\texttt{gpt-4.1-mini}} 
        & \rotatebox{90}{\texttt{gpt-4.1}} 
        & \rotatebox{90}{\texttt{o3-mini}} 
        & \rotatebox{90}{\texttt{o4-mini}}
        & \rotatebox{90}
        {\textbf{Average}}\\
      \midrule
      \multicolumn{21}{@{}l}{\textit{\textbf{JS Divergence (JSD)}}} \\
      Position       & 0.026 & 0.019 & 0.017 & 0.020 & 0.018 & 0.018 & 0.021 & 0.017 & 0.019 & 0.017 & 0.018 & 0.143 & 0.132 & 0.018 & 0.016 & 0.016 & 0.016 & 0.018 & 0.016 & 0.018 & 0.034 \\
      Speaker        & 0.010 & 0.011 & 0.009 & 0.009 & 0.012 & 0.010 & 0.010 & 0.008 & 0.008 & 0.008 & 0.011 & 0.076 & 0.065 & 0.008 & 0.010 & 0.009 & 0.008 & 0.009 & 0.011 & 0.010 & 0.018 \\
      Sentiment      & 0.039 & 0.037 & 0.034 & 0.037 & 0.038 & 0.037 & 0.036 & 0.041 & 0.045 & 0.048 & 0.041 & 0.084 & 0.091 & 0.036 & 0.036 & 0.033 & 0.038 & 0.036 & 0.033 & 0.040 & 0.043 \\
      Topic          & 0.058 & 0.046 & 0.043 & 0.047 & 0.050 & 0.048 & 0.053 & 0.051 & 0.060 & 0.059 & 0.056 & 0.199 & 0.175 & 0.041 & 0.046 & 0.047 & 0.047 & 0.046 & 0.044 & 0.049 & 0.062 \\
      Agent Action   & 0.171 & 0.166 & 0.159 & 0.165 & 0.173 & 0.168 & 0.170 & 0.162 & 0.179 & 0.177 & 0.176 & 0.229 & 0.226 & 0.160 & 0.161 & 0.164 & 0.166 & 0.164 & 0.163 & 0.173 & 0.172 \\
      Solution       & 0.041 & 0.026 & 0.025 & 0.031 & 0.028 & 0.026 & 0.035 & 0.025 & 0.039 & 0.033 & 0.032 & 0.097 & 0.093 & 0.032 & 0.025 & 0.023 & 0.028 & 0.028 & 0.026 & 0.021 & 0.037 \\
      Politeness     & 0.031 & 0.029 & 0.028 & 0.028 & 0.032 & 0.030 & 0.030 & 0.026 & 0.026 & 0.027 & 0.027 & 0.081 & 0.075 & 0.028 & 0.029 & 0.024 & 0.027 & 0.026 & 0.025 & 0.028 & 0.032 \\
      Urgency        & 0.029 & 0.022 & 0.021 & 0.024 & 0.027 & 0.022 & 0.025 & 0.026 & 0.026 & 0.033 & 0.027 & 0.065 & 0.063 & 0.021 & 0.030 & 0.024 & 0.027 & 0.027 & 0.022 & 0.030 & 0.029 \\
      Order          & 0.398 & 0.337 & 0.320 & 0.355 & 0.380 & 0.377 & 0.394 & 0.344 & 0.361 & 0.331 & 0.341 & 0.663 & 0.655 & 0.382 & 0.379 & 0.348 & 0.319 & 0.329 & 0.348 & 0.341 & 0.380 \\
      Emotion        & 0.097 & 0.135 & 0.137 & 0.117 & 0.126 & 0.122 & 0.122 & 0.125 & 0.106 & 0.118 & 0.125 & 0.073 & 0.085 & 0.132 & 0.123 & 0.133 & 0.120 & 0.127 & 0.111 & 0.107 & 0.118 \\
      Repetition     & 0.079 & 0.092 & 0.081 & 0.064 & 0.089 & 0.097 & 0.081 & 0.077 & 0.081 & 0.081 & 0.090 & 0.094 & 0.119 & 0.075 & 0.086 & 0.075 & 0.080 & 0.074 & 0.072 & 0.081 & 0.085 \\
      Disfluency     & 0.060 & 0.058 & 0.054 & 0.056 & 0.057 & 0.058 & 0.059 & 0.054 & 0.056 & 0.057 & 0.060 & 0.107 & 0.101 & 0.052 & 0.057 & 0.054 & 0.055 & 0.055 & 0.057 & 0.058 & 0.062 \\
      Length         & 0.019 & 0.016 & 0.014 & 0.017 & 0.017 & 0.017 & 0.018 & 0.017 & 0.017 & 0.016 & 0.017 & 0.085 & 0.074 & 0.016 & 0.015 & 0.015 & 0.016 & 0.016 & 0.015 & 0.016 & 0.023 \\
      Language       & 0.036 & 0.033 & 0.030 & 0.033 & 0.034 & 0.033 & 0.033 & 0.030 & 0.030 & 0.032 & 0.033 & 0.112 & 0.114 & 0.030 & 0.029 & 0.029 & 0.032 & 0.032 & 0.026 & 0.033 & 0.039 \\
      Entity         & 0.165 & 0.140 & 0.125 & 0.125 & 0.164 & 0.165 & 0.157 & 0.103 & 0.084 & 0.078 & 0.111 & 0.158 & 0.179 & 0.170 & 0.141 & 0.176 & 0.135 & 0.143 & 0.158 & 0.104 & 0.136 \\
      \midrule
      \textbf{Average} & 0.085 & 0.076 & 0.072 & 0.076 & 0.080 & 0.080 & 0.081 & 0.072 & 0.076 & 0.075 & 0.076 & 0.168 & 0.165 & 0.076 & 0.075 & 0.074 & 0.074 & 0.073 & 0.071 & 0.073 & -- \\
      \midrule
      \multicolumn{21}{@{}l}{\textit{\textbf{Coverage}}} \\
      Position       & 99.32 & 97.87 & 98.00 & 96.27 & 97.87 & 97.87 & 97.47 & 97.87 & 97.60 & 97.47 & 97.17 & 62.07 & 65.53 & 97.73 & 97.87 & 97.87 & 98.00 & 97.87 & 98.00 & 97.60 & 92.11 \\
      Speaker        & 100.00 & 98.00 & 98.00 & 96.67 & 98.00 & 98.00 & 98.00 & 98.00 & 98.00 & 98.00 & 97.33 & 72.33 & 77.67 & 98.00 & 98.00 & 98.00 & 98.00 & 98.00 & 98.00 & 98.00 & 94.80 \\
      Sentiment      & 90.34 & 90.74 & 92.36 & 89.33 & 90.34 & 91.54 & 89.82 & 90.31 & 88.44 & 88.54 & 89.30 & 56.97 & 61.67 & 92.02 & 92.09 & 92.13 & 91.73 & 91.36 & 90.47 & 88.26 & 86.95 \\
      Topic          & 78.02 & 82.61 & 83.26 & 80.48 & 80.60 & 82.42 & 78.74 & 79.62 & 75.58 & 75.16 & 78.09 & 40.74 & 44.95 & 83.53 & 81.84 & 81.45 & 81.35 & 82.29 & 81.46 & 78.48 & 74.64 \\
      Agent Action   & 65.13 & 66.25 & 68.85 & 65.83 & 62.64 & 66.63 & 64.48 & 67.23 & 61.98 & 62.78 & 62.89 & 37.02 & 40.96 & 67.13 & 66.96 & 67.44 & 66.70 & 68.02 & 66.38 & 63.70 & 62.61 \\
      Solution       & 84.50 & 87.56 & 87.31 & 84.48 & 85.71 & 87.50 & 85.27 & 85.86 & 82.61 & 85.40 & 86.83 & 51.71 & 55.31 & 85.15 & 87.06 & 87.03 & 87.26 & 86.27 & 87.39 & 86.69 & 82.17 \\
      Repetition     & 58.40 & 63.16 & 63.47 & 63.83 & 62.02 & 62.68 & 61.78 & 60.33 & 61.47 & 62.32 & 59.78 & 30.01 & 30.92 & 62.56 & 61.96 & 67.81 & 65.10 & 63.04 & 65.10 & 62.44 & 58.05 \\
      Disfluency     & 67.98 & 68.17 & 71.33 & 68.52 & 68.86 & 68.36 & 67.70 & 69.94 & 69.57 & 69.22 & 67.81 & 39.89 & 43.54 & 69.93 & 70.07 & 69.94 & 70.44 & 69.56 & 70.34 & 68.34 & 64.62 \\
      Politeness     & 96.20 & 96.89 & 96.39 & 94.50 & 95.00 & 96.28 & 94.94 & 94.94 & 94.00 & 94.61 & 94.28 & 66.22 & 68.39 & 96.50 & 95.94 & 95.50 & 95.33 & 96.44 & 94.44 & 93.67 & 91.97 \\
      Urgency        & 89.16 & 89.37 & 92.28 & 87.58 & 90.27 & 90.27 & 90.72 & 90.27 & 90.27 & 88.14 & 88.48 & 58.05 & 58.72 & 89.93 & 90.72 & 90.72 & 91.50 & 89.71 & 92.73 & 89.37 & 85.94 \\
      Length         & 85.68 & 83.53 & 85.60 & 83.00 & 82.80 & 83.93 & 81.90 & 82.17 & 83.23 & 82.53 & 82.23 & 53.16 & 56.98 & 84.00 & 83.33 & 83.63 & 83.53 & 84.03 & 84.97 & 83.27 & 79.79 \\
      Language       & 86.15 & 84.95 & 86.22 & 84.65 & 85.16 & 84.69 & 83.88 & 85.19 & 85.90 & 85.48 & 85.21 & 49.94 & 53.00 & 84.97 & 86.11 & 85.77 & 83.82 & 85.50 & 87.55 & 85.49 & 81.56 \\
      Entity         & 54.26 & 56.76 & 60.66 & 61.31 & 52.31 & 53.19 & 54.82 & 65.65 & 73.43 & 75.20 & 64.26 & 27.05 & 26.57 & 50.98 & 56.88 & 49.97 & 59.02 & 56.97 & 53.57 & 67.27 & 56.39 \\
      \midrule
      \textbf{Average} & 79.86 & 81.06 & 82.42 & 80.29 & 79.48 & 81.01 & 79.38 & 80.22 & 79.53 & 79.46 & 79.09 & 48.38 & 50.68 & 80.60 & 81.24 & 81.41 & 81.16 & 81.06 & 81.32 & 80.49 & -- \\
      \midrule
    \textbf{LLM Judge Score} 
    & 2.08  
    & 3.99  
    & 4.82  
    & 4.88  
    & 4.68  
    & 4.62  
    & 4.88  
    & 4.81  
    & 4.69  
    & 4.81  
    & 4.72  
    & 3.27  
    & 3.35  
    & 4.71  
    & 4.87  
    & 4.74  
    & 4.78  
    & 4.80  
    & 4.74  
    & 4.79  
    & 4.47
    \\
      \textbf{Compression Ratio} & 0.214 & 0.109 & 0.116 & 0.097 & 0.078 & 0.081 & 0.071 & 0.091 & 0.103 & 0.113 & 0.089 & 0.037 & 0.036 & 0.076 & 0.087 & 0.071 & 0.087 & 0.092 & 0.095 & 0.087 & 0.091 \\
      \textbf{Compression Factor} & 6.42 & 10.22 & 9.50 & 11.94 & 15.23 & 15.08 & 16.90 & 11.97 & 10.69 & 9.72 & 13.15 & 54.43 & 53.66 & 15.00 & 14.12 & 16.07 & 12.87 & 12.14 & 11.88 & 12.92 & 16.48 \\
      \bottomrule
    \end{tabular}%
  }
  \caption{Model performance on short transcripts (<3000 tokens).}
  \label{table:model_performance_1}
\end{table*}

\begin{table*}[!h]
  \centering
  \small
  \renewcommand{\arraystretch}{1.2}
  \resizebox{\textwidth}{!}{%
    \begin{tabular}{@{}l *{21}{>{\centering\arraybackslash}p{1.2cm}} @{}}
      \toprule
      \textbf{Metric / Bias} 
        & \rotatebox{90}{\texttt{llama-3.2-1b}} 
        & \rotatebox{90}{\texttt{llama-3.2-3b}} 
        & \rotatebox{90}{\texttt{llama-3.3-70b}} 
        & \rotatebox{90}{\texttt{llama-4-maverick}} 
        & \rotatebox{90}{\texttt{nova-micro}} 
        & \rotatebox{90}{\texttt{nova-lite}} 
        & \rotatebox{90}{\texttt{nova-pro}} 
        & \rotatebox{90}{\texttt{claude-3.5-haiku}} 
        & \rotatebox{90}{\texttt{claude-3.7-sonnet}} 
        & \rotatebox{90}{\texttt{claude-4-sonnet}} 
        & \rotatebox{90}{\texttt{deepseek-r1}} 
        & \rotatebox{90}{\texttt{gemini-2.0-flash}} 
        & \rotatebox{90}{\texttt{gemini-2.0-flash-lite}} 
        & \rotatebox{90}{\texttt{gpt-4o-mini}} 
        & \rotatebox{90}{\texttt{gpt-4o}} 
        & \rotatebox{90}{\texttt{gpt-4.1-nano}} 
        & \rotatebox{90}{\texttt{gpt-4.1-mini}} 
        & \rotatebox{90}{\texttt{gpt-4.1}} 
        & \rotatebox{90}{\texttt{o3-mini}} 
        & \rotatebox{90}{\texttt{o4-mini}}
        & \rotatebox{90}
        {\textbf{Average}}\\
      \midrule
      \multicolumn{21}{@{}l}{\textit{\textbf{JS Divergence (JSD)}}} \\
      Position       & 0.023 & 0.017 & 0.015 & 0.015 & 0.018 & 0.017 & 0.018 & 0.017 & 0.016 & 0.016 & 0.016 & 0.127 & 0.132 & 0.017 & 0.016 & 0.014 & 0.015 & 0.016 & 0.015 & 0.015 & 0.027 \\
      Speaker        & 0.016 & 0.017 & 0.014 & 0.016 & 0.019 & 0.017 & 0.016 & 0.014 & 0.011 & 0.010 & 0.014 & 0.077 & 0.085 & 0.013 & 0.014 & 0.015 & 0.014 & 0.013 & 0.015 & 0.014 & 0.020 \\
      Sentiment      & 0.040 & 0.041 & 0.037 & 0.039 & 0.037 & 0.039 & 0.038 & 0.043 & 0.044 & 0.047 & 0.046 & 0.102 & 0.097 & 0.036 & 0.039 & 0.036 & 0.039 & 0.038 & 0.034 & 0.041 & 0.045 \\
      Topic          & 0.057 & 0.053 & 0.049 & 0.050 & 0.056 & 0.052 & 0.055 & 0.058 & 0.057 & 0.063 & 0.059 & 0.192 & 0.204 & 0.048 & 0.052 & 0.051 & 0.051 & 0.050 & 0.048 & 0.058 & 0.063 \\
      Agent Action   & 0.165 & 0.168 & 0.161 & 0.164 & 0.169 & 0.168 & 0.172 & 0.175 & 0.174 & 0.178 & 0.177 & 0.228 & 0.230 & 0.163 & 0.169 & 0.162 & 0.167 & 0.164 & 0.164 & 0.173 & 0.170 \\
      Solution       & 0.040 & 0.031 & 0.028 & 0.025 & 0.025 & 0.023 & 0.027 & 0.029 & 0.033 & 0.033 & 0.034 & 0.126 & 0.124 & 0.025 & 0.028 & 0.022 & 0.025 & 0.023 & 0.023 & 0.025 & 0.038 \\
      Politeness     & 0.036 & 0.041 & 0.035 & 0.038 & 0.039 & 0.038 & 0.037 & 0.033 & 0.033 & 0.030 & 0.036 & 0.103 & 0.100 & 0.036 & 0.035 & 0.034 & 0.034 & 0.035 & 0.034 & 0.037 & 0.042 \\
      Urgency        & 0.023 & 0.026 & 0.022 & 0.021 & 0.022 & 0.024 & 0.024 & 0.025 & 0.027 & 0.025 & 0.025 & 0.090 & 0.078 & 0.022 & 0.020 & 0.023 & 0.020 & 0.024 & 0.021 & 0.023 & 0.029 \\
      Order          & 0.392 & 0.360 & 0.339 & 0.355 & 0.376 & 0.376 & 0.412 & 0.365 & 0.357 & 0.344 & 0.339 & 0.609 & 0.644 & 0.377 & 0.386 & 0.350 & 0.332 & 0.330 & 0.357 & 0.361 & 0.377 \\
      Emotion        & 0.118 & 0.159 & 0.138 & 0.136 & 0.144 & 0.138 & 0.134 & 0.142 & 0.122 & 0.131 & 0.133 & 0.129 & 0.098 & 0.154 & 0.140 & 0.138 & 0.130 & 0.127 & 0.126 & 0.112 & 0.136 \\
      Repetition     & 0.097 & 0.106 & 0.084 & 0.083 & 0.087 & 0.086 & 0.096 & 0.089 & 0.096 & 0.095 & 0.097 & 0.129 & 0.124 & 0.087 & 0.086 & 0.083 & 0.084 & 0.089 & 0.080 & 0.091 & 0.096 \\
      Disfluency     & 0.055 & 0.053 & 0.053 & 0.052 & 0.053 & 0.052 & 0.058 & 0.055 & 0.054 & 0.055 & 0.053 & 0.105 & 0.103 & 0.053 & 0.051 & 0.050 & 0.052 & 0.054 & 0.051 & 0.054 & 0.060 \\
      Length         & 0.016 & 0.016 & 0.014 & 0.016 & 0.016 & 0.016 & 0.017 & 0.015 & 0.016 & 0.015 & 0.016 & 0.079 & 0.091 & 0.015 & 0.016 & 0.014 & 0.015 & 0.014 & 0.015 & 0.016 & 0.023 \\
      Language       & 0.039 & 0.039 & 0.033 & 0.037 & 0.036 & 0.036 & 0.039 & 0.036 & 0.035 & 0.035 & 0.038 & 0.122 & 0.131 & 0.034 & 0.035 & 0.035 & 0.035 & 0.034 & 0.034 & 0.036 & 0.043 \\
      Entity         & 0.172 & 0.173 & 0.154 & 0.150 & 0.180 & 0.176 & 0.186 & 0.122 & 0.097 & 0.088 & 0.128 & 0.181 & 0.208 & 0.187 & 0.174 & 0.196 & 0.156 & 0.159 & 0.185 & 0.117 & 0.154 \\
      \midrule
      \textbf{Average} & 0.089 & 0.087 & 0.079 & 0.081 & 0.085 & 0.084 & 0.088 & 0.083 & 0.082 & 0.082 & 0.083 & 0.168 & 0.172 & 0.083 & 0.084 & 0.083 & 0.082 & 0.081 & 0.081 & 0.080 & -- \\
      \midrule
      \multicolumn{21}{@{}l}{\textit{\textbf{Coverage}}} \\
      Position       & 99.63 & 98.67 & 98.79 & 98.67 & 98.67 & 98.67 & 98.79 & 98.67 & 98.67 & 98.06 & 98.67 & 66.52 & 65.39 & 98.67 & 98.67 & 98.18 & 98.79 & 98.67 & 98.67 & 98.67 & 94.13 \\
      Speaker        & 100.00 & 98.79 & 98.79 & 98.79 & 98.79 & 98.79 & 98.79 & 98.79 & 98.79 & 98.18 & 98.79 & 75.76 & 74.55 & 98.79 & 98.79 & 98.18 & 98.79 & 98.79 & 98.79 & 98.79 & 96.36 \\
      Sentiment      & 89.69 & 91.48 & 93.15 & 92.06 & 90.76 & 91.68 & 91.04 & 88.92 & 89.94 & 88.86 & 89.77 & 60.36 & 59.60 & 92.52 & 92.13 & 91.91 & 91.67 & 91.34 & 92.32 & 89.89 & 87.94 \\
      Topic          & 76.03 & 78.51 & 81.30 & 80.72 & 78.87 & 80.06 & 77.38 & 76.83 & 75.40 & 74.04 & 76.35 & 40.47 & 39.77 & 81.45 & 79.40 & 79.85 & 79.99 & 80.57 & 79.60 & 75.70 & 72.72 \\
      Agent Action   & 68.58 & 65.57 & 69.04 & 69.14 & 67.34 & 67.71 & 65.54 & 65.68 & 64.10 & 64.85 & 63.48 & 40.85 & 41.35 & 68.94 & 68.02 & 69.14 & 68.38 & 68.84 & 70.04 & 65.58 & 64.56 \\
      Solution       & 81.55 & 84.90 & 87.57 & 85.35 & 86.98 & 88.29 & 85.19 & 85.56 & 83.71 & 84.24 & 83.27 & 46.02 & 44.81 & 85.93 & 86.10 & 88.28 & 88.24 & 86.79 & 87.61 & 86.74 & 81.68 \\
      Politeness     & 95.04 & 94.60 & 95.30 & 94.55 & 94.09 & 93.99 & 95.20 & 93.18 & 92.68 & 92.17 & 94.14 & 66.72 & 67.73 & 95.10 & 94.34 & 94.60 & 95.00 & 94.80 & 95.40 & 93.84 & 91.56 \\
      Urgency        & 95.91 & 93.84 & 94.95 & 95.15 & 94.55 & 93.84 & 92.93 & 93.33 & 93.84 & 93.23 & 93.23 & 59.95 & 57.22 & 94.39 & 94.85 & 94.65 & 95.25 & 93.64 & 94.85 & 93.74 & 90.60 \\
      Repetition     & 67.78 & 63.26 & 66.67 & 67.98 & 65.51 & 65.36 & 64.83 & 65.78 & 64.36 & 64.94 & 64.94 & 32.29 & 31.50 & 67.51 & 67.92 & 67.56 & 68.02 & 66.40 & 69.71 & 66.61 & 62.32 \\
      Disfluency     & 69.42 & 70.53 & 71.63 & 70.66 & 71.27 & 71.76 & 68.91 & 70.39 & 70.35 & 69.55 & 70.80 & 41.41 & 40.12 & 71.37 & 71.31 & 70.95 & 72.05 & 71.98 & 72.65 & 70.22 & 67.02 \\
      Length         & 87.33 & 86.24 & 87.36 & 85.85 & 86.67 & 86.09 & 86.24 & 86.73 & 86.58 & 85.64 & 86.70 & 56.45 & 56.36 & 86.85 & 85.85 & 87.33 & 86.61 & 88.09 & 87.94 & 85.52 & 82.48 \\
      Language       & 82.70 & 83.55 & 84.99 & 83.57 & 84.36 & 82.85 & 82.17 & 83.65 & 83.36 & 83.84 & 83.47 & 49.75 & 49.14 & 85.54 & 83.89 & 82.93 & 82.85 & 84.15 & 84.37 & 84.00 & 79.89 \\
      Entity         & 50.50 & 51.55 & 53.77 & 55.24 & 47.62 & 48.89 & 47.93 & 60.42 & 68.49 & 72.37 & 59.56 & 19.58 & 17.61 & 47.18 & 49.26 & 44.53 & 52.55 & 52.93 & 47.77 & 63.00 & 50.78 \\
      \midrule
      \textbf{Average} & 81.56 & 80.85 & 82.68 & 82.36 & 80.98 & 80.91 & 80.22 & 80.40 & 80.12 & 79.80 & 80.17 & 50.63 & 49.65 & 82.19 & 81.75 & 81.89 & 82.65 & 82.68 & 83.32 & 81.94 & -- \\
      \midrule
    \textbf{LLM Judge Score} 
    & 2.10  
    & 3.96  
    & 4.76  
    & 4.85  
    & 4.71  
    & 4.58  
    & 4.84  
    & 4.82  
    & 4.73  
    & 4.79  
    & 4.75  
    & 3.27  
    & 3.31  
    & 4.67  
    & 4.84  
    & 4.73  
    & 4.80  
    & 4.80  
    & 4.76  
    & 4.83  
    & 4.45
    \\

      \textbf{Compression Ratio} & 0.142 & 0.069 & 0.079 & 0.065 & 0.050 & 0.056 & 0.044 & 0.058 & 0.068 & 0.075 & 0.060 & 0.021 & 0.019 & 0.049 & 0.052 & 0.045 & 0.060 & 0.070 & 0.065 & 0.061 & 0.063 \\
      \textbf{Compression Factor} & 9.41 & 15.46 & 13.33 & 16.26 & 20.98 & 19.55 & 24.77 & 17.71 & 15.32 & 13.68 & 18.00 & 86.31 & 87.14 & 21.44 & 20.82 & 23.56 & 17.79 & 15.33 & 16.43 & 18.07 & 25.46 \\
      \bottomrule
    \end{tabular}%
  }
  \caption{Model performance on medium-length transcripts (3000-6000 tokens).}
  \label{table:model_performance_2}
\end{table*}

\begin{table*}[!h]
  \centering
  \small
  \renewcommand{\arraystretch}{1.2}
  \resizebox{\textwidth}{!}{%
    \begin{tabular}{@{}l *{21}{>{\centering\arraybackslash}p{1.2cm}} @{}}
      \toprule
      \textbf{Metric / Bias} 
        & \rotatebox{90}{\texttt{llama-3.2-1b}} 
        & \rotatebox{90}{\texttt{llama-3.2-3b}} 
        & \rotatebox{90}{\texttt{llama-3.3-70b}} 
        & \rotatebox{90}{\texttt{llama-4-maverick}} 
        & \rotatebox{90}{\texttt{nova-micro}} 
        & \rotatebox{90}{\texttt{nova-lite}} 
        & \rotatebox{90}{\texttt{nova-pro}} 
        & \rotatebox{90}{\texttt{claude-3.5-haiku}} 
        & \rotatebox{90}{\texttt{claude-3.7-sonnet}} 
        & \rotatebox{90}{\texttt{claude-4-sonnet}} 
        & \rotatebox{90}{\texttt{deepseek-r1}} 
        & \rotatebox{90}{\texttt{gemini-2.0-flash}} 
        & \rotatebox{90}{\texttt{gemini-2.0-flash-lite}} 
        & \rotatebox{90}{\texttt{gpt-4o-mini}} 
        & \rotatebox{90}{\texttt{gpt-4o}} 
        & \rotatebox{90}{\texttt{gpt-4.1-nano}} 
        & \rotatebox{90}{\texttt{gpt-4.1-mini}} 
        & \rotatebox{90}{\texttt{gpt-4.1}} 
        & \rotatebox{90}{\texttt{o3-mini}} 
        & \rotatebox{90}{\texttt{o4-mini}}
        & \rotatebox{90}
        {\textbf{Average}}\\
      \midrule
      \multicolumn{21}{@{}l}{\textit{\textbf{JS Divergence (JSD)}}} \\
      Position       & 0.027 & 0.019 & 0.016 & 0.016 & 0.017 & 0.015 & 0.018 & 0.016 & 0.017 & 0.016 & 0.017 & 0.021 & 0.019 & 0.018 & 0.018 & 0.014 & 0.014 & 0.015 & 0.014 & 0.017 & 0.017 \\
      Speaker        & 0.019 & 0.018 & 0.016 & 0.016 & 0.020 & 0.018 & 0.019 & 0.016 & 0.015 & 0.013 & 0.017 & 0.019 & 0.020 & 0.014 & 0.016 & 0.018 & 0.014 & 0.017 & 0.017 & 0.016 & 0.017 \\
      Sentiment      & 0.043 & 0.043 & 0.040 & 0.042 & 0.041 & 0.041 & 0.043 & 0.045 & 0.048 & 0.049 & 0.047 & 0.045 & 0.041 & 0.041 & 0.041 & 0.038 & 0.041 & 0.041 & 0.039 & 0.044 & 0.042 \\
      Topic          & 0.059 & 0.050 & 0.047 & 0.048 & 0.050 & 0.050 & 0.053 & 0.052 & 0.056 & 0.055 & 0.057 & 0.062 & 0.054 & 0.044 & 0.051 & 0.045 & 0.047 & 0.046 & 0.046 & 0.052 & 0.051 \\
      Agent Action   & 0.191 & 0.189 & 0.188 & 0.192 & 0.193 & 0.196 & 0.198 & 0.195 & 0.200 & 0.199 & 0.202 & 0.201 & 0.196 & 0.187 & 0.192 & 0.189 & 0.194 & 0.191 & 0.191 & 0.198 & 0.193 \\
      Solution       & 0.052 & 0.031 & 0.031 & 0.032 & 0.029 & 0.031 & 0.034 & 0.034 & 0.035 & 0.036 & 0.032 & 0.041 & 0.034 & 0.029 & 0.029 & 0.023 & 0.027 & 0.026 & 0.025 & 0.030 & 0.032 \\
      Politeness     & 0.039 & 0.040 & 0.038 & 0.038 & 0.040 & 0.039 & 0.040 & 0.036 & 0.035 & 0.033 & 0.039 & 0.039 & 0.040 & 0.036 & 0.037 & 0.034 & 0.035 & 0.035 & 0.033 & 0.037 & 0.037 \\
      Urgency        & 0.025 & 0.022 & 0.023 & 0.023 & 0.023 & 0.022 & 0.024 & 0.024 & 0.023 & 0.025 & 0.026 & 0.024 & 0.024 & 0.023 & 0.022 & 0.021 & 0.023 & 0.022 & 0.022 & 0.022 & 0.023 \\
      Order          & 0.394 & 0.366 & 0.343 & 0.356 & 0.385 & 0.363 & 0.370 & 0.368 & 0.357 & 0.358 & 0.354 & 0.404 & 0.397 & 0.381 & 0.388 & 0.354 & 0.325 & 0.336 & 0.352 & 0.345 & 0.365 \\
      Emotion        & 0.122 & 0.139 & 0.138 & 0.131 & 0.143 & 0.143 & 0.134 & 0.128 & 0.116 & 0.112 & 0.127 & 0.128 & 0.127 & 0.152 & 0.140 & 0.139 & 0.132 & 0.124 & 0.104 & 0.123 & 0.130 \\
      Repetition     & 0.093 & 0.085 & 0.085 & 0.085 & 0.085 & 0.088 & 0.085 & 0.084 & 0.088 & 0.082 & 0.080 & 0.091 & 0.086 & 0.074 & 0.091 & 0.077 & 0.076 & 0.080 & 0.075 & 0.083 & 0.083 \\
      Disfluency     & 0.051 & 0.048 & 0.046 & 0.047 & 0.045 & 0.049 & 0.050 & 0.048 & 0.050 & 0.050 & 0.050 & 0.052 & 0.052 & 0.046 & 0.048 & 0.045 & 0.048 & 0.045 & 0.052 & 0.048 & 0.048 \\
      Length        & 0.015 & 0.013 & 0.012 & 0.013 & 0.013 & 0.012 & 0.013 & 0.013 & 0.014 & 0.013 & 0.014 & 0.015 & 0.013 & 0.012 & 0.013 & 0.011 & 0.012 & 0.012 & 0.011 & 0.013 & 0.013 \\
      Language      & 0.044 & 0.041 & 0.038 & 0.039 & 0.040 & 0.038 & 0.042 & 0.040 & 0.040 & 0.039 & 0.042 & 0.046 & 0.043 & 0.035 & 0.038 & 0.035 & 0.037 & 0.036 & 0.040 & 0.038 & 0.039 \\
      Entity        & 0.172 & 0.158 & 0.153 & 0.133 & 0.187 & 0.176 & 0.179 & 0.120 & 0.101 & 0.090 & 0.121 & 0.169 & 0.187 & 0.183 & 0.180 & 0.194 & 0.146 & 0.148 & 0.174 & 0.112 & 0.152 \\
      \midrule
      \textbf{Average} & 0.092 & 0.085 & 0.082 & 0.083 & 0.086 & 0.085 & 0.086 & 0.085 & 0.085 & 0.084 & 0.083 & 0.092 & 0.090 & 0.083 & 0.086 & 0.080 & 0.080 & 0.081 & 0.081 & 0.082 & -- \\
      \midrule
      \multicolumn{21}{@{}l}{\textit{\textbf{Coverage}}} \\
      Position       & 98.06 & 97.24 & 97.93 & 97.52 & 97.86 & 97.59 & 97.93 & 97.59 & 97.86 & 97.52 & 97.38 & 95.93 & 97.24 & 98.28 & 98.55 & 97.59 & 97.59 & 97.59 & 97.59 & 97.93 & 97.67 \\
      Speaker        & 98.26 & 97.24 & 97.93 & 97.59 & 97.93 & 97.59 & 97.93 & 97.59 & 97.93 & 97.59 & 97.59 & 96.03 & 97.24 & 98.62 & 98.62 & 97.59 & 97.59 & 97.59 & 97.59 & 97.93 & 97.76 \\
      Sentiment      & 87.76 & 88.98 & 90.07 & 89.43 & 88.61 & 89.83 & 88.41 & 88.96 & 88.56 & 87.60 & 88.34 & 85.44 & 86.51 & 91.83 & 89.37 & 90.33 & 90.63 & 88.83 & 88.91 & 88.33 & 88.76 \\
      Topic          & 74.26 & 77.90 & 79.96 & 78.32 & 76.73 & 76.52 & 75.32 & 74.02 & 73.78 & 71.25 & 74.30 & 69.93 & 73.43 & 80.76 & 78.83 & 80.32 & 77.74 & 77.53 & 77.52 & 73.44 & 76.04 \\
      Agent Action   & 68.65 & 70.58 & 72.43 & 70.12 & 69.08 & 69.33 & 67.04 & 67.59 & 66.43 & 65.70 & 65.61 & 64.76 & 66.94 & 72.41 & 70.02 & 72.99 & 70.12 & 71.10 & 67.22 & 67.22 & 68.96 \\
      Solution       & 77.72 & 83.90 & 85.42 & 84.89 & 84.52 & 85.13 & 83.71 & 81.77 & 81.77 & 81.04 & 83.08 & 78.78 & 82.61 & 85.34 & 85.70 & 87.00 & 85.04 & 86.43 & 83.99 & 83.99 & 83.27 \\
      Politeness     & 94.76 & 95.20 & 95.89 & 95.40 & 94.94 & 94.51 & 94.91 & 93.42 & 93.13 & 92.44 & 93.88 & 91.87 & 92.96 & 96.35 & 95.92 & 95.29 & 95.17 & 94.57 & 94.52 & 93.45 & 94.19 \\
      Urgency        & 91.55 & 92.30 & 93.88 & 92.33 & 92.76 & 92.33 & 92.70 & 92.82 & 93.25 & 91.78 & 92.41 & 91.72 & 91.38 & 93.94 & 93.16 & 92.99 & 92.93 & 93.51 & 92.87 & 92.87 & 92.58 \\
      Repetition     & 58.09 & 60.72 & 57.83 & 60.01 & 59.11 & 59.44 & 57.65 & 56.24 & 57.89 & 59.98 & 58.12 & 55.44 & 54.24 & 62.07 & 57.26 & 63.71 & 61.17 & 63.80 & 58.16 & 60.16 & 59.07 \\
      Disfluency     & 67.05 & 66.95 & 68.69 & 66.41 & 68.56 & 66.86 & 66.38 & 68.03 & 66.54 & 66.54 & 66.29 & 63.16 & 65.37 & 69.17 & 68.02 & 70.43 & 67.84 & 69.49 & 65.60 & 65.60 & 66.99 \\
      Length        & 87.43 & 88.74 & 89.29 & 87.91 & 88.47 & 88.74 & 87.16 & 87.00 & 86.74 & 86.50 & 86.40 & 85.22 & 87.02 & 89.79 & 89.38 & 89.86 & 89.24 & 89.09 & 87.28 & 87.28 & 87.85 \\
      Language      & 80.84 & 82.54 & 83.73 & 82.47 & 82.17 & 82.27 & 81.46 & 81.31 & 81.33 & 80.85 & 81.61 & 78.11 & 80.59 & 84.09 & 82.99 & 84.76 & 83.07 & 82.92 & 81.57 & 81.57 & 82.01 \\
      Entity        & 49.09 & 50.11 & 51.03 & 54.83 & 44.19 & 46.50 & 46.29 & 57.75 & 63.84 & 68.04 & 58.06 & 46.89 & 44.27 & 43.46 & 46.11 & 42.04 & 53.26 & 47.19 & 42.09 & 61.08 & 50.70 \\
      \midrule
      \textbf{Average} & 79.05 & 80.42 & 81.31 & 80.41 & 79.73 & 80.10 & 79.25 & 79.67 & 79.48 & 79.10 & 78.96 & 75.91 & 77.96 & 82.12 & 80.65 & 81.64 & 80.76 & 81.10 & 78.93 & 80.12 & -- \\
      \midrule
    \textbf{LLM Judge Score} 
    & 2.06  
    & 4.11  
    & 4.80  
    & 4.87  
    & 4.65  
    & 4.64  
    & 4.84  
    & 4.84  
    & 4.73  
    & 4.82  
    & 4.69  
    & 4.54  
    & 4.66  
    & 4.73  
    & 4.85  
    & 4.70  
    & 4.76  
    & 4.75  
    & 4.72  
    & 4.77  
    & 4.61
    \\

      \textbf{Compression Ratio} & 0.094 & 0.043 & 0.045 & 0.039 & 0.029 & 0.034 & 0.027 & 0.033 & 0.041 & 0.045 & 0.039 & 0.024 & 0.024 & 0.030 & 0.027 & 0.027 & 0.040 & 0.049 & 0.041 & 0.040 & 0.040 \\
      \textbf{Compression Factor} & 14.08 & 24.94 & 23.22 & 27.57 & 37.03 & 33.48 & 41.80 & 31.11 & 25.24 & 23.03 & 28.29 & 51.16 & 48.52 & 34.69 & 38.28 & 38.76 & 26.42 & 21.66 & 26.24 & 27.08 & 30.79 \\
      \bottomrule
    \end{tabular}%
  }
  \caption{Model performance on long transcripts (>6000 tokens).}
  \label{table:model_performance_3}
\end{table*}

\subsection{Correlations Between Metrics}
To understand the relationships between traditional quality metrics and our bias framework, we computed the Pearson correlation coefficients between them (Table~\ref{table:correlation_coefficients}). We observe a strong positive correlation between higher compression and higher bias (JSD), and a strong negative correlation between compression and coverage. This confirms the intuitive idea that more aggressive summarization leads to greater information loss and distortion. Conversely, the correlation between the LLM Judge Score and our bias metrics is weak, highlighting that holistic quality scores often fail to capture these fine-grained fidelity issues.

\begin{table*}[!h]
  \centering
  \small
  \renewcommand{\arraystretch}{1.2}
  \resizebox{\textwidth}{!}{%
    \begin{tabular}{@{}l *{4}{>{\centering\arraybackslash}p{3cm}} @{}}
    \toprule
    \textbf{Metric} & 
    \textbf{LLM Judge vs JS Divergence} & 
    \textbf{LLM Judge vs Coverage} & 
    \textbf{Compression vs JS Divergence} & 
    \textbf{Compression vs Coverage} \\
    \midrule
    Turn Length & -0.3499 & 0.2896 & 0.9212 & -0.9148 \\
    Speaker & -0.3603 & 0.2577 & 0.9327 & -0.9316 \\
    Position & -0.4203 & 0.2836 & 0.9001 & -0.9306 \\
    Urgency & -0.3336 & 0.3296 & 0.8897 & -0.9170 \\
    Solution & -0.5516 & 0.4536 & 0.7894 & -0.8506 \\
    Politeness & -0.3672 & 0.2801 & 0.9178 & -0.9044 \\
    Language Complexity & -0.3859 & 0.3423 & 0.9086 & -0.9170 \\
    Sentiment & -0.2970 & 0.3437 & 0.8317 & -0.8934 \\
    Disfluency & -0.4061 & 0.3541 & 0.8727 & -0.9006 \\
    Topic & -0.3696 & 0.3551 & 0.8959 & -0.8438 \\
    Information Repetition & -0.5058 & 0.3443 & 0.6078 & -0.8831 \\
    Emotion Shift & 0.2949 & -- & -0.1854 & -- \\
    Entity Type & -0.2833 & 0.3267 & 0.4364 & -0.7538 \\
    Agent Action & -0.2619 & 0.2866 & 0.8490 & -0.8500 \\
    Temporal Sequence & -0.4708 & -- & 0.8701 & -- \\
    \bottomrule
  \end{tabular}%
  }
  \caption{Pearson correlation coefficients between key metrics. Strong correlations exist between compression and bias metrics, while correlations with LLM Judge scores are weak, underscoring the need for our framework.}
  \label{table:correlation_coefficients}
\end{table*}

\subsection{Model Clustering and Outlier Analysis}
To visualize the behavioral similarities between the evaluated models, we performed a Principal Component Analysis (PCA) on the models' JSD and Coverage bias vectors (where each vector consists of a model's 15 JSD scores and 13 coverage scores). As shown in Figure~\ref{figure:1000}, the PCA plot reveals that most of the 20 LLMs, regardless of family or scale, cluster tightly in a specific region. This dense clustering provides strong visual evidence for our central claim that these biases are systemic and not specific to a few poorly performing models. The plot also clearly identifies the two Gemini models as significant outliers, exhibiting a distinct and more severe bias profile compared to the rest of the field.

\begin{figure*}[!ht]
    \centering
    \includegraphics[width=0.9\linewidth]{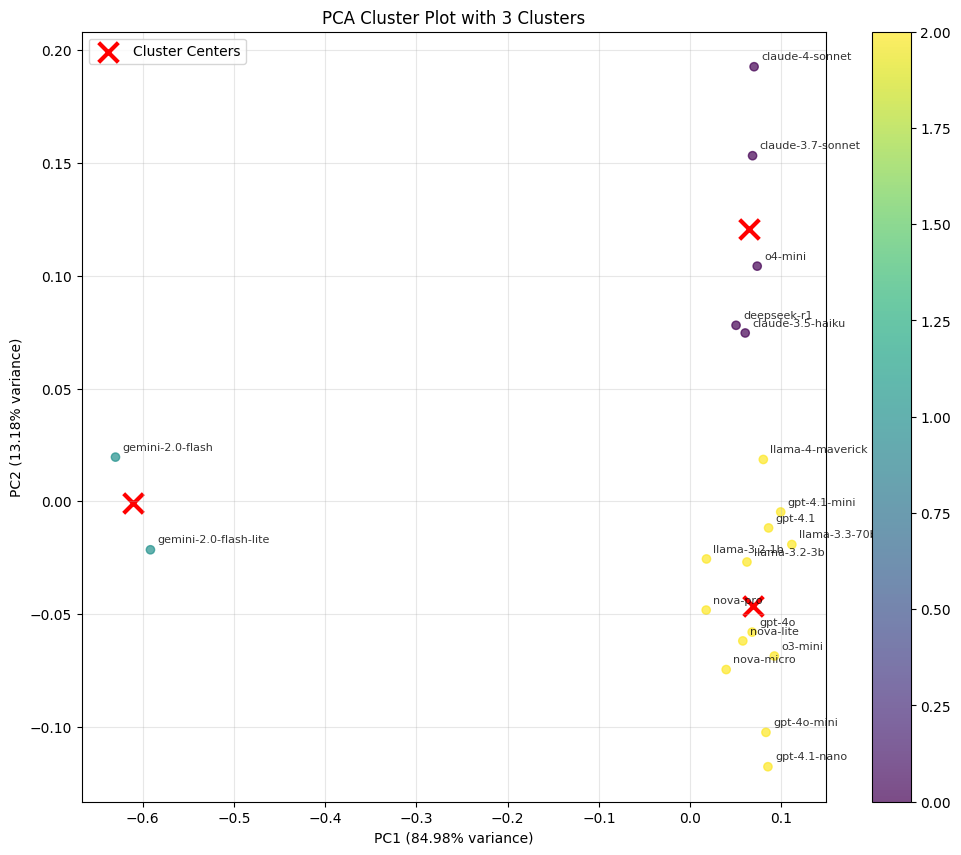}
    \caption{PCA projection of the 20 LLMs based on their 15-dimensional JSD bias vectors. The tight clusters towards the right indicates a shared, systemic bias profile across most models. The two Gemini models are clear outliers with a distinctly different and higher bias profile.}
    \label{figure:1000}
\end{figure*}

\subsection{Fine-Grained Label Representation Analysis}
Beyond aggregate scores, our framework allows for an analysis of which specific labels are systematically over- or under-represented. Figure~\ref{figure:over-rep} illustrates the labels with the most significant positive (over-represented) and negative (under-represented) skew, averaged across all models. This analysis reveals a consistent narrative strategy: models tend to amplify labels related to problem statements (e.g., `Negative` sentiment, `Issue` topic) while omitting labels related to conversational context and resolution (e.g., `Rapport-Building`, `Directives`). This provides a deeper explanation for the observed biases, linking them to the models' implicit assumptions about what is important in a conversation.

\begin{figure*}
    \centering
    \includegraphics[width=0.9\linewidth]{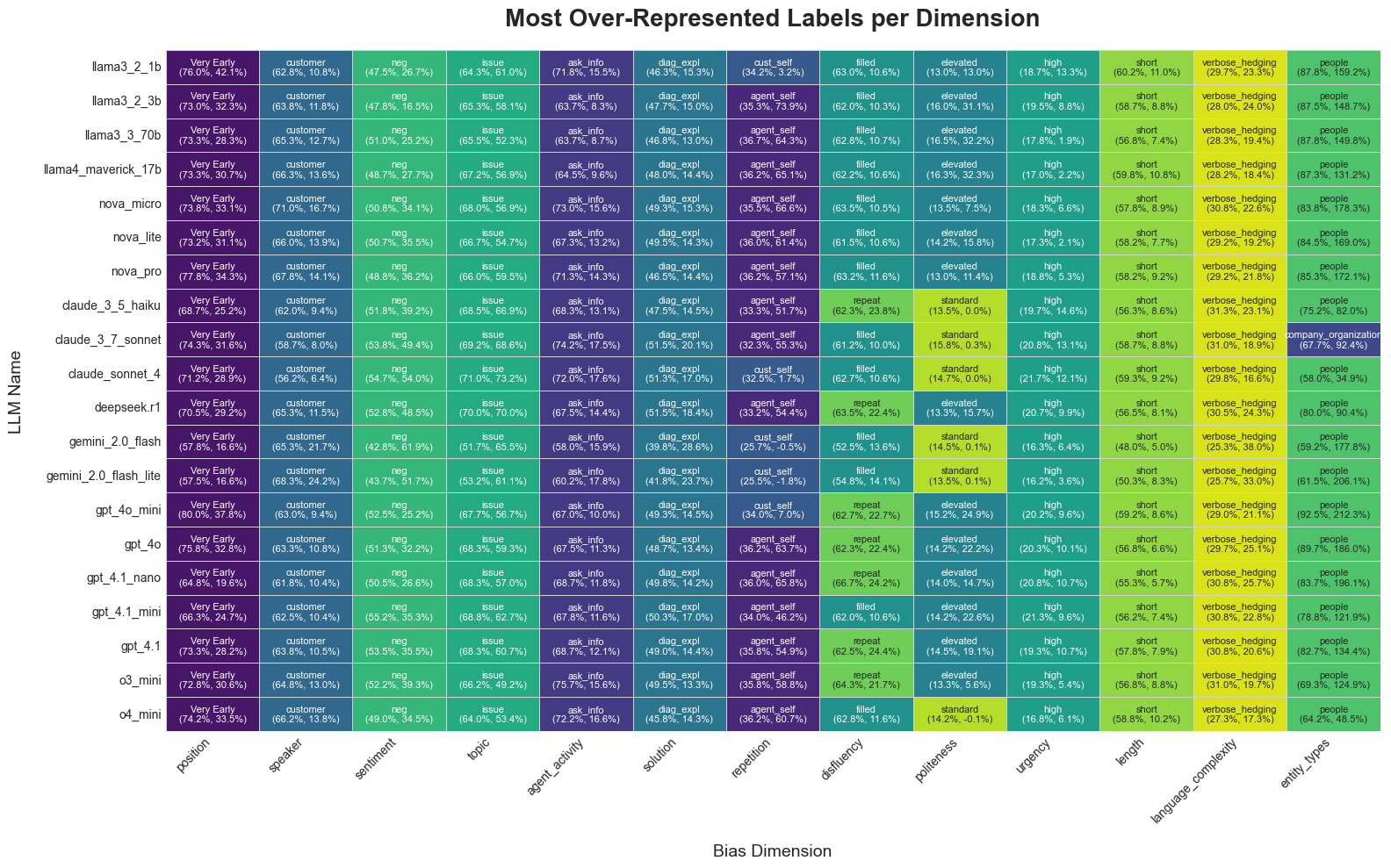}
    \includegraphics[width=0.9\linewidth]{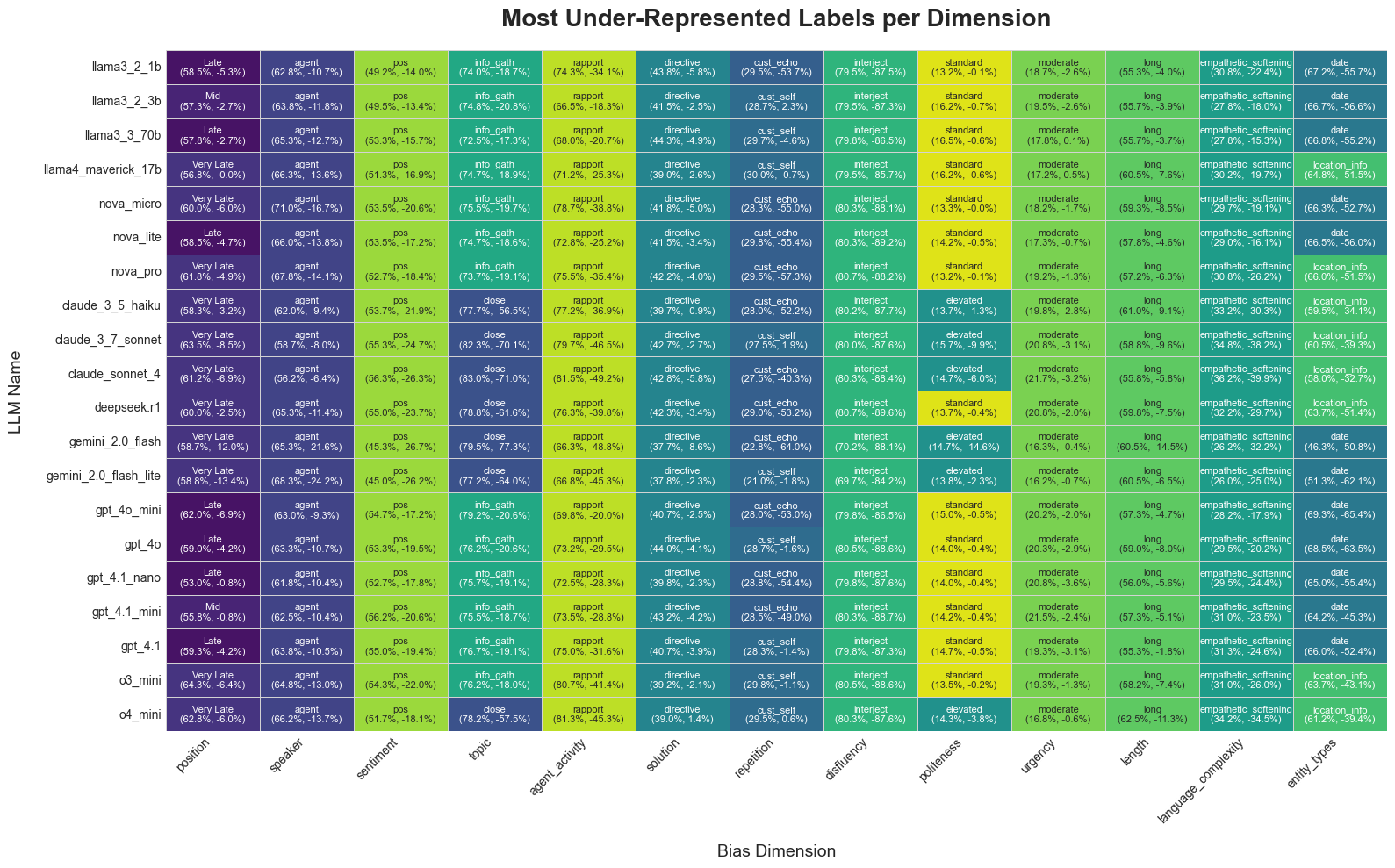}
    \caption{Systematic over- and under-representation of specific labels across all 20 LLMs. The top plot depicts over-represented labels; the bottom plot depicts under-represented labels. Each label is annotated with two percentages indicating (1) the proportion of summaries in which it is over- or under-represented and (2) the average magnitude of that deviation. The results reveal a consistent, cross-model tendency to construct simplified, problem-centric narratives while underrepresenting interactional and resolution-focused content. }
    \label{figure:over-rep}
\end{figure*}

\onecolumn

\definecolor{lightgray}{gray}{0.95}

\section{Prompts}
\label{appendix:prompts}

\begin{tcolorbox}[breakable, colframe=orange!85!black, colback=orange!10!white, title= Prompt for Semantic Proposition Extractor]
You are a semantic analysis assistant. Your task is to decompose the given paragraph into atomic semantic propositions. Each proposition must preserve a minimal, standalone unit of meaning and reflect a single assertion or fact conveyed by the text.

\textbf{Guidelines:}
\begin{enumerate}[label=\arabic*.]
    \item Use the original words where possible; do not paraphrase unnecessarily.
    \item Resolve pronouns if possible.
    \item A proposition should typically follow the (subject; predicate; object/modifier) structure.
    \item Include time, place, and recipient details as separate propositions when appropriate.
    \item Do not explain or justify. Just return the list of propositions.
\end{enumerate}

Next, extract entities from the summary and categorize them into the following predefined types:
\begin{itemize}
    \item \textbf{people}: Agent name, Customer name, 3rd parties
    \item \textbf{identifiers}: Ticket ID, Account No., Policy No.
    \item \textbf{phone\_number}: Phone numbers
    \item \textbf{email}: Email addresses
    \item \textbf{time\_info}: Time, Duration, Deadlines
    \item \textbf{date}: Dates
    \item \textbf{location\_info}: Address, City, Branch
    \item \textbf{products\_services}: Items discussed or complained about
    \item \textbf{monetary}: Price, Refund, Discount
    \item \textbf{company\_organization}: Mentioned institutions
    \item \textbf{others}: Miscellaneous/Unclassified entities
\end{itemize}

\textbf{Return a JSON object} with two keys: \texttt{propositions} and \texttt{entities}.

\begin{itemize}
    \item \texttt{propositions}: an object where keys are sequential numeric strings (e.g., \texttt{"1"}, \texttt{"2"}) and values are the proposition texts.
    \item \texttt{entities}: an object with the exact keys listed above, each containing a list of extracted entities (even if empty).

    \item "The overall JSON structure should be: 
    {
      "propositions": {
        "1": "John filed a complaint.",
        "2": "The issue occurred yesterday at 10 AM."
      },
      "entities": {
        "people": ["John"],
        "identifiers": ["WR123X62"],
        "phone\_number": [],
        "email": [],
        "time\_info": ["10 AM"],
        "date": ["yesterday"],
        "location\_info": [],
        "products\_services": ["mobile insurance"],
        "monetary": ["\$112"],
        "company\_organization": ["accolade"],
        "others": []
      }
    }
    \end{itemize}
    \textbf{User Prompt:}
    \newline
   Process the following summary to extract semantic propositions and entities and provide the output in JSON format:\texttt{\textbackslash n\textbackslash nSummary:\textbackslash n"}"
\end{tcolorbox}

\begin{tcolorbox}[breakable, colframe=orange!85!black, colback=orange!10!white,  title=Prompt for Summarization]

\textbf{System Prompt}: You are a helpful assistant designed to summarize text.

\textbf{User Prompt Templates:}
\begin{enumerate}[label=\arabic*.]
    \item Summarize the following dialog. \texttt{<dialog> \{transcript\} </dialog>}
    \item Please provide a summary of the contact-center conversation transcript. \texttt{<transcript> \{transcript\} </transcript>}
    \item Generate a summary of the conversation. \texttt{<conversation> \{transcript\} </conversation>}
\end{enumerate}
\end{tcolorbox}

\begin{tcolorbox}[breakable, colframe=orange!85!black, colback=orange!10!white, title= Prompt for Transcript Labeling]
You are a transcript analysis assistant. Your task is to annotate each turn in a conversation transcript using a fixed set of linguistic and conversational dimensions, and separately extract entities mentioned across the entire transcript.

Each turn begins like: \texttt{"1: Speaker: ..."} Analyze each turn independently.

\textbf{Dimensions (Fixed Order with Short Labels)}

Each turn must be annotated in the following order. Always include all dimensions. Empty lists are allowed where applicable.

\begin{center}
\begin{tabular}{|l|l|l|}
\hline
\textbf{Key} & \textbf{Dimension Name} & \textbf{Type} \\
\hline
\texttt{sent} & Sentiment & single value \\
\texttt{topic} & Topic Category & single value \\
\texttt{agent} & Agent Action & single value \\
\texttt{sol} & Solution Type & list \\
\texttt{rep} & Information Repetition & single value \\
\texttt{disf} & Disfluency Patterns & list \\
\texttt{lang} & Language Complexity Patterns & list \\
\texttt{polite} & Politeness & single value \\
\texttt{urgency} & Urgency & single value \\
\hline
\end{tabular}
\end{center}

\textbf{Output Format (JSON)}

{
  "map": [
    [1, "neutral", "greet", "ask\_info", [], "no\_rep", [], [], "minimal", "low"],
    [2, "pos", "diag", "escalate", ["diag\_expl"], "cust\_self", ["filled"], ["plain", "formal"], "standard", "high"]
  ],
  "entity": {
    "people": ["Alex"],
    "phone\_number": [9512384859],
    "monetary": ["\$100"],
    ... (and other entity categories)
  }
}

\texttt{map}: List of arrays — one for each turn. Each array must contain 10 elements:
\texttt{[turn\_number, sent, topic, agent, sol, rep, disf, lang, polite, urgency]}

\texttt{entity}: Dictionary of extracted entities. Entity extraction is a separate task — do not confuse with turn-level annotation.

\textbf{Allowed Values and Glossary:}

\textbf{1. sent - Sentiment }
\begin{center}
\begin{tabular}{|l|p{10cm}|}
\hline
\textbf{Code} & \textbf{Meaning} \\
\hline
\texttt{very\_pos} & Strongly positive tone \\
\texttt{pos} & Moderately positive tone \\
\texttt{neg} & Moderately negative tone \\
\texttt{very\_neg} & Strongly negative tone \\
\texttt{info} & Information content or presence of factual tokens (dates, names, IDs) -- high priority over \texttt{neutral} \\
\texttt{neutral} & Does not have information and contains explicit neutral-emotion cues (e.g., ``okay,'' ``fine,'' ``so-so,'' ``not sure'') \\
\hline
\end{tabular}
\end{center}

\textbf{2. topic - Topic Category}
\begin{center}
\begin{tabular}{|l|l|}
\hline
\textbf{Code} & \textbf{Description} \\
\hline
\texttt{greet} & Greetings, introductions \\
\texttt{id\_verif} & ID or account verification \\
\texttt{issue} & Customer's reason for contact \\
\texttt{info\_gath} & Agent probing/investigating \\
\texttt{prod\_inq} & Product or service questions \\
\texttt{diag} & Diagnosis or troubleshooting \\
\texttt{soln} & Proposing a solution \\
\texttt{action} & Performing an action \\
\texttt{transact} & Payments, refunds, orders \\
\texttt{offers} & Service offers or upgrades \\
\texttt{sales} & Sales, upselling, persuasion \\
\texttt{resolve\_conf} & Confirming issue is resolved \\
\texttt{next} & Next steps, follow-ups \\
\texttt{close} & Farewell, call closure \\
\texttt{empathy} & Expressing care or rapport \\
\texttt{complaint} & Handling complaints/escalation \\
\texttt{policy} & Explaining rules or terms \\
\texttt{feedback} & Requesting feedback or surveys \\
\texttt{sched} & Appointments, scheduling \\
\texttt{billing} & Billing/payment issues \\
\texttt{compliance} & Compliance or regulations \\
\texttt{misc} & Miscellaneous \\
\hline
\end{tabular}
\end{center}

\textbf{3.agent - Agent Action}:
\begin{center}
\begin{tabular}{|l|l|l|}
\hline
\textbf{Code} & \textbf{Category} & \textbf{Notes} \\
\hline
\texttt{ask\_info} & Request Information & ``Could you confirm your order?'' \\
\texttt{give\_info} & Provide Information & Facts or explanations not tied to a fix \\
\texttt{check\_under} & Confirm Understanding & ``Do you see the change on your end?'' \\
\texttt{rapport} & Build Rapport & Empathy, friendliness, thank-you \\
\texttt{backchannel} & Acknowledgement / Cue & ``Uh-huh,'' ``Okay,'' ``Got it.'' \\
\texttt{escalate} & Escalate / Transfer Action & ``I'm connecting you to billing.'' \\
\texttt{compliance} & Compliance / Verification & Identity, policy, legal checks \\
\texttt{idle} & Passive / No-Op Response & Silence gaps marked or minimal reply \\
\texttt{other} & Other Conversational Act & Anything else (e.g., small talk) \\
\hline
\end{tabular}
\end{center}

\textbf{4.sol - Solution Type (multi-select)}
\begin{center}
\begin{tabular}{|l|l|}
\hline
\textbf{Code} & \textbf{Description} \\
\hline
\texttt{diag\_expl} & Diagnostic explanation \\
\texttt{advisory} & General advice \\
\texttt{root\_cause} & Explaining root cause \\
\texttt{directive} & Concrete steps or commands \\
\texttt{preventive} & Prevent future issues \\
\texttt{escalate} & Escalation or transfer \\
\texttt{self\_help} & Do-it-yourself instructions \\
\texttt{partial} & Incomplete or partial fix \\
\texttt{rejected} & Offered but not applied \\
\texttt{followup} & Future action promised \\
\texttt{expect} & Sets realistic timelines \\
\texttt{reassure} & Emotional closure \\
\texttt{no\_soln} & No solution given \\
\hline
\end{tabular}
\end{center}

\textbf{5. rep - Repetition}
\begin{center}
\begin{tabular}{|l|l|}
\hline
\textbf{Code} & \textbf{Description} \\
\hline
\texttt{no\_rep} & No repetition present \\
\texttt{cust\_self} & Customer repeats self \\
\texttt{agent\_self} & Agent repeats self \\
\texttt{cust\_echo} & Customer echoes agent \\
\texttt{agent\_echo} & Agent echoes customer \\
\hline
\end{tabular}
\end{center}

\textbf{6.disf - Disfluencies (multi-select)}
\begin{center}
\begin{tabular}{|l|l|}
\hline
\textbf{Code} & \textbf{Description} \\
\hline
\texttt{filled} & ``uh'', ``um'', etc. \\
\texttt{silent} & Silent pauses \\
\texttt{repeat} & Word/phrase repetition \\
\texttt{false\_start} & Incomplete start \\
\texttt{repair} & Self-correction \\
\texttt{prolong} & Stretched sounds \\
\texttt{stutter} & Repeated syllables \\
\texttt{marker} & Discourse filler (``like'', ``you know'') \\
\texttt{interject} & ``oh!'', ``hmm'' \\
\texttt{cutoff} & Abandoned utterance \\
\texttt{placeholder} & ``sort of'', ``you know what I mean'' \\
\texttt{overlap} & Overlapping talk \\
\hline
\end{tabular}
\end{center}

\textbf{7. lang - Language Complexity (multi-select)} 
\begin{center}
\begin{tabular}{|l|p{4cm}|p{5cm}|}
\hline
\textbf{Code} & \textbf{Description} & \textbf{Example} \\
\hline
\texttt{standard\_clear} & Clear, direct, and easily understood language. & The default if no other specific complexities are prominently featured. \\
\texttt{simple\_syntax} & Predominantly short, declarative sentences. & ``I can help. What is your name? The account is open.'' \\
\texttt{complex\_syntax} & Long, multi-clause, or convoluted sentences. & ``Given the information you've provided, and after checking the system, it appears that the issue, which started last Tuesday, will require a technician to resolve it.'' \\
\texttt{technical\_terms} & Specialized terms related to a specific domain. & ``Modem,'' ``IP address,'' ``deductible,'' ``API endpoint.'' \\
\texttt{industry\_jargon} & Terms/phrases specific to an industry/company. & ``Tier 2 escalation,'' ``SKU,'' ``churn rate,'' ``SOP.'' \\
\texttt{acronyms\_abbreviations} & Use of shortened forms of words or phrases. & ``ASAP,'' ``ID,'' ``ETA,'' ``KYC.'' \\
\texttt{info\_dense} & Highly concise; packed with specific information. & ``Policy AX47 requires form B2, due COB Friday for Q3 processing.'' \\
\texttt{verbose\_hedging} & Wordy, uses fillers, qualifiers, or vague language. & ``Well, you know, it's sort of like, I guess maybe we could perhaps try to see...'' \\
\texttt{formal\_register} & Polished, professional, often more structured. & ``We wish to inform you...'', ``It is imperative that...'' \\
\texttt{informal\_colloquial} & Conversational, casual, everyday language. & ``No worries!'', ``Gonna check that for ya.'', ``Awesome!'' \\
\texttt{empathetic\_softening} & Language used to show understanding or soften news. & ``I understand this must be frustrating...'', ``Unfortunately...'', ``I'm afraid...'' \\
\texttt{abrupt\_blunt} & Overly direct, lacking typical softeners/politeness. & ``No. Can't do that. Next.'' (Extreme example) \\
\texttt{idioms\_slang} & Figurative expressions or informal slang. & ``Bite the bullet'', ``cool'', ``spill the beans.'' \\
\texttt{passive\_voice\_prominent} & Significant use of passive voice construction. & ``The account was accessed'', ``A decision will be made.'' (When frequent) \\
\hline
\end{tabular}
\end{center}

\textbf{8. polite - Politeness} 
\begin{center}
\begin{tabular}{|l|p{11cm}|}
\hline
\textbf{Code} & \textbf{Description} \\
\hline
\texttt{none} & No politeness cues (no please/thank you/etc.) \\
\texttt{minimal} & One-off courtesy (``thank you'', ``please'') \\
\texttt{standard} & Expected level (``please let me know'', ``thanks for waiting'') \\
\texttt{elevated} & Multiple markers + honorifics (``sir/madam'', ``kindly'') \\
\texttt{impolite} & Impoliteness cues) \\
\hline
\end{tabular}
\end{center}

\textbf{9. urgency - Urgency}
\begin{center}
\begin{tabular}{|l|p{11cm}|}
\hline
\textbf{Code} & \textbf{Description} \\
\hline
\texttt{none} & No urgency language \\
\texttt{low} & Mild timeframe hints (``when you can'', ``at your convenience'') \\
\texttt{moderate} & Moderate urgency (``soon'', ``shortly'') \\
\texttt{high} & Strong urgency (``ASAP'', ``urgent'') \\
\texttt{critical} & Extreme immediacy (``immediately'', ``right now'', ``without delay'') \\
\hline
\end{tabular}
\end{center}

\textbf{Entity Extraction (Separate Task)}

Extract entities from the full transcript, not turn-by-turn. Group into these categories (keys in entity block):
\begin{itemize}
  \item \texttt{people}
  \item \texttt{identifiers}
  \item \texttt{phone\_number}
  \item \texttt{email}
  \item \texttt{time\_info}
  \item \texttt{date}
  \item \texttt{location\_info}
  \item \texttt{products\_services}
  \item \texttt{monetary}
  \item \texttt{company\_organization}
  \item \texttt{others}
\end{itemize}
\textbf{User Prompt:}
\newline
Analyze the following transcript segment:\textbackslash n<transcript>\textbackslash n\{segment\_turns\}</transcript>

\end{tcolorbox}

\begin{tcolorbox}[breakable, colframe=orange!85!black, colback=orange!10!white, title=Prompt for Turn to Proposition Mapping ]

Your task is to map each turn in a transcript to the summary propositions it expresses.

You will receive:
\begin{enumerate}
  \item A set of numbered summary propositions.
  \item A transcript segment containing turns, each starting with a turn number like \texttt{"X: Speaker: ..."}, where \texttt{X} is the turn number.
\end{enumerate}

\textbf{Your Task:}
\begin{itemize}
  \item For each turn, identify which summary propositions (by their original number) are semantically expressed in that turn.
  \item A proposition matches a turn if the information in the proposition is present in the turn or can be reasonably inferred from it.
  \item Focus only on semantic content matching, not other analysis.
\end{itemize}

\textbf{Output Requirements:}
\begin{itemize}
  \item Produce a JSON object where:
  \begin{itemize}
    \item Keys are turn numbers (e.g., \texttt{"1"}, \texttt{"2"}).
    \item Values are lists of 0-based indices of matched summary propositions.
    \item If no matches are found for a turn, do not include that turn in the output.
  \end{itemize}
  \item A proposition can match multiple turns. If so, include its index in each relevant turn.
\end{itemize}

\textbf{JSON Format Example:}
\begin{verbatim}
{
  "0": [0, 2],
  "2": [1]
}
\end{verbatim}

\textbf{Example Input:}
\begin{itemize}
  \item Summary Propositions: \\
  \texttt{0. Agent name is Sarah.} \\
  \texttt{1. The sky is blue.} \\
  \texttt{2. The grass looks dead.}
  \item Transcript:
\begin{verbatim}
0: Agent: Hi, I am Sarah. Beautiful blue sky today!
1: Customer: The grass looks dead.
\end{verbatim}
\end{itemize}

\textbf{Example Output:}
\begin{verbatim}
{
  "0": [0, 1],
  "1": [2]
}
\end{verbatim}

\textbf{User Prompt:}
\newline
Map the following dialogue turns to the summary propositions:\textbackslash n<propositions>\{summary\_proposition\_string\}</propositions>\textbackslash n<transcript>\{segment\_turns\}

</transcript>

\end{tcolorbox}

\begin{tcolorbox}[breakable, colframe=orange!85!black, colback=orange!10!white, title=Prompt for Summary Labeling ]

Your task is to annotate each proposition (atomic-unit of summary) using a fixed set of conversational and linguistic dimensions.

Each proposition is about either the agent or the customer, and may express actions, emotions, or procedural events.

\#\#\# Dimensions (Fixed Order with Short Labels)

Each proposition must be annotated in the following order. Always include all dimensions for each proposition. Use [] for empty values in list-type fields.

\begin{center}
\begin{tabular}{|l|l|l|}
\hline
\textbf{Key} & \textbf{Dimension Name} & \textbf{Type} \\
\hline
\texttt{sent} & Sentiment & single value \\
\texttt{spk} & Speaker & single value \\
\texttt{topic} & Topic Category & single value \\
\texttt{agent} & Agent Action & single value \\
\texttt{sol} & Solution Type(s) & list \\
\texttt{lang} & Language Complexity Patterns & list \\
\texttt{polite} & Politeness & single value \\
\texttt{urgency} & Urgency & single value \\
\hline
\end{tabular}
\end{center}

\textbf{Output Format}

Return a compact JSON object with:

\textbf{Keys:} Proposition index as a string  
\newline
\textbf{Values}: List of 8 values in **fixed order**:  
  [sent, spk, topic, agent, sol, lang, polite, urgency]

\textbf{Example}

{
  "0": ["very\_pos", "customer", "empathy", "ask\_info", [], ["simple\_syntax"], minimal, high],
  \newline
  "1": ["neutral", "agent", "offers", "give\_info", ["advisory"], ["info\_dense"], standard, none],
  \newline
  "2": ["very\_neg", "agent", "diag", "check\_under", ["diag\_expl"], ["standard\_clear"], elevated, low]
}

\textbf{Allowed Values and Glossary}

\textbf{1. sent - Sentiment}

\begin{center}
\begin{tabular}{|l|p{11cm}|}
\hline
\textbf{Code} & \textbf{Meaning} \\
\hline
\texttt{very\_pos} & Strongly positive tone \\
\texttt{pos} & Moderately positive tone \\
\texttt{neg} & Moderately negative tone \\
\texttt{very\_neg} & Strongly negative tone \\
\texttt{info} & Information content or presence of factual tokens (dates, names, IDs) -- high priority to this over \texttt{neutral} \\
\texttt{neutral} & Does not have information and contains explicit neutral-emotion cues \\
\hline
\end{tabular}
\end{center}

\textbf{2. spk - Speaker}

agent, customer, misc

\textbf{3. topic - Topic Category}

\begin{center}
\begin{tabular}{|l|p{5cm}|}
\hline
\textbf{Code} & \textbf{Description} \\
\hline
\texttt{greet} & Greetings, introductions \\
\texttt{id\_verif} & ID or account verification \\
\texttt{issue} & Customer's reason for contact \\
\texttt{info\_gath} & Agent probing/investigating \\
\texttt{prod\_inq} & Product or service questions \\
\texttt{diag} & Diagnosis or troubleshooting \\
\texttt{soln} & Proposing a solution \\
\texttt{action} & Performing an action \\
\texttt{transact} & Payments, refunds, orders \\
\texttt{offers} & Service offers or upgrades \\
\texttt{sales} & Sales, upselling, persuasion \\
\texttt{resolve\_conf} & Confirming issue is resolved \\
\texttt{next} & Next steps, follow-ups \\
\texttt{close} & Farewell, call closure \\
\texttt{empathy} & Expressing care or rapport \\
\texttt{complaint} & Handling complaints/escalation \\
\texttt{policy} & Explaining rules or terms \\
\texttt{feedback} & Requesting feedback or surveys \\
\texttt{sched} & Appointments, scheduling \\
\texttt{billing} & Billing/payment issues \\
\texttt{compliance} & Compliance or regulations \\
\texttt{misc} & Miscellaneous \\
\hline
\end{tabular}
\end{center}

\textbf{4. agent - Agent Action}

\begin{center}
\begin{tabular}{|l|l|p{8cm}|}
\hline
\textbf{Code} & \textbf{Category} & \textbf{Notes} \\
\hline
\texttt{ask\_info} & Request Information & ``Could you confirm your order?'' \\
\texttt{give\_info} & Provide Information & Facts or explanations not tied to a fix \\
\texttt{check\_under} & Confirm Understanding & ``Do you see the change on your end?'' \\
\texttt{rapport} & Build Rapport & Empathy, friendliness, thank-you \\
\texttt{backchannel} & Acknowledgement / Cue & ``Uh-huh,'' ``Okay,'' ``Got it.'' \\
\texttt{escalate} & Escalate / Transfer Action & ``I'm connecting you to billing.'' \\
\texttt{compliance} & Compliance / Verification & Identity, policy, legal checks \\
\texttt{idle} & Passive / No-Op Response & Silence gaps marked or minimal reply \\
\texttt{other} & Other Conversational Act & Anything else (e.g., small talk) \\
\hline
\end{tabular}
\end{center}

\textbf{5. sol - Solution Type (multi-select)}

\begin{center}
\begin{tabular}{|l|p{5cm}|}
\hline
\textbf{Code} & \textbf{Description} \\
\hline
\texttt{diag\_expl} & Diagnostic explanation \\
\texttt{advisory} & General advice \\
\texttt{root\_cause} & Explaining root cause \\
\texttt{directive} & Concrete steps or commands \\
\texttt{preventive} & Prevent future issues \\
\texttt{escalate} & Escalation or transfer \\
\texttt{self\_help} & Do-it-yourself instructions \\
\texttt{partial} & Incomplete or partial fix \\
\texttt{rejected} & Offered but not applied \\
\texttt{followup} & Future action promised \\
\texttt{expect} & Sets realistic timelines \\
\texttt{reassure} & Emotional closure \\
\texttt{no\_soln} & No solution given \\
\hline
\end{tabular}
\end{center}

\textbf{6. lang - Language Complexity (multi-select)}

\begin{center}
\begin{tabular}{|l|p{8cm}|}
\hline
\textbf{Code} & \textbf{Description} \\
\hline
\texttt{standard\_clear} & Clear, direct, and easily understood language. \\
\texttt{simple\_syntax} & Predominantly short, declarative sentences. \\
\texttt{complex\_syntax} & Long, multi-clause, or convoluted sentences. \\
\texttt{technical\_terms} & Specialized terms related to a specific domain. \\
\texttt{industry\_jargon} & Terms/phrases specific to an industry/company. \\
\texttt{acronyms\_abbreviations} & Use of shortened forms of words or phrases. \\
\texttt{info\_dense} & Highly concise; packed with specific information. \\
\texttt{verbose\_hedging} & Wordy, uses fillers, qualifiers, or vague language. \\
\texttt{formal\_register} & Polished, professional, often more structured. \\
\texttt{informal\_colloquial} & Conversational, casual, everyday language. \\
\texttt{empathetic\_softening} & Language used to show understanding or soften news. \\
\texttt{abrupt\_blunt} & Overly direct, lacking typical softeners/politeness. \\
\texttt{idioms\_slang} & Figurative expressions or informal slang. \\
\texttt{passive\_voice\_prominent} & Significant use of passive voice construction. \\
\hline
\end{tabular}
\end{center}

\textbf{7. polite - Politeness}

\begin{center}
\begin{tabular}{|l|p{11cm}|}
\hline
\textbf{Code} & \textbf{Description} \\
\hline
\texttt{none} & No politeness cues (no please/thank you/etc.) \\
\texttt{minimal} & One-off courtesy (“thank you”, “please”) \\
\texttt{standard} & Expected level (“please let me know”, “thanks for waiting”) \\
\texttt{elevated} & Multiple markers + honorifics (“sir/madam”, “kindly”) \\
\texttt{impolite} & Impoliteness cues \\
\hline
\end{tabular}
\end{center}

\textbf{8. urgency - Urgency}

\begin{center}
\begin{tabular}{|l|p{11cm}|}
\hline
\textbf{Code} & \textbf{Description} \\
\hline
\texttt{none} & No urgency language \\
\texttt{low} & Mild timeframe hints (“when you can”, “at your convenience”) \\
\texttt{moderate} & Moderate urgency (“soon”, “shortly”) \\
\texttt{high} & Strong urgency (“ASAP”, “urgent”) \\
\texttt{critical} & Extreme immediacy (“immediately”, “right now”, “without delay”) \\
\hline
\end{tabular}
\end{center}

\textbf{Important Instructions}

* Always include all 8 fields per proposition in the exact order: sent, spk, topic, agent, sol, lang, polite, urgency  
\newline
* For sol and lang, output a list of applicable codes or an empty list ([]) if none apply. 
\newline
* Use only the short-form codes provided above.
\newline
+
\newline
IMPORTANT: You must analyze ALL \{len(summary\_propositions)\} propositions in the list. Do not skip any propositions. " 
        Output a JSON object where keys are proposition indices (0-based, from 0 to \{len(summary\_propositions)-1\}) and values are objects containing: You must include entries for indices 0 through \{len(summary\_propositions)-1\}. 

\textbf{User Prompt:}
\hspace*{1em}
'Please analyze the sentiment and determine the speaker for ALL \{len(summary\_propositions)\} propositions below. ' \\
\hspace*{1em}'Make sure to include entries for indices 0 through \{len(summary\_propositions)-1\}:\textbackslash n\textbackslash n\{summary\_propositions\}' \\

\end{tcolorbox}

\clearpage

\section{Bias Mitigation}
\label{appendix:mitigation}

To demonstrate that the fine-grained analysis provided by our \textit{BlindSpot} framework is actionable, we conducted a preliminary experiment in bias mitigation. The goal was to use the specific, systemic biases identified in our main analysis to construct a targeted system prompt and then measure its impact on model behavior.

\subsection{Constructing a Targeted System Prompt}
Our main findings revealed consistent patterns of bias across most models, such as over-representing negative sentiment while under-representing agent rapport-building and resolution steps from the middle of the conversation. Based on these insights, we constructed a single, detailed system prompt designed to explicitly counteract these observed tendencies.

The prompt, shown in full in Box~\ref{box:bias-mitigation-prompt}, instructs the model to focus on high-fidelity, balanced summarization and provides a checklist of ``Correction and Balancing Guidelines.'' These guidelines directly map to the bias dimensions where we observed the most significant issues, such as \textit{Sentiment Balance}, \textit{Positional Coverage}, and \textit{Topic and Activity Coverage}. By making the model explicitly aware of its potential blind spots, we hypothesized that we could steer its summarization process towards a more faithful representation of the source transcript.

\begin{tcolorbox}[breakable, colframe=orange!85!black, colback=orange!10!white, 
  title=Constructed System Prompt for Bias Mitigation, 
  label={box:bias-mitigation-prompt}
]

Your task is to summarize the following dialog with a focus on high fidelity and balance. Based on an analysis of previous outputs, apply the following corrections to ensure a more accurate and balanced summary.

\textbf{Correction and Balancing Guidelines}

\begin{enumerate}
    \item \textbf{Sentiment Balance:}
    \begin{itemize}
        \item Ensure both positive and negative sentiments are represented if they appear in the transcript.
        \item \textbf{Specifically Include:} Positive sentiments expressed by the customer, especially those related to agreement or satisfaction with a solution.
    \end{itemize}

    \item \textbf{Speaker Representation:}
    \begin{itemize}
        \item Provide a balanced representation of contributions from both the customer and the agent.
        \item \textbf{Specifically Include:} Key agent responses, clarifying questions, and de-escalation efforts.
    \end{itemize}

    \item \textbf{Positional Coverage:}
    \begin{itemize}
        \item Draw information equitably from all parts of the conversation.
        \item \textbf{Specifically Include:} Details from the Mid, Late, and Very Late segments of the conversation, which often contain resolution steps and final agreements.
    \end{itemize}

    \item \textbf{Topic and Activity Coverage:}
    \begin{itemize}
        \item Broaden the scope of topics and activities included in the summary.
        \item \textbf{Topics to Include:} Information gathering/probing by agent, Call closure, ID verification, and Expression of empathy.
        \item \textbf{Agent Activities to Include:} Rapport-building, Asking for information, and Checking for understanding.
    \end{itemize}

    \item \textbf{Solution and Repetition Types:}
    \begin{itemize}
        \item \textbf{Solution Types:} Ensure representation of directive solutions (concrete, actionable steps).
        \item \textbf{Repetition Types:} Include all forms of significant repetition, such as:
        \begin{itemize}
            \item customer repeating self,
            \item agent repeating customer, and
            \item customer repeating agent.
        \end{itemize}
    \end{itemize}

    \item \textbf{Linguistic and Structural Elements:}
    \begin{itemize}
        \item \textbf{Disfluencies:} Include meaningful interjections (oh!, hmm) and incomplete starts if they indicate hesitation or a change of thought.
        \item \textbf{Turn Length:} Represent information from both very long and very short conversational turns if they are relevant.
        \item \textbf{Chronological Order:} Narrate events in the sequence they occurred in the transcript. Do not reorder them.
    \end{itemize}

    \item \textbf{Factual and Emotional Fidelity:}
    \begin{itemize}
        \item \textbf{Entity Representation:} Include a wider range of entities beyond people and organizations.
        \item \textbf{Specifically Include:} Dates, Locations, Product/Case IDs, Monetary values, Times, Phone numbers, and Emails.
        \item \textbf{Emotional Tone:} Reflect the emotional state of the speakers accurately. Avoid amplifying, attenuating, or neutralizing emotions expressed in the transcript.
    \end{itemize}
\end{enumerate}

\textbf{Final Instruction} \\
Produce a summary of the following dialog that strictly adheres to all the guidelines above. The final output should be a balanced, factually accurate, and structurally faithful representation of the original conversation.

\end{tcolorbox}

\subsection{Mitigation Results}
We applied the mitigation prompt to a representative subset of ten models from our main evaluation, including small and large variants from four major model families. We then re-calculated the average \textit{Fidelity Gap} (JSD) and \textit{Coverage} scores and compared them to the baseline performance.

Table~\ref{table:mitigation_results} details the absolute change in performance for each model after applying the targeted prompt. A negative change in JSD indicates a reduction in bias (an improvement), while a positive change in Coverage indicates better information retention (an improvement). The results clearly show that this simple intervention was effective. All evaluated models exhibited a net improvement, with lower average JSD and higher average Coverage. 

\begin{table*}[!hbt]
  \centering
  \small
  \renewcommand{\arraystretch}{1.2}
  \begin{tabular}{@{}l c c@{}}
    \toprule
    \textbf{Model} & 
    \textbf{\(\Delta\) Avg. Coverage (\%)} & 
    \textbf{\(\Delta\) Avg. JSD} \\
    & (\(\uparrow\) better) & (\(\downarrow\) better)  \\
    \midrule
    \texttt{claude-4-sonnet} & +4.87 & -0.0118 \\
    \texttt{nova-pro} & +4.09 & -0.0070  \\
    \texttt{llama-4-maverick} & +3.59 & -0.0070  \\
    \texttt{nova-lite} & +3.06 & -0.0030  \\
    \texttt{llama-3.2-3b} & +2.36 & -0.0012 \\
    \texttt{gpt-4.1} & +2.00 & -0.0011  \\
    \texttt{claude-3.5-haiku} & +1.99 & -0.0039  \\
    \texttt{gpt-4.1-mini} & +1.57 & +0.0027  \\
    \texttt{gpt-4o} & +1.45 & -0.0003  \\
    \texttt{o4-mini} & +1.07 & -0.0089  \\
    \midrule
    \textbf{Average Change} & \textbf{+2.61} & \textbf{-0.0041}  \\
    \bottomrule
  \end{tabular}
  \caption{Impact of the targeted mitigation prompt on model performance. The table shows the absolute change (\(\Delta\)) in average Coverage, and average JSD (Fidelity Gap) compared to the baseline. With one minor exception (\texttt{gpt-4.1-mini} on JSD), the prompt led to improved fidelity and quality across all models.}
  \label{table:mitigation_results}
\end{table*}

The results also highlight the scaling effect discussed in the main paper. More capable models like \texttt{claude-4-sonnet} demonstrated the largest improvements, suggesting they are better able to follow the complex set of corrective instructions. This experiment, while not a comprehensive study on mitigation techniques, successfully validates the core premise of our work: that by systematically identifying and understanding specific operational biases, we can generate targeted, actionable feedback to create more faithful and reliable summarization systems.

\subsection{Impact on Summary Compression}
\label{subsec:mitigation_compression}

A key question when adding detailed instructions to a prompt is whether it impacts summary length and conciseness. An ideal intervention would reduce bias without making summaries overly verbose. To measure this, we analyzed the change in the \textbf{Compression Factor} (the ratio of transcript tokens to summary tokens; higher is more compressed) before and after applying the mitigation prompt.

Table~\ref{table:mitigation_compression} shows that for the majority of models, the targeted prompt led to summaries that were \textbf{less compressed} (i.e., a lower Compression Factor). For instance, the compression factor for \texttt{nova-pro} decreased significantly from 34.7 to 15.2, and for \texttt{llama-3.2-3b}, it fell from 20.5 to 7.7. This result is expected; the prompt explicitly asks for more information to be included (e.g., details from later parts of the conversation, more entity types, rapport-building moments), which naturally increases the length of the output summary.

Interestingly, this trend was not universal. Models like \texttt{gpt-4o} and \texttt{o4-mini} became \textit{more} compressed, suggesting they were able to integrate the complex instructions more efficiently without a linear increase in length. This indicates a potential difference in how various model architectures handle detailed, constraint-based prompting.

Overall, while the prompt successfully reduced bias, it often came at the cost of reduced compression. This highlights a fundamental trade-off between summary fidelity and succinctness, suggesting that future work could focus on achieving bias mitigation while adhering to stricter length constraints.

\begin{table*}[!ht]
  \centering
  \small
  \renewcommand{\arraystretch}{1.2}
  \begin{tabular}{@{}l c c c@{}}
    \toprule
    \textbf{Model} & 
    \textbf{Compression Factor (Baseline)} & 
    \textbf{Compression Factor (Mitigated)} & 
    \textbf{\(\Delta\) Compression Factor} \\
    \midrule
    \texttt{o4-mini} & 22.1 & 64.9 & +42.82 \\
    \texttt{gpt-4o} & 29.8 & 31.5 & +1.65 \\
    \texttt{gpt-4.1-mini} & 22.3 & 22.4 & +0.12 \\
    \texttt{gpt-4.1} & 18.9 & 18.9 & -0.05 \\
    \texttt{llama-4-maverick} & 23.2 & 11.2 & -11.99 \\
    \texttt{llama-3.2-3b} & 20.5 & 7.7 & -12.77 \\
    \texttt{claude-sonnet-4} & 18.5 & 8.5 & -9.97 \\
    \texttt{claude-3.5-haiku} & 24.3 & 18.1 & -6.21 \\
    \texttt{nova-lite} & 28.0 & 13.5 & -14.50 \\
    \texttt{nova-pro} & 34.7 & 15.2 & -19.50 \\
    \bottomrule
  \end{tabular}
  \caption{Change in summary compression after applying the targeted mitigation prompt. The table shows the Compression Factor (higher is more compressed) before and after the intervention. A negative delta (\(\Delta\)) indicates that the mitigated summary was longer and less compressed than the baseline.}
  \label{table:mitigation_compression}
\end{table*}

\end{document}